\documentclass[letterpaper,twocolumn,10pt]{article}
\usepackage{usenix2019}
\usepackage{comment}

\newcommand{\name}{\textsc{Gati}}

\usepackage{algorithm, algorithmicx, algpseudocode}
\usepackage[T1]{fontenc}
\usepackage{amsmath, amssymb, amsthm, amsfonts}
\usepackage{multirow}
\usepackage{hyperref}
\usepackage{color}
\usepackage{xcolor}
\usepackage{alltt}
\usepackage[draft]{pgf}
\usepackage{tikz}
\usetikzlibrary{backgrounds}
\usetikzlibrary{calc}
\usetikzlibrary{positioning}
\usetikzlibrary{automata}
\usetikzlibrary{shapes}
\usetikzlibrary{arrows}
\usetikzlibrary{shadows}
\usetikzlibrary{decorations.pathreplacing}
\usetikzlibrary{decorations.text}
\usetikzlibrary{shapes.multipart}
\usetikzlibrary{fit}
\usetikzlibrary{patterns}
\usepackage{subfig}
\usepackage{amsmath}
\usepackage[medium, compact]{titlesec}
\titlespacing*{\section}{1pt}{3.5pt}{2pt}
\titlespacing*{\subsection}{1pt}{3pt}{1.5pt}
\titlespacing*{\subsubsection}{1pt}{3pt}{1.5pt}
\usepackage[font=footnotesize]{caption}
\titlespacing*{\section}{1pt}{3.5pt}{2pt}
\titlespacing*{\subsection}{1pt}{3pt}{1.5pt}
\titlespacing*{\subsubsection}{1pt}{3pt}{1.5pt}

\newcommand{\mytilde}{\raise.17ex\hbox{$\scriptstyle\mathtt{\sim}$}}

\newcommand{\aditya}[1]{\textcolor[rgb]{0.00,0.00,1.00}{\{AA: #1\}}}

\newcommand{\arjun}[1]{\textcolor[rgb]{0.00,0.5,0.5}{\{AB: #1\}}}

\newcommand{\myvec}[1]{\protect\overrightarrow{#1}}

\newcommand{\secref}[1]{{\S\ref{#1}}}
\DeclareMathOperator*{\argmax}{arg\,max}

\usepackage{xcolor}
\usepackage{etoolbox} 

\makeatletter 
\patchcmd{\maketitle}%
  {\def\@makefnmark{\rlap{\@textsuperscript{\normalfont\@thefnmark}}}}%
  {\def\@makefnmark{\rlap{\@textsuperscript{\normalfont\color{red}\@thefnmark}}}}%
  {}
  {}
\makeatother 

\begin{document}

\title{\Large \bf Accelerating Deep Learning Inference via Learned Caches}
\author{
{Arjun Balasubramanian\thanks{Equal Contribution. Work done while at UW-Madison.}~}\\
Amazon Web Services
\and
{Adarsh Kumar\footnotemark[1]}\\
Amazon
\and
{ Yuhan Liu}\\ University of Wisconsin - Madison
\and
{ Han Cao}\\ University of Wisconsin - Madison
\and
{Shivaram Venkataraman}\\ University of Wisconsin - Madison
\and
{Aditya Akella}\\ University of Wisconsin - Madison}


\maketitle


\noindent{\bf Abstract:}
Deep Neural Networks (DNNs) are witnessing increased adoption in multiple domains
owing to their high accuracy in solving real-world problems. 
However, this high accuracy has been achieved by building deeper networks,
posing a fundamental challenge to the low latency inference desired by user-facing applications.
Current low latency solutions trade-off on accuracy or
fail to exploit the inherent temporal locality in prediction serving workloads.

We observe that caching hidden layer outputs of the DNN can introduce a form of late-binding
where inference requests only consume the amount of computation needed. This enables a mechanism for achieving low latencies, coupled with an ability to exploit temporal locality.
However, traditional caching approaches incur high memory overheads and lookup latencies,
leading us to design \emph{learned caches} - caches that consist of simple ML models that are continuously updated.
We present the design of \name{}, an end-to-end prediction serving system
that incorporates learned caches for low-latency DNN inference.
Results show that \name{} can reduce inference latency by up to 7.69$\times$
on realistic workloads.

\section{Introduction}

Machine learning models based on deep neural networks (DNNs) have
surpassed human-level accuracy on tasks ranging from speech
recognition~\cite{speechparity}, image classification~\cite{resnet,inception} to machine translation~\cite{translationparity}.
As a result,
several enterprises are now deploying DNNs as a part of
their applications, many of which are user-facing and hence
latency-sensitive~\cite{deepcpu, grandslam, clipper}. This gain in accuracy
has largely come from models becoming more complex, typically
\emph{deeper} or having more layers. As each layer of DNN inference
depends on the output from the previous layer, such deep models
exacerbate inference latency, hurting the performance of applications
that rely on them. For instance, the top-5
classification accuracy for ImageNet~\cite{imagenet} increased from 71\% in 2012 to
97\% in 2015 but the models became 20$\times$ more computationally
expensive, and inference latency has correspondingly worsened
\textasciitilde 15$\times$. 

A number of recent efforts have been aimed at reducing inference latency,
such as: (a) quantization-based approaches that perform computation at
lower precision~\cite{bnn,twn}, (b) model pruning-based approaches that prune
dependencies across layers~\cite{stateofpruning}, and (c) distillation
based techniques that teach a smaller model to behave like a larger
model~\cite{distillation} (\secref{subsubsec:current_solutions}).
These approaches face two key issues. (1) 
All the above approaches trade-off some fixed amount of accuracy for
latency improvements as shown in Figure~\ref{fig:motivation}b.
In other words, they bind to a {\em specific point in the
latency-accuracy trade-off space} for {\em all} prediction inputs.
However, as prior work has
shown~\cite{noscope,cascades}, cheaper DNNs (e.g., Resnet-18)
suffice for easier inputs while deeper models (e.g., Resnet-152) are
only necessary to handle more difficult examples. This suggests that
postponing the binding to during inference can, in theory, lead to lower
latencies for easier inputs without significantly
impacting accuracy.
 (2) Prediction serving
systems typically serve user facing applications which are often
temporally dominated by a small number of
classes~\cite{cascades,focus}. Thus requests have a notion of {\em
  temporal locality}.  Existing approaches like distillation can leverage this temporal
locality by retraining or fine-tuning the model. However, this only
improves the accuracy, not the latency of future requests, and retraining could
also be expensive for deep models~\cite{learninginsitu}.  


\begin{figure}[t]
  \centering
   \includegraphics[width=0.35\textwidth]{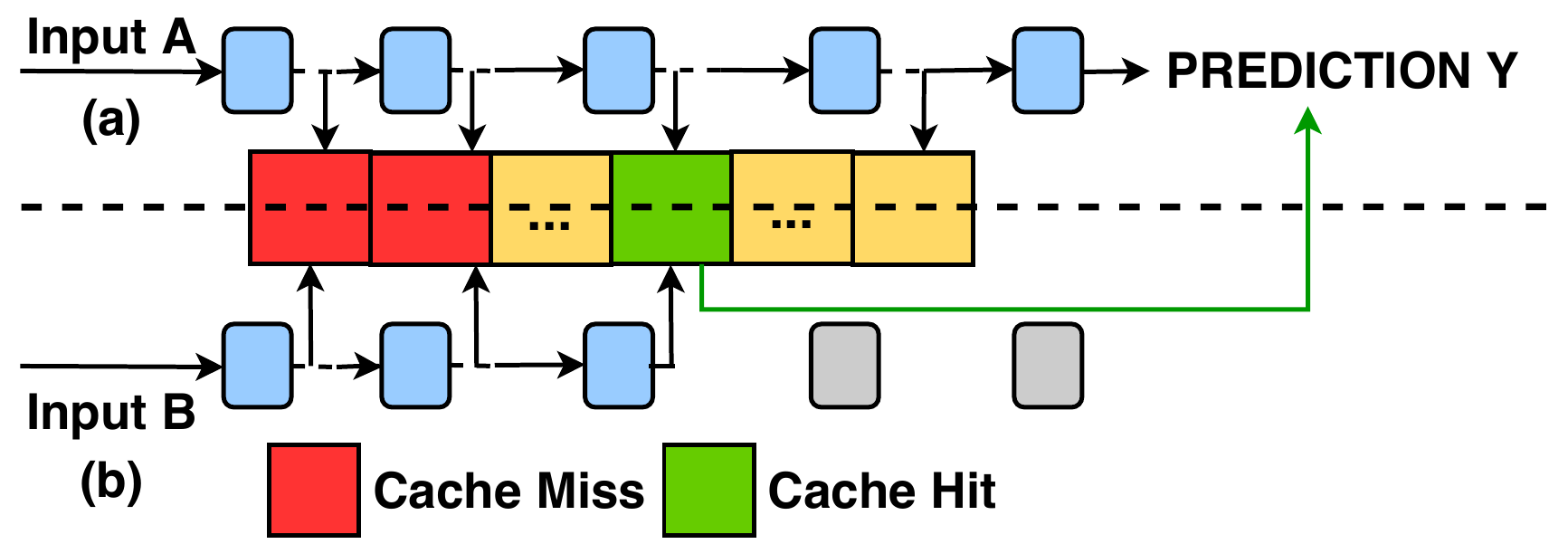}
	\caption{\textbf{A caching mechanism where (a) Hidden layer outputs along with associated predictions are cached.
	(b) During inference, we do a cache lookup after every layer. Cache hits yield faster predictions.}}
    \label{fig:traditional_caching}
\end{figure}

Traditionally, caches are used in web-services
to improve latency for workloads with temporal locality. This paper
explores using caches to {\em both} provide lower latencies for easier
inputs and exploit temporal locality.  To apply such a caching-based
approach to inference, we propose that during inference, if the result
of any intermediate computation matches what we have seen earlier, we
can skip the computations of the remaining layers of the DNN
(Figure~\ref{fig:traditional_caching}). Caches at deeper layers in a
model capture information about progressively harder-to-classify
inputs. Our approach to caching thus paves the way for {\em
  late-binding the work performed for inference based on the input's
  hardness}, allowing for lowering of prediction latency, depending
upon the layer at which we observe a cache hit, without impacting
accuracy. Intermediate (or hidden) layer caches can be initially
populated using user-provided validation data and updated to hold the
intermediate outputs for recent inputs, thereby exploiting temporal
locality.



However, using a direct approach of storing hidden layer outputs as a
cache leads to many challenges: hidden layer outputs in DNNs are
high dimensional (e.g., 262144 floats for
  block 3\footnote{ResNet-50 consists of 16 blocks, where each block has 3 convolutional layers and a residual connection.} of ResNet-50) needing significant memory for storage and high
latency at lookup (\secref{subsec:caching-proposal}). To address these
challenges, we propose using {\em simple machine learning models to represent
  the cache} at each layer. Given the hidden layer output, such
a {\em learned cache} model at a layer predicts if we have a cache-hit,
and if so the final result to use. This provides improved latency during
cache-hits;
cache misses continue inference on the base
model leading to no loss in accuracy.

Using ML models for the learned cache means that we need a methodology
to both determine the predicted output of the base model and if this
can be considered as a cache hit.  We address this problem using a
novel approach where the learned cache for a given layer of the base
DNN uses two trained networks, a \emph{predictor} network that
generates the predicted output and a \emph{selector} network that
decides if the prediction can be used as a cache hit.

We develop {\name}, a system for DNN inference that automatically
integrates learned caches into the inference workflow given a {\em
  pre-trained} base DNN model and a validation dataset. During an
initial deployment phase, {\name} automatically explores model
architectures for predictor and selector networks, estimating their
hit-rate and accuracy using the validation dataset.
Using this, {\name} determines the
optimal set of layers that should include a learned cache and the
architecture to use to minimize the overall average latency, taking
into account memory and compute resources available.

At runtime, \name{} performs two main functions: query planning when
using a directed acyclic graph (DAG) of models to process a user
query, and incremental retraining of learned caches to provide
temporal locality. Using learned caches enables run-time refinement of
query plans, or {\em incremental replanning}, in systems like
Nexus~\cite{nexus} that execute a DAG of models. We describe how
\name{} can incrementally improve prediction accuracy given a latency
SLO for a DAG of models, based on which models in the DAG obtained a
cache hit.

Finally, temporal shifts in input data might
lead to worse hit rate for learned caches.  For example, if we train
an object detection model where the validation data consists of video
snippets of car traffic, our learned caches will give a higher hit rate for
input frames with cars. However the class distribution might change
over time. We handle that in {\name} by designing an online retraining
scheme that periodically and quickly retrains predictor and selector
networks when their hit-rate is worse than the expected hit rate.


We evaluate {\name} against both systems and algorithmic techniques for low
latency inference using popular datasets~\cite{cifar10-data,speech-data}
and video streams~\cite{bend-data,oxford-data,sittard-data}.
Results show that \name{} can opportunistically get cache hits for easier inputs
and improve average latency by up to {\bf 1.95$\times$} over the base model
and {\bf 1.39$\times$} over existing low latency techniques on image 
datasets.
On real-world videos that exhibit temporal locality, \name{} gives up to
{\bf 7.69$\times$} improvement in average latency.
Additionally, when using a DAG of models (Figure~\ref{fig:motivation}a), we see that 
incremental replanning 
 simultaneously improves the overall accuracy, with respect to ground truth, by {\bf \textasciitilde
 1\%} and prediction latency by {\bf 1.26$\times$}.

\section{Background and Motivation}
A Deep Neural Network (DNN) consists of a sequence of layers,
each of which is a mathematical function on the output of the previous layer.
We refer to the first layer of the DNN that accepts the input 
as the {\em input layer} and the final layer that represents 
the prediction as the {\em output layer}.
We term the remaining intermediate layers of the DNN as {\em hidden layers}.


\subsection{Prediction Serving and DNN Inference}
\label{subsec:prediction-serving}

\begin{figure}[t!]
	\centering
	\subfloat[][]{\includegraphics[width=0.30\columnwidth]{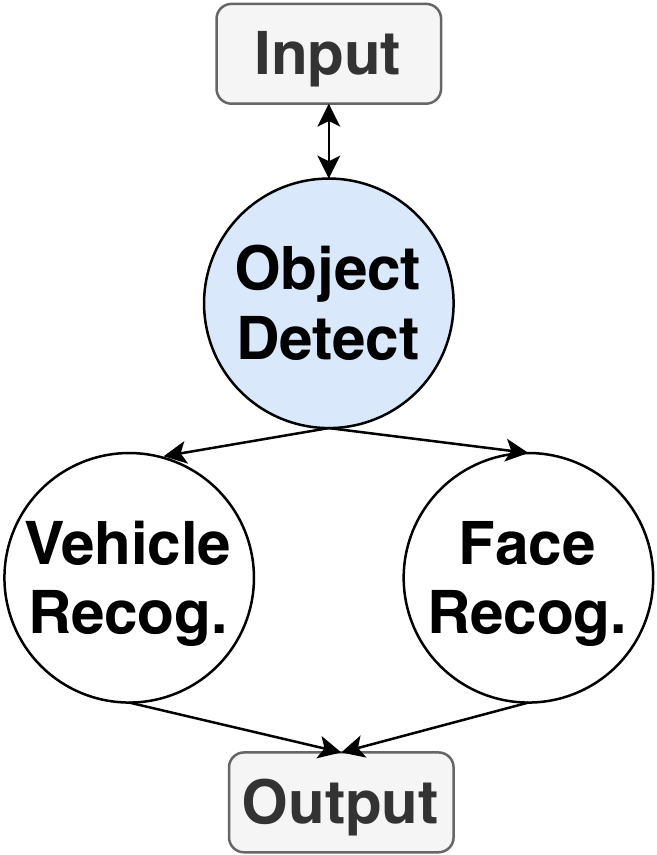}}
	\subfloat[][]{\includegraphics[width=0.58\columnwidth]{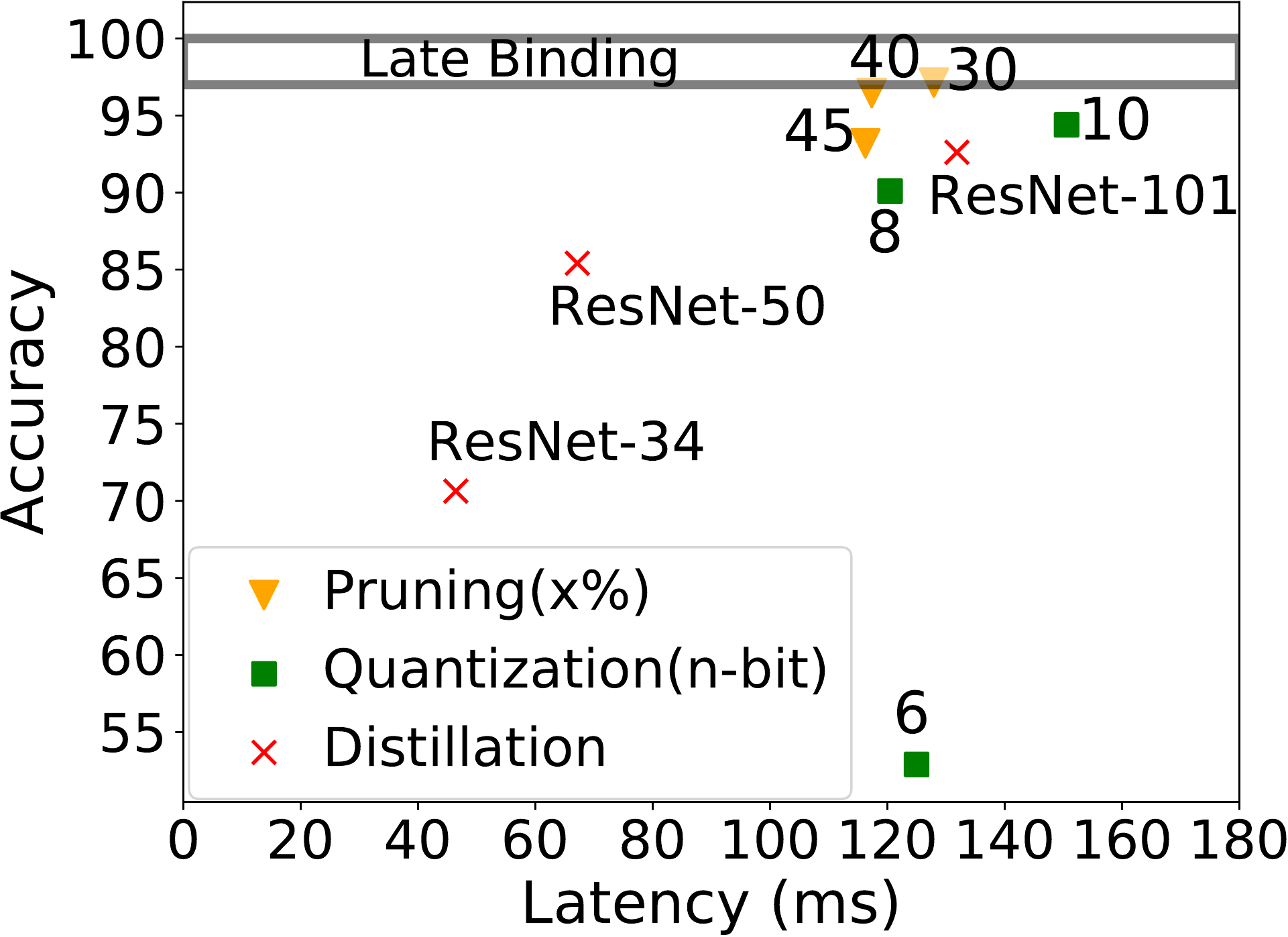}}
	\caption{\label{fig:motivation} \textbf{(a) DAG of models to be executed for a traffic
	analysis application
	(b) Latency-accuracy trade-off for ResNet-152 on CIFAR-100.
	Accuracies are wrt. base ResNet-152 model. \name{} enables late-binding.}}
\end{figure}

A prediction serving system serves machine learning models used for
inference~\cite{clipper,nexus,tfserving}.  It accepts queries from clients
which may require performing inference on a single model or a
sequence of models.  For instance, consider a traffic analysis
application (Figure~\ref{fig:motivation}a).  The
analysis would first require object detection followed by either face
or vehicle recognition depending upon the object, in order to deliver
the final output.  We can therefore view the query as a DAG of ML
models that need to be executed.  Client requests are often associated
with a latency SLO within which the entire query needs to be
executed. Prediction serving systems are deployed on a fleet of
servers consisting of CPUs and accelerators (GPUs, TPUs, custom
ASICs~\cite{tpu,brainwave,inferentia}). 

DNN inference in prediction serving systems must meet two
requirements: (1) {\bf Low Latency. } As trained models
often support user-facing applications, low inference
latency (~\textasciitilde 10-100ms)~\cite{deepcpu} is key.
However, executing DNN inference is computationally intensive
and imposes high latency today~\cite{nexus,noscope,deepcpu}.
Though some hardwares (e.g., GPUs) offer higher throughput when
executing requests in batches, many systems perform minimal
or no batching to keep the promise of low latency~\cite{deepcpu,paritymodels,brainwavebatching,fbcpu}.
(2) {\bf High Accuracy. }
With DNNs being increasingly deployed in critical applications such as
fraud detection and personal assistants, predictions need to have
high accuracy.


\subsubsection{Current state of DNN Inference}
\label{subsubsec:current_solutions}



Increasingly, {\em deeper} DNN models with many layers are being used
to achieve high accuracy, but they severely impact latency. Viewing
the dependencies among DNN computations performed during inference as
a computation graph~\cite{tensorflow}, inference latency depends on:
(i) the per-node latencies in the graph, and (ii) the critical path
length.  
Existing techniques to reduce inference latency can be categorized
on how they manipulate the computation graph.  All of these
techniques come with an inherent trade-off in accuracy as shown in
Figure~\ref{fig:motivation}b:

\noindent{\bf Reducing cost of individual nodes.}
This is used in model quantization~\cite{bnn,twn,integerquantization} based techniques which
perform computations at lower precision
rather than 32-bit floating point,
thereby reducing the cost of individual computations. 
We see (Figure~\ref{fig:motivation}b) that quantization can 
result in \textasciitilde 5-10\% loss in accuracy,
while giving 1.2$\times$-1.5$\times$ latency improvement.


\noindent{\bf Reducing length of critical path. } Model
distillation~\cite{distillation} is an example here, where a smaller DNN (say
ResNet-50) termed the {\em student} is taught by a deeper trained DNN
(say ResNet-152).
We again see that it can result in \textasciitilde
8-30\% loss in accuracy,
while giving 2.7$\times$-4.1$\times$ improvement in latency.

\noindent{\bf Reducing the number of nodes and edges. } Network
pruning~\cite{stateofpruning} encompasses a broad set of techniques that fall into
this category.  For example, Network Slimming~\cite{networkslimming} exploits
channel-level sparsity and removes insignificant channels from
convolutional neural networks.
Our experiments show that pruning can result in
\textasciitilde 3-7\% loss in accuracy,
while giving 2.6$\times$-2.7$\times$ improvement in latency.

\subsubsection{Opportunities in DNN Inference}
\label{subsubsec:opportunities_motivation}
Next, we present opportunities to improve DNN inference:

\noindent {\bf (O1) Opportunity from DNN model architecture. }
Existing solutions for lowering latencies pick a {\em fixed} point on
the latency-accuracy trade-off space
(Figure~\ref{fig:motivation}b). Thus, they {\em early-bind}, as they
require committing to a particular latency and accuracy value prior to
DNN deployment.

We observe that deeper DNNs are built to offer an incremental
accuracy benefit over their shallower counterparts.  For instance,
ResNet-50~\cite{resnet} has an accuracy of 94.2\% on CIFAR-10~\cite{cifar10-data} while
a shallower ResNet-18 has an accuracy of 93.3\%.  
Intuitively, ResNet-18 seems to suffice for obtaining accurate
predictions for 93.3\% seemingly "easy" inputs, while the "extra"
layers in ResNet-50 offer an accuracy benefit for an additional
\textasciitilde 1\% "hard" inputs.  This raises the question - {\em
  can we obtain lower latencies without trading-off accuracy
  by opportunistically deciding the number of layers to compute?}


Achieving this necessitates {\em late-binding} the decision of the
number of layers that need to be computed at inference time, thereby
obtaining progressively lower latencies for easier inputs without
trading off on accuracy (Figure~\ref{fig:motivation}b). In
practice we find that late-binding can 
improve average latency by \textasciitilde 1.95$\times$
for real-world workloads (\secref{subsec:eval_late_binding}).

Today, using an ensemble of models~\cite{clipper} or cascades of
specialized models~\cite{noscope,cascades} are some techniques that enable a
limited spectrum of progressive latencies. However, 
as shown in our experiments (~\secref{subsec:eval_late_binding}),
these techniques either inflate tail latencies or require extra
resources. 



\noindent {\bf (O2) Opportunity from workload. }
Since prediction serving systems typically serve
user-facing web applications, requests have a skewed distribution
dominated by few classes~\cite{webtemporallocality,OpenImages}.
Video analytics represent another important workload class
that uses prediction serving systems.
Both workloads exhibit {\em temporal locality},
where certain classes may be
popular for a given window of time~\cite{focus,cascades}.
Current solutions can adapt to temporal locality by
retraining or fine-tuning the model~\cite{learninginsitu}.
However, this only improves accuracy and not
latency and retraining can be expensive for deep models~\cite{learninginsitu}.

We note that temporal locality has been exploited in other domains
~\cite{usenixtemporallocality,webtemporallocality,nvmtemporallocality,3dtemporallocality,graphtemporallocality} 
like web-caching to extract latency benefits. Given an ability to late-bind
the amount of computation depending on whether an input is "easy" or
"hard", we see an opportunity to similarly exploit temporal locality
and opportunistically obtain lower latencies for the "easier"
frequently occurring requests.

\noindent {\bf (O3) Opportunity from system architecture. } Prediction
serving systems typically consist of a query scheduler responsible for
orchestrating the execution of multiple DNN models constituting a
query~\cite{nexus}.  The scheduler's {\em query planner} determines
how to apportion the available query latency SLO amongst different DNN
models that constitute the query, and computes a query evaluation
plan (QEP), i.e., which DNN model is to be executed at each stage of
the query.

Current approaches offer fixed latencies and hence the query planner
determines a fixed QEP prior to query execution~\cite{nexus}.
Late-binding however provides an opportunity for {\em incremental
  replanning} - this can distribute the ``saved'' latency from
early execution of an "easy" input to downstream model
executions. This could help use more expensive/complex downstream DNNs,
resulting in higher accuracy for the query while not violating
the latency SLO (\secref{subsec:scheduler}).

\subsection{A Proposal for Caching}
\label{subsec:caching-proposal}

To exploit the above opportunities, we propose adopting the use
of caches to complement DNNs during inference.
Inference involves performing a forward-pass on the DNN.
During the forward pass, if the result of any intermediate computation
matches what we have seen earlier, we can skip the computations of the
remaining layers of the DNN and directly arrive at the final prediction.


Inspired by the design of multi-level caches, we propose associating a cache
with each hidden layer of the DNN. 
Figure~\ref{fig:traditional_caching} illustrates a simple mechanism
to realize the caching of DNN computations.
As a pre-processing step, we can cache the hidden layer outputs 
along with the final predictions from the validation dataset.
During inference, 
if a hidden layer output matches what was cached earlier, we deem this as a {\em cache hit}
and skip performing computations for the remaining layers and directly deliver the final prediction.
Caching thus paves the way for {\em late-binding inference computation}.


\begin{table}[t]
\centering
\begin{scriptsize}
\begin{tabular}{ l | c | c | c }
	\textbf{Scheme} & {\bf Lookup Latency (GPU)} & {\bf Memory} & \textbf{Accuracy} \\
\hline
\textbf{k-NN (k=50)} & 2075.5 ms & 5000 MB & 79.84 \% \\
\hline
\textbf{LSH} & 71.4 ms & 5000 MB & 39.77\% \\
\hline
{\bf k-means (k=100)} & 1.511 ms & 333 MB & 76.06\% \\
\end{tabular}
\end{scriptsize}
	\caption{\textbf{Overheads incurred by caching mechanisms at a single layer [ResNet-18, Block 3].
	$k$-NN returns the majority label amongst $k$ nearest neighbors as prediction. $k$-means clusters
	hidden layer outputs and returns a
	representative label from the nearest cluster as prediction.}}
\label{tab:trad_vs_learned}
\end{table}

\noindent {\bf Challenges in caching DNN computations. }
While caching holds promise as also observed in prior work~\cite{freeze-vision},
{\em high dimensionality} of hidden layer outputs
introduces a number of challenges in designing caches.
Cache lookups for DNN computations require distance-based similarity
search such as $k$-nearest neighbors ($k$-NN), $k$-means clustering
or locality sensitive hashing (LSH)~\cite{lsh}
to infer cache hits and obtain predictions (Table~\ref{tab:trad_vs_learned}).
High dimensionality means that the memory overhead associated with such caches
would be high (as in $k$-NN). This makes caching challenging on
accelerators such as GPUs that have limited memory. 
Further, the fidelity of distance-based
search degrades in high-dimension~\cite{highdimension}, which could
result in errors in cache lookup as observed in Table~\ref{tab:trad_vs_learned}.
Techniques such as 
LSH which can deal with high dimensions cannot be used due to the high memory
overheads associated with storing the hidden layer outputs. High dimensionality also exacerbates lookup latencies.


\subsection{Towards Learned Caches}
To address the challenges mentioned above,
we propose adopting a {\em learning-based approach} for caching.
Instead of complementing each layer of the DNN with a traditional cache consisting of data entries,
our key idea is to use a {\em simple}
machine learning model (Figure~\ref{fig:cache_structure}a).
We train a model that mimics a cache lookup function and term this as a {\em learned cache}.
The interface exposed by a learned cache remains the same; i.e.,
the learned cache takes in a hidden layer output and delivers a prediction in the event of a cache hit.

Using a learned cache alleviates the issues associated with high
dimensionality.  First, ML models are effective at handling high dimension
inputs, thereby allaying concerns about the fidelity
of cache lookups.  Second, a learned cache does not actually store any data
but is instead a succinct representation encapsulated by the weights
of the model.  The size of a learned cache depends on
the model architecture and is independent of the number of items that
constitute the cache.  Thus, a learned cache can be used to
encapsulate a large number of data entries without additional
memory overheads.  Third, the latency of a learned cache lookup is
dictated by the cost of executing an ML model. By choosing an
appropriate model architecture, it is possible to keep lookup
latencies below an acceptable limit.  
Additionally, simple ML models with a small number of parameters
can be retrained quickly using recent data, thereby allowing
learned caches to effectively exploit temporal locality.


%

\section{Learned Cache Design}
\label{sec:learned_cache}





\begin{figure}[t!]
	\subfloat[][]{\includegraphics[scale=0.38]{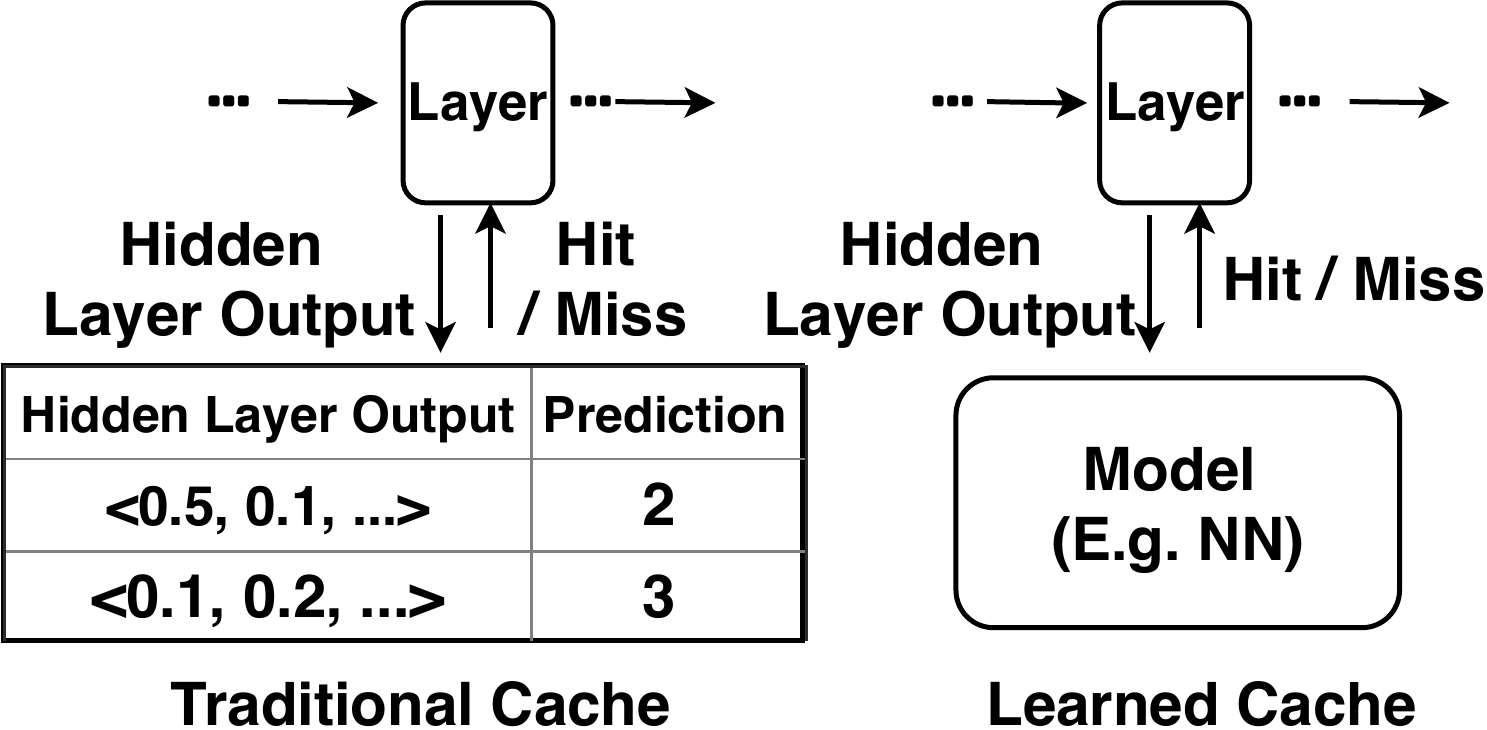}}
	\hspace{0.2cm}
	\subfloat[][]{\includegraphics[scale=0.25]{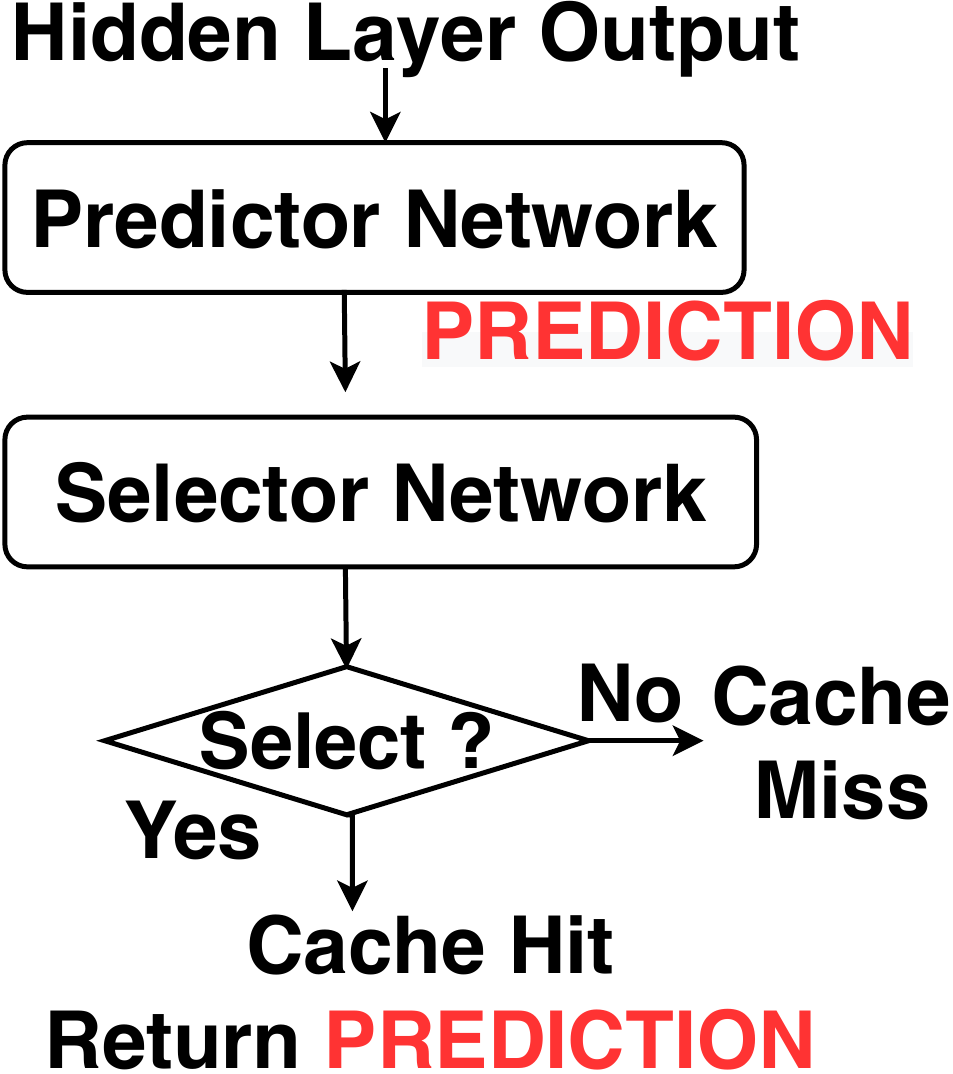}}
	\caption{\label{fig:cache_structure} \textbf{(a) Difference between traditional and learned cache. (b) Structure and operation of a learned cache.}}
\end{figure}

Given our goal of using ML models to represent caches, we observe
that ML models are directly amenable to generating a prediction;
e.g., for a classification problem we can train a model that predicts
the final class given the hidden layer output so far. However, to avoid
returning incorrect results we also need to determine if the
prediction is correct or not. Thus, we observe that we can decompose
the cache lookup operation into two sub-operations as shown in
Figure~\ref{fig:cache_structure}b:

\noindent(i) A {\em prediction operation} that computes an output
prediction given a hidden layer output.
\noindent(ii) A {\em selection operation} that determines whether the
prediction is a {\em cache hit} or not.

We model the two sub-operations using two different {\em neural
  networks}: a predictor network and a selector network. We 
use neural networks as they are universal function approximators and
provide a direct mechanism to approximate the base DNN model. Next, we
describe the neural network architectures that can be used to build
predictor and selector networks, and systems requirements that guide
their design.

\begin{table}[t]
  \centering
  \begin{scriptsize}
  \begin{tabular}{ c | c | c }
	  & \textbf{Selector $T=1$} & \textbf{Selector $T=0$} \\
    \hline
	  \textbf{Predictor Correct} & True Positive (TP) & False Negative (FN) \\
    \hline
	  \textbf{Predictor Wrong} & False Positive (FP) & True Negative (TN) \\
  \end{tabular}
  \end{scriptsize}
	\caption{\textbf{Confusion matrix for learned caches}}
  \label{tab:confusion_matrix}
\end{table}

\subsection{Predictor and Selector Networks}
\label{subsec:predictor_selector_design}
Using a neural network for predictors and selectors opens up a large design space. We start by
discussing systems requirements that guide our design. 

\noindent\textbf{Latency, Memory Usage.} As the selector and predictor
networks are run in sequence, the cache lookup latency is the sum of
their execution times. The memory usage is similarly the total amount
of memory required to store model parameters as well as the runtime
memory used while doing inference on these networks. Thus, we target
using a \emph{simple network with a limited number of hidden layers}
for low-latency and memory.

\noindent{\bf Hit-Rate, Accuracy.} We define hit-rate to be fraction of
lookups that yield a cache hit. Since learned caches are approximate,
we also need to account for the accuracy of the prediction, which we
define as the fraction of lookups that do not yield incorrect cached
predictions.  Learned caches need \emph{a high hit rate with
  accuracy above a certain threshold} (say 97\%) relative to base
model. Note that false negatives (FN;
Table~\ref{tab:confusion_matrix}), where the learned cached returned a
miss but the prediction was correct, do not hurt accuracy as we fall
back to the base DNN on a cache miss.

Given these requirements we next present various design choices and
how they meet the requirements above.

\begin{table}[t]
  \centering
  \begin{scriptsize}
  \begin{tabular}{ c | c | c  }
	  \textbf{Name} & \textbf{Architecture} & \textbf{FLOPS}\\
    \hline
	  \textbf{Fully-Connected} & Input $\xrightarrow[]{\text{FC}}$ Hidden Layer $\xrightarrow[]{\text{FC}} Output$ & 33.5M \\
	  FC(h) & (h) & \\
    \hline
	  \textbf{Pooling} & Input $\xrightarrow[]{\text{Pool}}$ Hidden Layer $\xrightarrow[]{\text{FC}}$ Output & 4.3M \\
     	  Pool(h)& (h) & \\
    \hline
	  \textbf{Convolution} & Input $\xrightarrow[]{\text{Conv}}$ Hidden Layer $\xrightarrow[]{\text{FC}}$ Output & 0.3M \\
	  Conv(k,s) & (k,s) & \\
  \end{tabular}
  \end{scriptsize}
	\caption{\textbf{Model architectures currently supported in \name{} for the predictor network. All of them
	use ReLU as the activation function.}}
  \label{tab:exploration_architectures}
\end{table}

\noindent\textbf{Predictor Network Design.}  We restrict predictor
networks to simple neural networks to satisfy latency, memory
requirements, and consider three possibilities
(Table~\ref{tab:exploration_architectures}): (a) a fully connected
architecture (FC) that can learn non-linear combinations of inputs and
project it onto the output dimension, (b) a convolution based
architecture that limits non-linear combinations to local regions of
the input and (c) a pooling-based architecture that reduces the
spatial size of the inputs before projecting them onto the output
dimension. We choose these three options as they represent different
points in terms of FLOPs required
(Table~\ref{tab:exploration_architectures}). Our system is extensible
and can include other model architectures.

\noindent\textbf{Selector Network Design.} While the predictor network attempts to mimic the base
DNN, the selector network has a simpler role as it only needs to perform a binary classification of whether we have a cache hit or miss. We find that using a simple neural network with one hidden layer
that projects the output of the predictor network onto an output layer
that enables a binary decision to be made is sufficient and other architectures do not provide any
significant benefits.



\noindent\textbf{Training Learned Caches.}
Consider a base DNN model with $N$ hidden layers $L_1$, .., $L_N$
and that we are given $M$ input samples from a validation dataset, $X_1$, ..,
$X_M$, to construct learned caches.
To train a predictor network, we first run a forward pass of the base DNN over 
the $M$ input samples. During the forward pass for each input $X_j$, we collect the hidden layer output at layer $i$ 
as $HO_{i,j}$ and also record the final prediction of the base DNN as $Y_{j}$, where $Y_{j}$ is a
vector representing the distribution of class probabilities for classification ($\argmax_{j} Y_{j}$ is
predicted class).
The collection of data ${<HO_{i,1}, Y_{1}>, .., <HO_{i,M}, Y_{M}>}$
is then used to train a predictor network at layer $i$ of the DNN.
Given that the predictor network is trying to mimic the behavior of the rest of the base DNN,
we borrow insights from distillation~\cite{distillation} and use a loss function that takes into account the true labels and also the distribution of class probabilities.

A selector network at layer $i$ is a function $T=\sigma_{i}(PR)$, where $PR$ is the class distribution prediction from the predictor network and $T$ is a binary decision.
To train a selector network at layer $i$, we first construct ground truth labels ($G$) in the
following manner:
\[
  G_{i,j} = \begin{cases}
    1, & \text{if } \argmax\limits_{j} PR_{i,j} = \argmax\limits_{j} Y_{j}\\
    0, & \text{otherwise} 
  \end{cases}
\]
The collection of data ${<PR_{i,1}, G_{i,1}>, .., <PR_{i,M}, G_{i,M}>}$
can be used to train a selector network at layer $i$. A key function of the selector network is to reduce the number of
false positives (FP) so as to achieve high accuracy. To this end, we employ a custom cross-entropy loss
function that levies a higher loss penalty for FPs in comparison for FNs, since FNs do not impact accuracy as discussed
earlier.

\begin{table}[t]
	\begin{scriptsize}
  \begin{tabular}{ c | c | c | c | c | c | c }
	  \textbf{Block} & \textbf{Arch.} & \textbf{Accuracy} & \textbf{Hit} & \textbf{CPU} & \textbf{GPU} & \textbf{Memory} \\
	  & & & \textbf{Rate} & \textbf{Latency} & \textbf{Latency} & \textbf{Cost} \\
    \hline
	  3 & FC(1024) & 97.3\% & 38.8\% & 6.08 ms & 0.43 ms & 268 MB \\
    \hline
	  3 & Pool(8192) & 96.7\% & 34.1\% & 1.32 ms & 0.53 ms & 33 MB \\
    \hline
	  3 & Conv(3,1) & 96.2\% & 20.4\% & 1.66 ms & 0.48 ms & 2 MB \\
    \hline
	  6 & FC(1024) & 99.5\% & 62.9\% & 2.94 ms & 0.47 ms & 134 MB \\
    \hline
	  6 & Pool(8192) & 96.2\% & 54.4\% & 0.64 ms & 0.5 ms & 33 MB \\
    \hline
	  6 & Conv(3,1) & 99.3\% & 49.4\% & 0.68 ms & 0.49 ms & 0.8 MB \\
  \end{tabular}
		\end{scriptsize}
	\caption{\textbf{Trade-off space exposed by different model architectures for predictor networks
	at the end of the $3^{rd}$ and $6^{th}$ ResNet-18 block.}}
  \label{tab:exploration_tradeoff}
\end{table}



\subsection{Predictor Design Trade-offs}
We now discuss the trade-offs involved in choosing the appropriate
architecture for predictor networks by considering three learned cache
variants for the $3^{rd}$ and $6^{th}$ blocks in ResNet-18 as
illustrated in Table~\ref{tab:exploration_tradeoff}. 
We train all variants per the procedure described earlier (\secref{subsec:predictor_selector_design}).

\noindent{\bf Hit-rate vs. System resources. }
From Table~\ref{tab:exploration_tradeoff}, we notice that
architectures that are more computationally expensive take up more
systems resources (have larger lookup latencies and memory cost), but
offer greater hit rate and hence greater reduction in the average
end-to-end inference latency (see FC(1024) vs. Pool(8192) for both
block 3 and 6). 

\noindent{\bf Hardware dependent behavior. }
From Table~\ref{tab:exploration_tradeoff}, we notice that the nature of the trade-off 
depends on the target hardware due to differences in the underlying lookup latencies.
For instance, we notice that FC(1024) has both a better hit rate and lookup latency
on GPU compared to Pool(8192), while FC(1024) has higher lookup latency on CPU.
This is because the fully-connected layer requires a large dense matrix multiplication and 
this operation can be effectively parallelized across many thread blocks available on a GPU.

\noindent{\bf Role of base DNN layer. } From
Table~\ref{tab:exploration_tradeoff}, we also see that the dynamics of the
trade-off space varies across layers of the base DNN.  We see that all
architectures have a higher hit rate at block 6 compared to block
3 indicating that caching is easier as we get closer to the output layer. 
Additionally, the
lookup latencies and memory costs also reduce at block 6 since dimensionality of
the hidden layer output reduces as we go deeper in the base DNN. 

Collectively the three observations indicate that the appropriate choice of network architecture
depends on the systems resources available, the hardware being used, and the base-DNN layer being
considered. This motivates the need for a scheme that can reason about various learned cache
variants at each layer of the DNN and collectively optimize the system for end-to-end latency
benefits.  We next present a general approach to address this challenge of composing learned
caches for end-to-end benefits.

\section{Composing Learned Caches}
\label{sec:static-construction}
\label{subsec:composition_phase}




The trade-offs presented in the previous section indicate that a
number of learned cache variants could be applicable for each layer of
the DNN. To allow for systematic exploration, our system first runs an
{\em exploration phase}, where for each variant corresponding to a
base DNN layer, we compute the expected hit rate, accuracy,
lookup latency, and memory cost.

\begin{figure}[t]
        \centering
        \includegraphics[width=0.7\columnwidth]{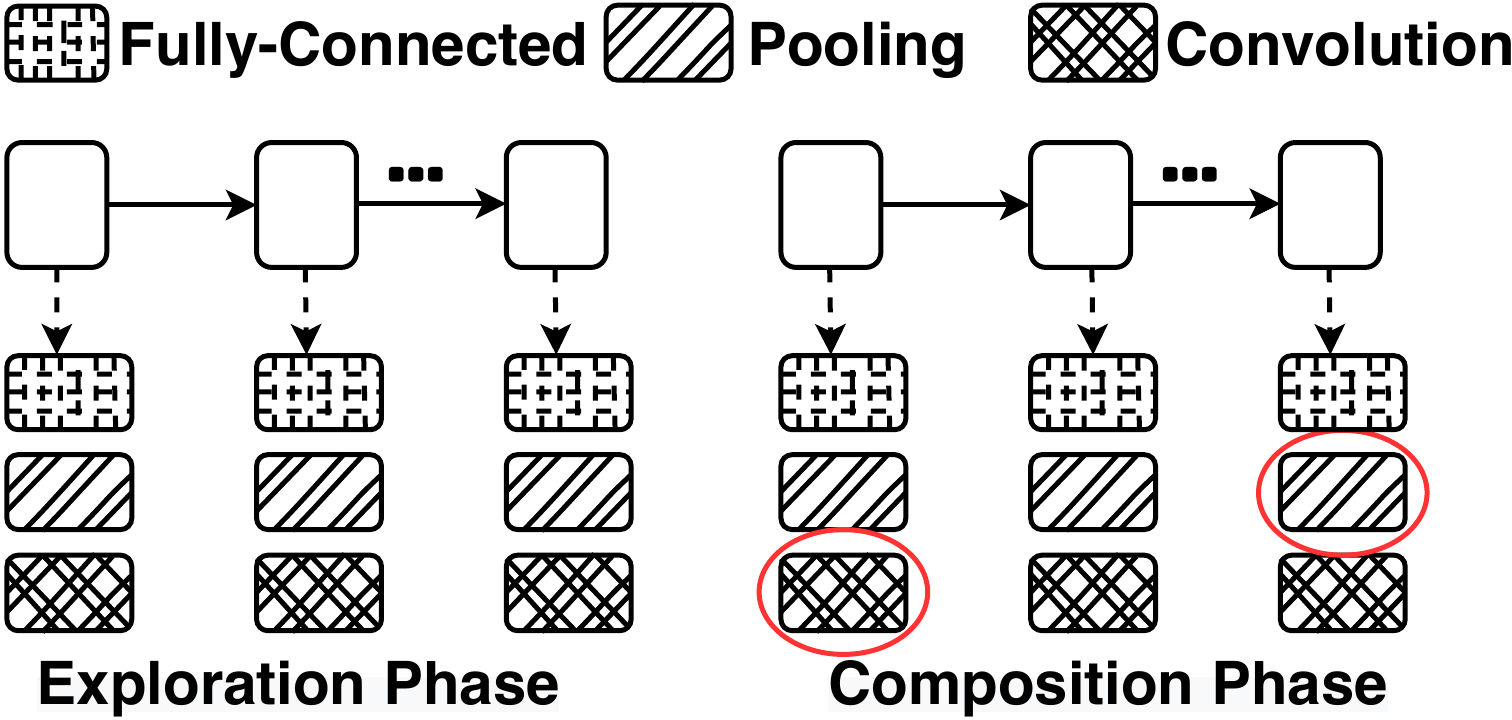}
        \caption{\textbf{(a) Exploration phase - \name{} considers multiple cache variants at each layer of the DNN
        (b) Composition phase - \name{} chooses a subset of variants to complement the base DNN during inference.}}
        \label{fig:exploration_composition_phase}
\end{figure}

Then, we use these as inputs to select a subset of these learned cache
variants for inference. \name{} accomplishes this in a {\em
  composition phase} by formulating it as an optimization problem
(Figure~\ref{fig:exploration_composition_phase}). 
The above two phases are one-time operations performed
when inference service owners upload trained DNNs.
Once the learned caches are deployed,
\name{} exploits temporal locality by fine-tuning or retraining
the chosen cache variants in an online manner (\secref{subsec:overall_system_design}).




Consider the following toy example that illustrates the problem
underlying the final choice of the optimal learned caches to use
during inference.  Suppose we have an aggregate memory budget of 564
MB available for learned caches. 
Table~\ref{tab:exploration_tradeoff} shows that a
potential choice can be to greedily select architectures
at earlier layers that offer high hit-rates.
This would mean selecting FC(1024) at block 3 which yields
cache hits for 38.75\% of requests. However,
it uses up all of the available memory budget and hence
the remaining 62.25\% of requests would incur the
latency of running the entire base DNN.
The latency for the 38.75\% cached requests would
be the time to compute the first three blocks of ResNet
plus the lookup latency (6.08 ms).

Another potential choice can be to select Pool(8192) at block 3 and
FC(1024) at block 6. Combined, they have a memory cost of 264 MB,
which is within the memory budget. We would get cache hits for 34.1\%
of requests at block 3. Interestingly, due to a lower lookup latency,
the latency for these requests would be {\em lesser} than the latency
of requests with cache hits at block 3 in the first option.  Further,
an additional 28.8\% of requests would get cache hits at block 6, and
the remaining 37.1\% of requests would incur the latency of running
the entire base DNN.  The second option is preferable since it can
significantly reduce the average latency of requests, even though a
small additional fraction of requests achieve higher latency relative
to the first option. Composing an optimal set of caches thus requires
a global view of the trade-off space across layers of the DNN.

\noindent {\bf Goal: } The goal of the composition phase is to jointly
select a global set of learned caches that {\em minimize the expected
  average latency}, while ensuring minimal degradation in
accuracy and meeting compute and memory constraints.

We can formulate this problem as a mix-integer quadratic program 
but we find that the formulation is intractable while handling deep networks. 
We present the optimization problem in full detail in Appendix~\ref{subsec:appendix_optimization}. \name{} uses a relaxation of the formulation whose key details we explain below.

Let us assume that we have a DNN with $N$ layers and $K$ cache
variants at each layer.  We index each variant by its layer $i$ and
variant number $j$.  From the exploration phase, we obtain the
following metrics for each variant - Hit Rate ($H_{i,j}$), Accuracy
($A_{i,j}$), Lookup Latency ($T_{i,j}$), and Memory Cost
($M_{i,j}$). Additionally, we profile the latency for the computation
of each layer ($L_i$). Binary variable $b_{i,j}$
indicates if learned cache variant $j$ at layer $i$ is chosen.
$b_{i,j}=1$ means that the variant is selected.

We use a three-step approach to simplify the composition problem
formulation:

\noindent{\bf Step 1. Accuracy Filter: } We do not consider variants 
whose accuracy $A_{i,j}$ is below a minimum accuracy threshold $A$.

\noindent{\bf Step 2. Score Computation: }
Motivated by the example from before, we consider two factors in determining
the importance of a particular variant:
(i) the hit rate of the variant and 
(ii) the {\em latency gain} obtained by using the model variant in the event of a cache hit.
We compute latency gain (LG) for a learned cache variant
as the ratio of running time for the entire DNN to the running time
assuming that a cache hit is obtained at the given learned cache variant.

\begin{equation}
     LG_{i,j} = \sum_{k=1}^{N}L_{k}/(\sum_{k=1}^{i}L_{i} + T_{i,j})
\end{equation}

We prefer higher hit rates and higher latency gains. However, these are
fundamentally at odds with each other since higher latency gains are
obtained using variants at earlier layers of the base DNN where the
hit rates would be lower, and vice versa.  To balance these two
factors, we compute a score ($S$ - higher is better) that captures
the benefit of using a variant:

\begin{equation}
\label{eqn:model_score}
	S_{i,j} = \alpha.(H_{i,j}) + (1 - \alpha).(LG_{i,j})
\end{equation} 

\noindent where $\alpha$ is a knob that lies in [0,1] and controls
the relative importance of hit rate and latency gain.

\noindent{\bf Step 3. Resource Constraints: } We constrain the total
memory occupied by chosen variants to be within a memory budget
$M$. For computation, to avoid latency inflation, we wish to run the learned caches
asynchronously while the computation of the base DNN proceeds. To minimize the amount
of resources required, we specify a computational constraint that we can
atmost perform one cache lookup at a given point of time 
\footnote{Our profiling on GPUs suggests that running more than two
concurrent models using MPS or CUDA streams imposes a 10-15\% overhead.}.



Finally, the objective of our formulation is to {\em maximize} the 
sum of scores for chosen variants:

\begin{equation}
	\text{\textbf{max.}} \sum_{i=1}^{N}\sum_{j=1}^{K}b_{i,j}.S_{i,j}
\end{equation}

The computed values of $b_{i,j}$ then determine which learned cache
variants should be used along with the base DNN during inference.
We next describe how the above composition phase is integrated into the end-to-end query lifecycle
and present the design of our system \name{}.

\section{\name{} System Design}
\label{sec:system-design}




We design \name{}, an end-to-end prediction serving system that
leverages learned caches to speed up DNN inference serving
(Figure~\ref{fig:system-design}). Users interact with \name{} by
issuing a query along with a latency SLO (query completion
deadline). Similar to~\cite{nexus}, \name{} determines the dataflow
DAG of DNNs that need to be executed for a query. Like prior work~\cite{nexus}, \name{}
considers simple DAGs with chains or fork-join dependencies, that
have a single input and a single output prediction.
In addition, \name{}
allows the inference service owner to specify an array of possible DNN
models with different accuracies that can be used at each node of the
dataflow DAG.


\begin{table}[t]
  \centering
  \begin{scriptsize}
  \begin{tabular}{ l | c | c | c }
	  \textbf{Module} & \textbf{DNN} & \textbf{Latency} & \textbf{Accuracy} \\
    \hline
                               & ResNet-18 & 27.36 ms & 91.1\% \\
	  \textbf{Obj. Detect} & ResNet-34 & 41.05 ms & 92.9\% \\
	                       & ResNet-50 & 54.5 ms & 94.1\% \\
    \hline
	                       & SE-LResNet9E-IR & 17.38 ms & 95.5\% \\
			       & SE-LResNet18E-IR~\cite{arcface} & 36.75 ms & 97.6\% \\
          \textbf{Face} & SE-LResNet50E-IR & 58.34 ms & 98.1\% \\
                               & SE-LResNet101E-IR & 110.32 ms & 99.1\% \\
    \hline
	                       & ResNet-9 & 16.14 ms & 90.2\% \\
	                       & ResNet-18 & 23.68 ms & 91.8\% \\
          \textbf{Vehicle} & ResNet-50 & 54.12 ms & 92.6\% \\
                               & ResNet-101 & 111.42 ms & 93.4\% \\
  \end{tabular}
  \end{scriptsize}
	\caption{\textbf{DNN model options at each node of traffic analysis application.}}
  \label{tab:query_replanning_models}
\end{table}

\noindent {\bf Example: } Consider the traffic analysis application as
shown in Figure~\ref{fig:motivation}a which represents a DAG of models
that need to be executed. 
As shown in
Table~\ref{tab:query_replanning_models}, the inference service
provider can specify an array of possible DNNs that can be used for
each node of the DAG. These options can vary from cheap DNNs that have lower
accuracy to more expensive DNNs that have higher accuracy.

\subsection{Query Planner}
\label{subsec:scheduler}
Given a query, \name{}'s query planner formulates a query evaluation
plan (QEP) that captures what DNN model to pick for
execution at each node of the dataflow DAG, while ensuring that the
latency SLO for the query is met.

\noindent{\bf Example: }
Consider a query for the traffic analysis application (Figure~\ref{fig:motivation}a)
with a latency SLO of 80 ms.
First, the planner needs to split the available latency SLO budget amongst
the different nodes in the dataflow DAG.
There can be multiple policies~\cite{grandslam,nexus,archipelago} to compute these {\em partial
budgets} (\secref{subsubsec:split_policy}).  Let us assume a simple
policy that divides the latency budget equally among all nodes: this
will allocate 40 ms for object detection and 40 ms for face/vehicle
recognition.

Next, the planner needs to choose a DNN with maximum accuracy that can be executed within the
partial latency budget. For object detection, state-of-the-art static planners~\cite{nexus,infaas} would choose ResNet-18, and
SE-LResNet18E-IR and ResNet-18 for face and vehicle recognition respectively.
This constitutes the QEP for a latency SLO of 80 ms.

\noindent{\bf Learned caches enable incremental replanning: }
Consider a scenario where learned caches are used and, because of cache hit, say
object detection can execute in 20 ms instead of 27 ms. 
This leaves 60 ms instead of 53 ms for face/vehicle recognition.
With a new latency budget of 60 ms, we can now use {\em SE-LResNet50E-IR} and {\em ResNet-50}
for face and vehicle recognition respectively, without violating the latency SLO.
These options have higher accuracy than the originally chosen ones.
Thus, learned caches enable {\em incremental replanning} of
the yet-to-be-traversed DAG nodes, leading to higher end-to-end
accuracy for the query.

\floatname{algorithm}{Pseudocode}
\begin{algorithm}[t!]
\begin{scriptsize}
\begin{algorithmic}[1]

\State {$\myvec{PLB}$ \Comment{Map holding partial latency budgets}} \label{alg:partial-budgets}
\Statex
\State {$\triangleright$ Prepare a QEP for DAG D with LatencySLO L and begin execution}
\Procedure{OnReceiveQuery}{DAG D, LatencySLO L}
	\State $\myvec{PLB}$ = \textsc{ComputePartialBudgetWithPolicy(D,L)}
	\State \textsc{DNN} = \textsc{PickBestModel(D.root, $\myvec{PLB}$)} \label{alg:best-model}
	\State \textsc{ExecuteDNN(DNN)}
\EndProcedure
\Statex
\State {$\triangleright$ Called when a DNN D finishes execution}
\Procedure{OnExecutionComplete}{DNN D} \label{alg:execution-complete}
        \State {$\triangleright$ Redistribute the saved latency amongst downstream nodes}
	\State \textsc{SavedLatency} = \textsc{D.BaseModelLatency} - $\myvec{PLB}$\textsc{(D)}
        \State $\myvec{PLB}$ = \textsc{RedistributeLatencyWithPolicy($\myvec{PLB}$,SavedLatency)}
	\ForAll{\textsc{node} $\in D.downstreamNodes$}
	    \State \textsc{DNN} = \textsc{PickBestModel(D.root, $\myvec{PLB}$)} 
	    \State \textsc{ExecuteDNN(DNN)}
	\EndFor
\EndProcedure
\end{algorithmic}
\end{scriptsize}
\caption{Query Replanning Algorithm}
\label{alg:replanning-algorithm}
\end{algorithm}

\subsection{Incremental Replanning Algorithm}

\name{}'s query planner realizes incremental replanning by deferring
the selection of a particular DNN at a node to when the node needs to be
scheduled for execution (Algorithm~\ref{alg:replanning-algorithm}).

\noindent (i) On receiving a query, the planner allocates partial latency budgets to each node of
the 
DAG according to a policy to compute partial latency budgets (described below)
and begins query execution at the root of the DAG. 
(ii) When a node needs to be executed, the planner picks
the highest accuracy DNN at that node that can be executed within the partial
latency budget (Line~\ref{alg:best-model} in Algorithm~\ref{alg:replanning-algorithm}).
(iii) When a DNN finishes execution, the planner computes the
{\em saved} latency by subtracting the observed execution time from the allocated partial latency budget.
This saved latency is redistributed amongst downstream nodes according to
the partial latency budget allocation policy.

\subsubsection {Computing Partial Latency Budgets}
\label{subsubsec:split_policy}
The query planner considers all possible critical paths from the input
to the output in the DAG to compute partial latency budgets.  First,
\name{} computes a partial latency budget for {\em each} node by
considering one DAG-input-to-DAG-output path at a time.  Borrowing from
planners that support the execution of a DAG for prediction serving
tasks~\cite{nexus,grandslam}, \name{} currently supports two policies
to compute partial latency budgets.

\noindent {\bf Equal Split.}
This splits the latency SLO equally amongst each node
in a path.
If $L$ is the latency SLO and there are $N$ nodes in the  path,
then each node receives a budget of $\frac{L}{N}$.

\noindent {\bf Proportional Split. }
This policy splits the latency SLO among nodes
in a critical path in proportion to the {\em maximum} execution time
of any DNN option in that node.
If there are $N$ nodes in an input-output path and $C_{i}$
is the maximum execution time among options at node $i$,
then the partial latency budget for node $i$ is computed as -

\begin{equation}
	L_{i} = \frac{C_{i}}{\sum_{j=1}^{N}C_{j}}*L
\end{equation}

Next, \name{} computes the final partial latency budget at a node as 
the {\em minimum} partial latency budget available 
for the node from all possible  paths, so as to meet
the latency SLO for any  path that the request might go through.

\name{} allows any policy to be plugged into the incremental replanning
framework as shown in Algorithm~\ref{alg:replanning-algorithm}, which
we also evaluate in \secref{subsec:eval_replanning}.

\subsection{Retraining Learned Caches}
\label{subsec:cache_update}
\name{} retrains learned caches 
to exploit the temporal
locality inherently present in online workloads (\secref{subsubsec:opportunities_motivation}): retraining achieves the effect of fine-tuning the caches, 
leading to better hit rates. 
We sample input queries that users issue to \name{} and
maintain a window of samples obtained over time.
Then, we periodically retrain the predictor
and selector network for each learned cache variant chosen in the composition phase 
using a mix of data from the window of samples and
the original validation dataset provided by the user~\cite{catastrophicforgetting}.
While picking sampled inputs, we weigh recent samples more heavily~\cite{learninginsitu}.
We obtain the training data needed for retraining by running a forward
pass over the base DNN on the chosen sampled inputs
as described earlier in \secref{subsec:predictor_selector_design}.

We have two parameters to control retraining: the number of samples stored and the frequency at
which retraining is triggered.  Our evaluation shows that using \textasciitilde 20\% samples we can
perform retraining within 500ms, owing to the simplicity of the predictor/selector networks' designs, 
and that a retraining period of 15 minutes
works well for realistic workloads.
We present more detailed results in
Section~\ref{subsubsec:online-insights}.



\subsection{Overall System Design}
\label{subsec:overall_system_design}

\begin{figure}[t]
  \centering
   \includegraphics[width=0.42\textwidth]{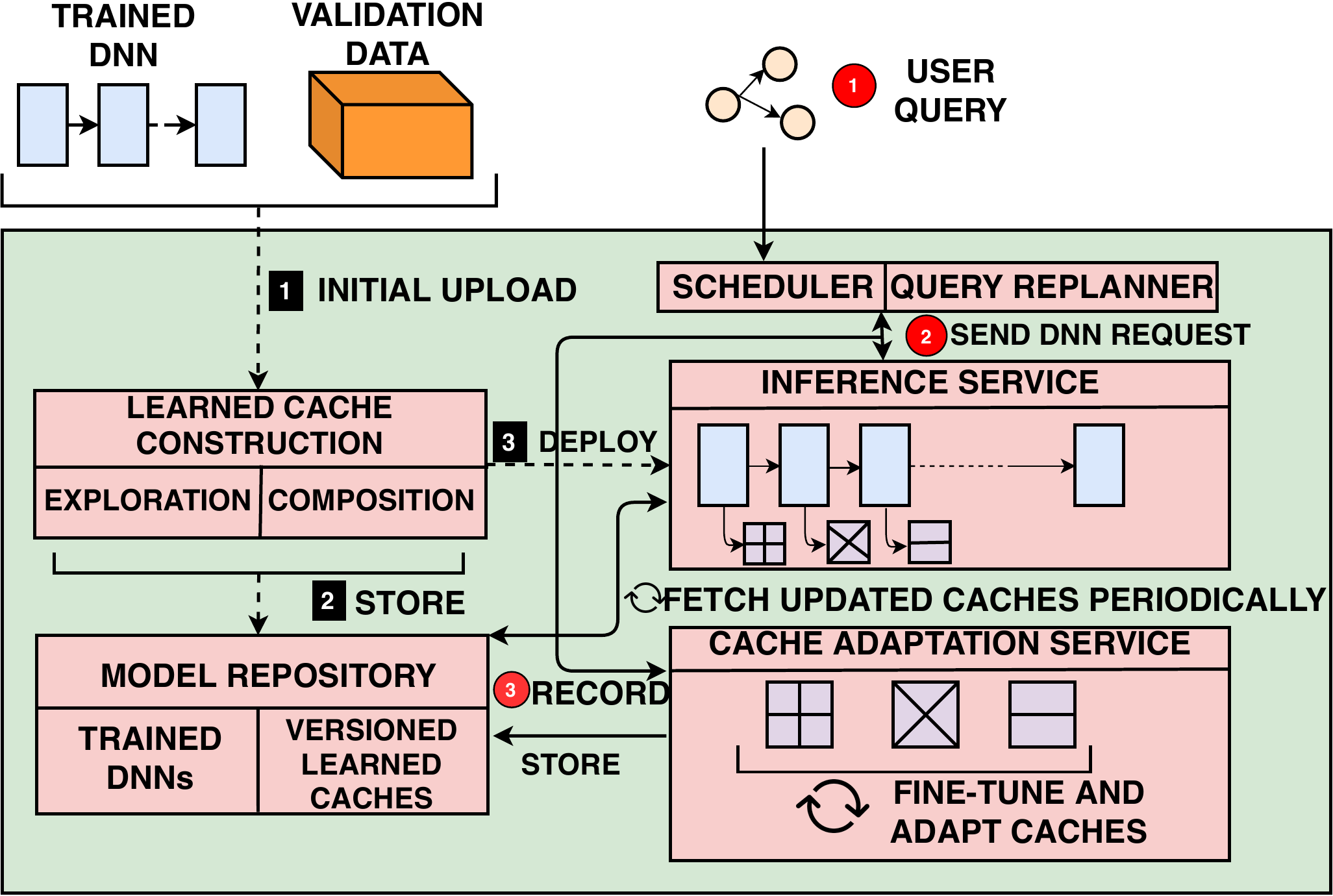}
        \caption{\textbf{End-to-end system design of \name{}. Dashed lines show deployment phase; solid lines show online phase.}}
    \label{fig:system-design}
\end{figure}

As outlined in Figure~\ref{fig:system-design},
users interact with \name{} in two ways:

\noindent (i) {\em Deployment Phase}:
Inference service owners 
begin DNN deployment by uploading
a trained model and its validation dataset.
Given this, \name{} constructs an initial set of learned caches by exploring multiple variants
and composing an optimal set of learned caches (\secref{subsec:composition_phase}).
The base DNN and along with learned caches are then deployed to the inference service.

\noindent (ii) {\em Online Phase}: 
Queries issued to \name{} go through a front-end
query planner that formulates a QEP for the query and handles
its execution. The cache adaptation service records samples
of the query inputs from the inference service 
and periodically retrains the learned
caches. 
The inference service periodically pulls new versions of learned caches. 

\section{Implementation}
\label{sec:implementation}

We developed a prototype of \name{} that implements the above
system design. Each of the services (query scheduler, inference service,
cache adaptation service) are written as Apache Thrift~\cite{thrift}
services that communicate using RPCs.
Our prototype currently supports DNNs written in PyTorch~\cite{pytorch}
and supports inference on CPUs and GPUs.

For the exploration of learned cache variants we load the PyTorch model
from the checkpoint and add hooks to save hidden layer outputs by traversing the list
of \texttt{Modules} in the model.
The composition phase uses Gurobi~\cite{gurobi} to compute the optimal
set of learned caches after exploration.


At runtime, the query scheduler accepts queries using a REST API. 
Each DNN and its learned caches reside on the same dedicated instance.
We use similar hooks as in the exploration phase so that cache lookups can be performed
during inference. The learned caches lookups are done asynchronous to the execution of the
base DNN.
On CPUs, we achieve this by running the base DNN execution and learned cache lookup
in different processes. Each process is pinned to a dedicated set of CPU cores that do not
overlap and the intermediate layer output is shared between processes using pipes.
This ensures isolation between the computations and guarantees that the execution time does not
exceed that of the vanilla base DNN. 
On GPUs, we overlap computations using CUDA streams~\cite{cudastreams}. The base DNN computation
proceeds on the default stream while the learned cache computations are issued on a different
stream.
Our cluster deployment design of \name{} is similar to existing prediction serving
systems~\cite{clipper,nexus,paritymodels} and thus \name{} inherits their horizontal scaling and fault-tolerance properties.


\section{Evaluation}
\label{sec:evaluation}

We use our prototype implementation of \name{}
to evaluate if \name{} can deliver on the opportunities
identified in \secref{subsubsec:opportunities_motivation}.
Our evaluation shows the following:

\noindent (i) Across a range of datasets and state-of-the-art DNNs, \name{} maintains high accuracy
     and offers 
     up to 1.95$\times$ improvement in average latency compared to the base DNN (\secref{subsec:eval_late_binding}).

\noindent (ii) \name{} exploits temporal locality in real-world videos, giving up to 7.69$\times$ improvement in average latency (\secref{subsec:eval_online}).

\noindent (iii) Incremental replanning, when using a DAG of models, helps overcome the fundamental latency-accuracy trade-off
	    by simultaneously improving accuracy, with respect to the ground truth, by \textasciitilde 1\%
      and prediction latency by 1.26$\times$
      (\secref{subsec:eval_replanning}).


\noindent {\bf Testbed Setup: }
We deployed \name{} on a heterogeneous cluster
comprising of 2 p3.2xlarge GPU instances and 12 c5.4xlarge instances on AWS~\cite{awsec2} thus measuring
how \name{} helps on both CPU and GPUs.
Each GPU instance has 1 NVIDIA Tesla V100 GPU.
For GPU experiments, we use 1 GPU instance for the inference service
and the other for the cache adaptation service.
For CPU inference, we use the c5.4xlarge instance and
allocate 8 vCPUs each for the base model and learned cache computation.
Each service (query scheduler, inference service, and cache adaptation service) 
runs as a daemon.
Since the construction of initial caches is a one-time operation,
we use multiple AWS GPU spot instances and CloudLab~\cite{cloudlab}
for exploring multiple cache variants.

\noindent {\bf Default Parameters: }
We consider six learned cache variants -
FC(1024), FC(512), Pool(8192), Pool(4096), Conv(3,1), Conv(5,2).
For the composition phase, we choose the target accuracy $A$
to be 97\%.
We evaluate higher accuracy targets in~\secref{subsubsec:other-workloads}.
We use a batch size of 1 in all of our experiments and
evaluate larger batch sizes in \secref{subsec:eval_online}.

\noindent {\bf Learned Cache Training Methodology: }
We use the validation data from each dataset to train the 
predictor and selector networks.
The initial cache construction uses 80\% of the training data,
sampled at random for training each predictor/selector network.
The remaining 20\% is used for obtaining the metrics 
needed by the composition phase (\secref{sec:static-construction}).

\subsection{Late Binding Benefits}
\label{subsec:eval_late_binding}
We first evaluate the latency benefits of using learned caches
during inference. 
To study this aspect in isolation, we disable the cache
adaptation service for these experiments.

\noindent {\bf Workload: }
Our evaluation considers popular image classification
(CIFAR-10 and CIFAR-100~\cite{cifar10-data}) and speech recognition
(Google Commands~\cite{speech-data}) tasks.
We consider 4 different base DNN architectures: VGG-16~\cite{vgg}, ResNet-18,
ResNet-50, ResNet-152~\cite{resnet}. These represent state-of-the-art DNNs
that have different number of layers.
We measure accuracy, latency using the respective test datasets.

\noindent {\bf Baselines: }
We compare against existing techniques (\secref{subsubsec:current_solutions}):
(i) {\em Quantization} ($n$-bit).
(ii) {\em Model distillation}.
(iii) {\em Model pruning} ($x$\%).
We compare against an approach that employs
a {\em cascade} of specialized models having varying number
of layers, similar to \cite{noscope,cascades}.
We assume that models in a cascade are run serially
to enable a fair comparison with \name{} which uses
only a single hardware resource for inference.


\subsubsection{Learned caches with ResNet-50}
\label{subsubsec:res50_results}

\begin{table}[t]
  \centering
  \begin{scriptsize}
  \begin{tabular}{ c | c | c}
	  \textbf{Model (Dataset)} & \textbf{CPU Latency Gain} & \textbf{GPU Latency Gain}\\
    \hline
	  ResNet-50 (CIFAR-10) & 1.95$\times$ & 1.63$\times$ \\
    \hline
	  ResNet-18 (CIFAR-10) & 1.72$\times$ & 1.28$\times$ \\
    \hline
	  ResNet-152 (CIFAR-100) & 1.24$\times$ & 1.21$\times$ \\
    \hline
	  VGG-16 (Google Commands) & 1.96$\times$ & 1.54$\times$ \\
  \end{tabular}
  \end{scriptsize}
	\caption{\textbf{Overall latency benefits of \name{} over base models. Accuracies meet the 97\% target.}}
  \label{tab:other_cpu_macro_results}
\end{table}

\begin{figure}[t!]
	\subfloat[][Latency distribution]{\includegraphics[width=0.5\columnwidth]{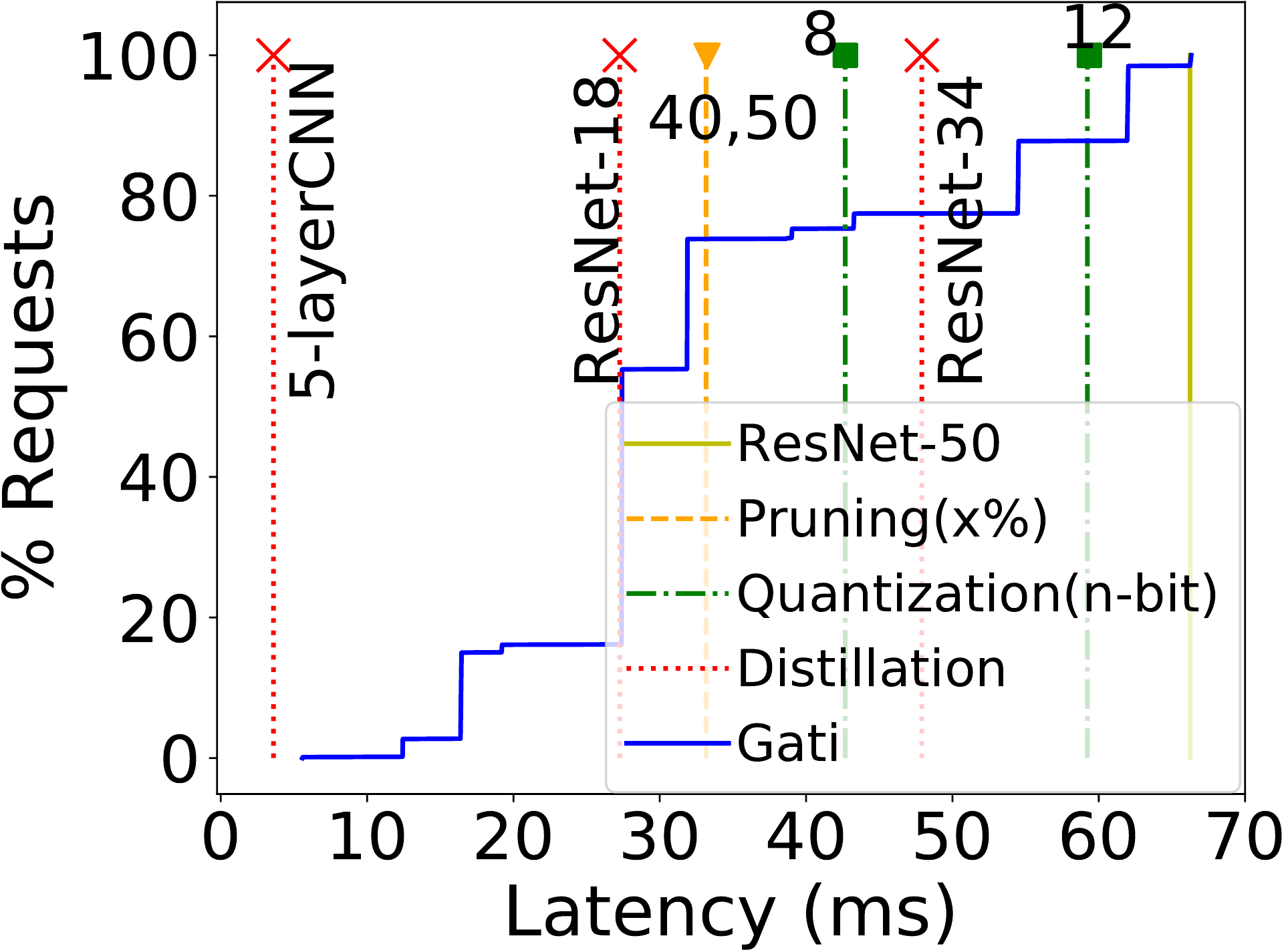}}
	\subfloat[][Latency vs. accuracy]{\includegraphics[width=0.5\columnwidth]{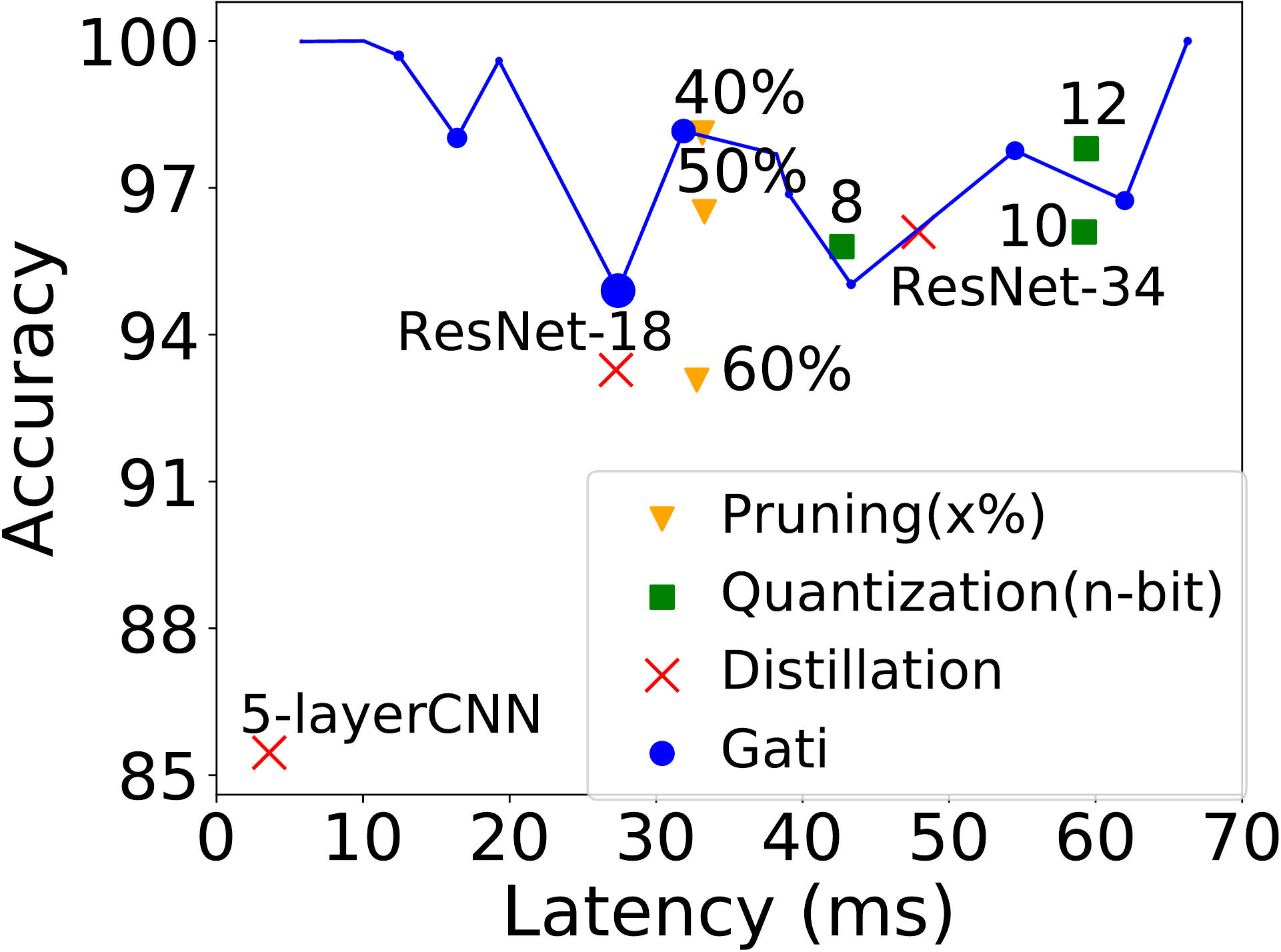}}
	\caption{\label{fig:res50-macro-results-cpu} \textbf{[CPU inference] Benefits of late-binding for 
	ResNet-50 on CIFAR-10 dataset. In (b),
	Size of each \name{} marker is proportional to \% of requests at the corresponding latency point.}}
\end{figure}

We first consider Resnet-50 with CIFAR-10 data as the base DNN and use CPUs for inference. 
From Figure~\ref{fig:res50-macro-results-cpu}a, we observe that
\name{} exhibits an average latency of \textasciitilde {34 ms}, which is
{\bf 1.95$\times$} lower than the latency of running the entire DNN.
Figure~\ref{fig:res50-macro-results-cpu}(b) shows that \name{} exhibits
a spectrum of latencies with an accuracy of {\bf 96.97\%} with respect
to the base ResNet-50 model, which is very close to the accuracy target of 97\%.
The learned caches occupy 1277.5 MB memory.

\noindent {\bf Against quantization: }
\name{} outperforms both 8 and 10-bit quantization, improving average latencies by a factor
of 1.25$\times$ and 1.74$\times$ respectively while also improving accuracy by 0.57\% and 0.97\% respectively.
Against 12-bit quantization, \name{}'s accuracy is \textasciitilde1\% lower but it provides
lower latency for \textasciitilde87\% of requests.

\noindent {\bf Against distillation: }
\name{} outperforms ResNet-50 distilled to ResNet-34 with an 
average latency improvement of 1.41$\times$
and an accuracy improvement of 0.57\%. 
ResNet-50 distilled to ResNet-18 and a simple 5-layer CNN
have better average latency than \name{} but incur significant trade-off on accuracy.

\noindent {\bf Against pruning: }
\name{} has better accuracies than pruning at 50\%, 60\%, and 70\%, while also improving average
latency by a factor of 1.39$\times$-1.42$\times$. Pruning at 40\% has higher accuracy, but
\name{} offers better latency for 70\% of the requests.

\begin{figure}[t!]
	\centering
	\subfloat[][]{\includegraphics[width=0.45\columnwidth]{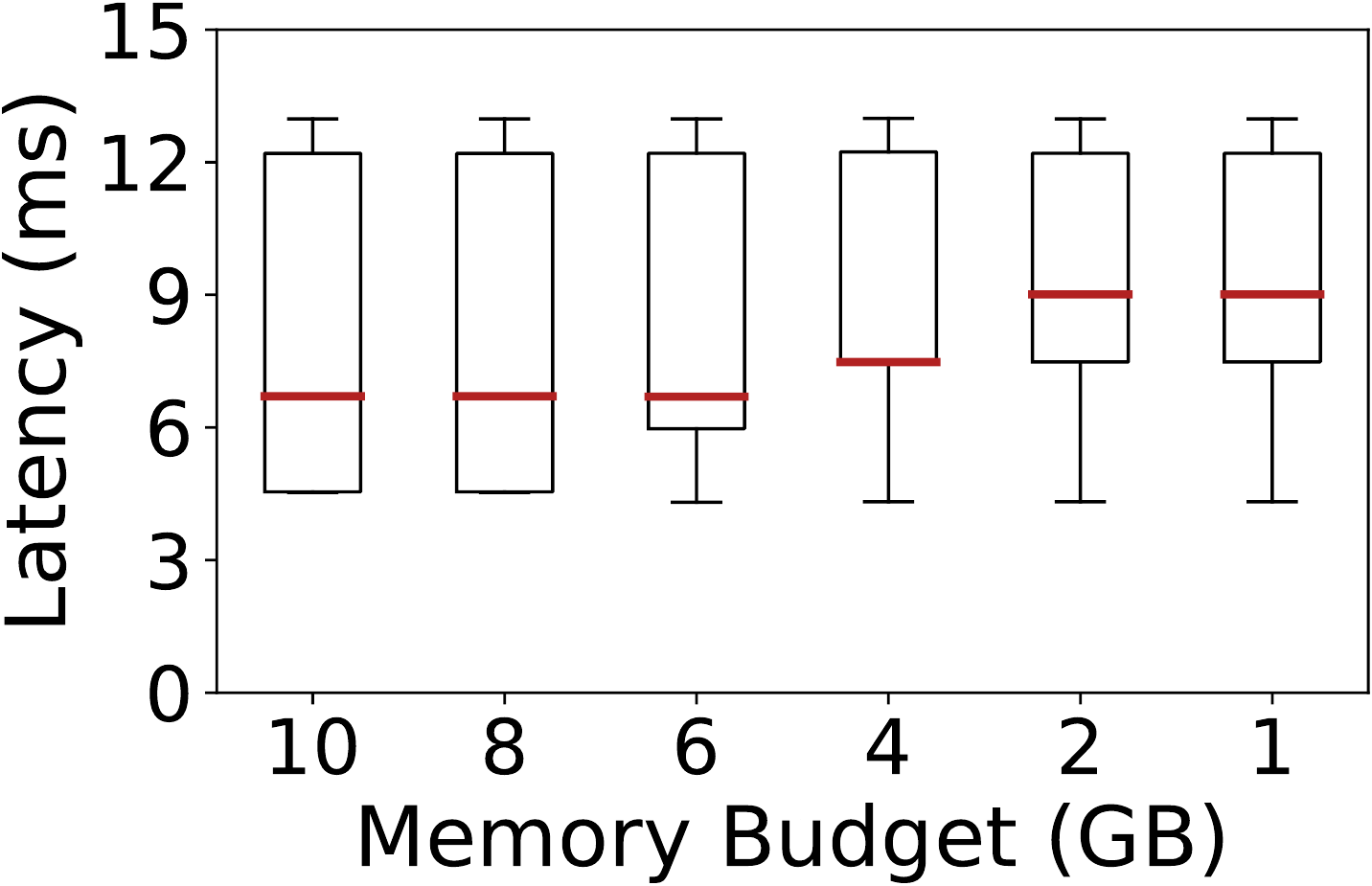}}
	\hspace{0.2cm}
	\subfloat[][]{\includegraphics[width=0.45\columnwidth]{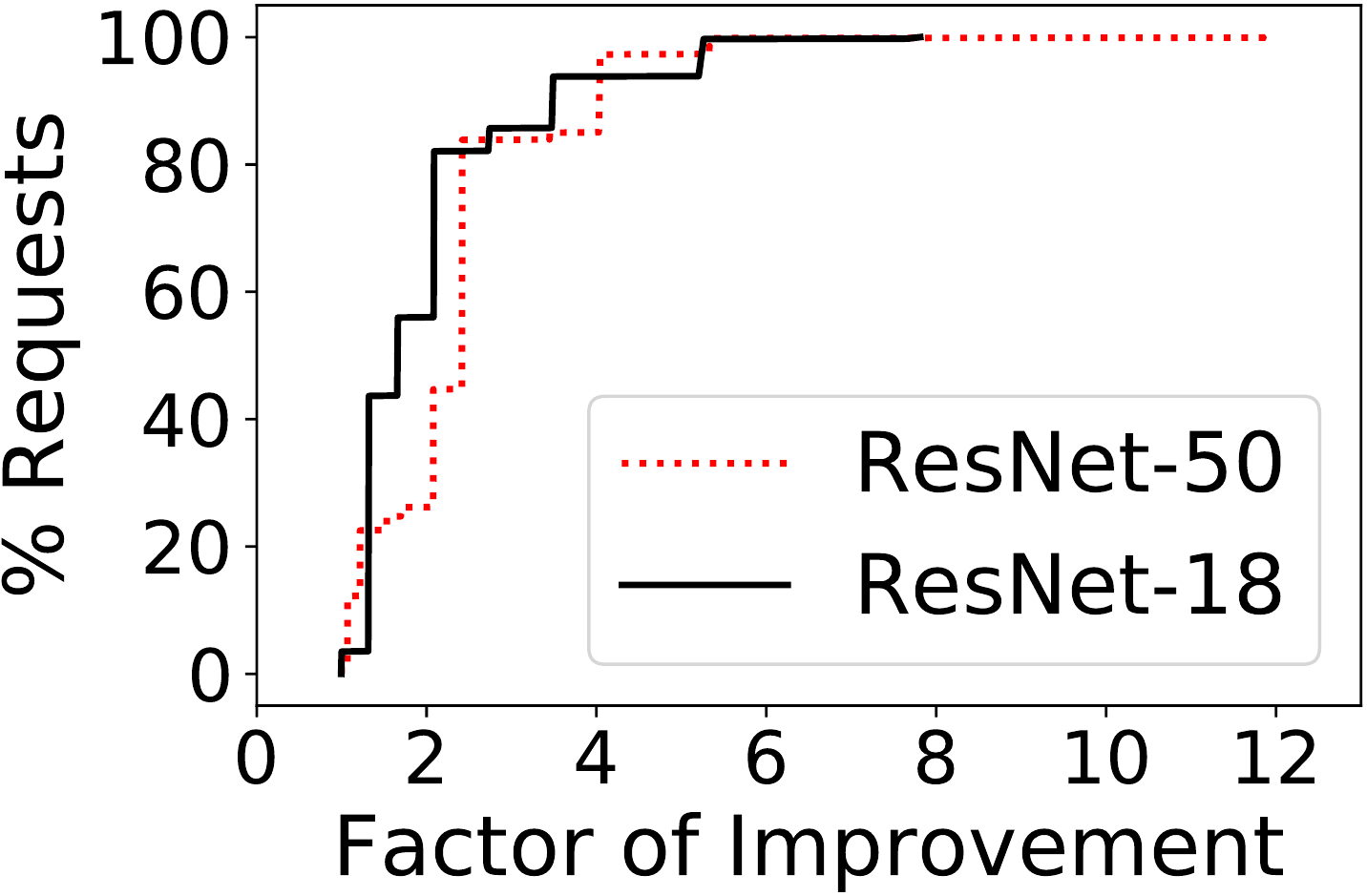}}
	\caption{\label{fig:micro} \textbf{(a) [GPU Inference, CIFAR-10] Memory budget impact (ResNet-50). 
	(b) [CPU Inference, CIFAR-10] ResNet-18 vs. ResNet-50.}}
\end{figure}

\begin{figure*}[t!]
        \captionsetup[subfloat]{captionskip=0.0pt}
        \centering
        \subfloat[][]{%
                \includegraphics[scale=0.34]{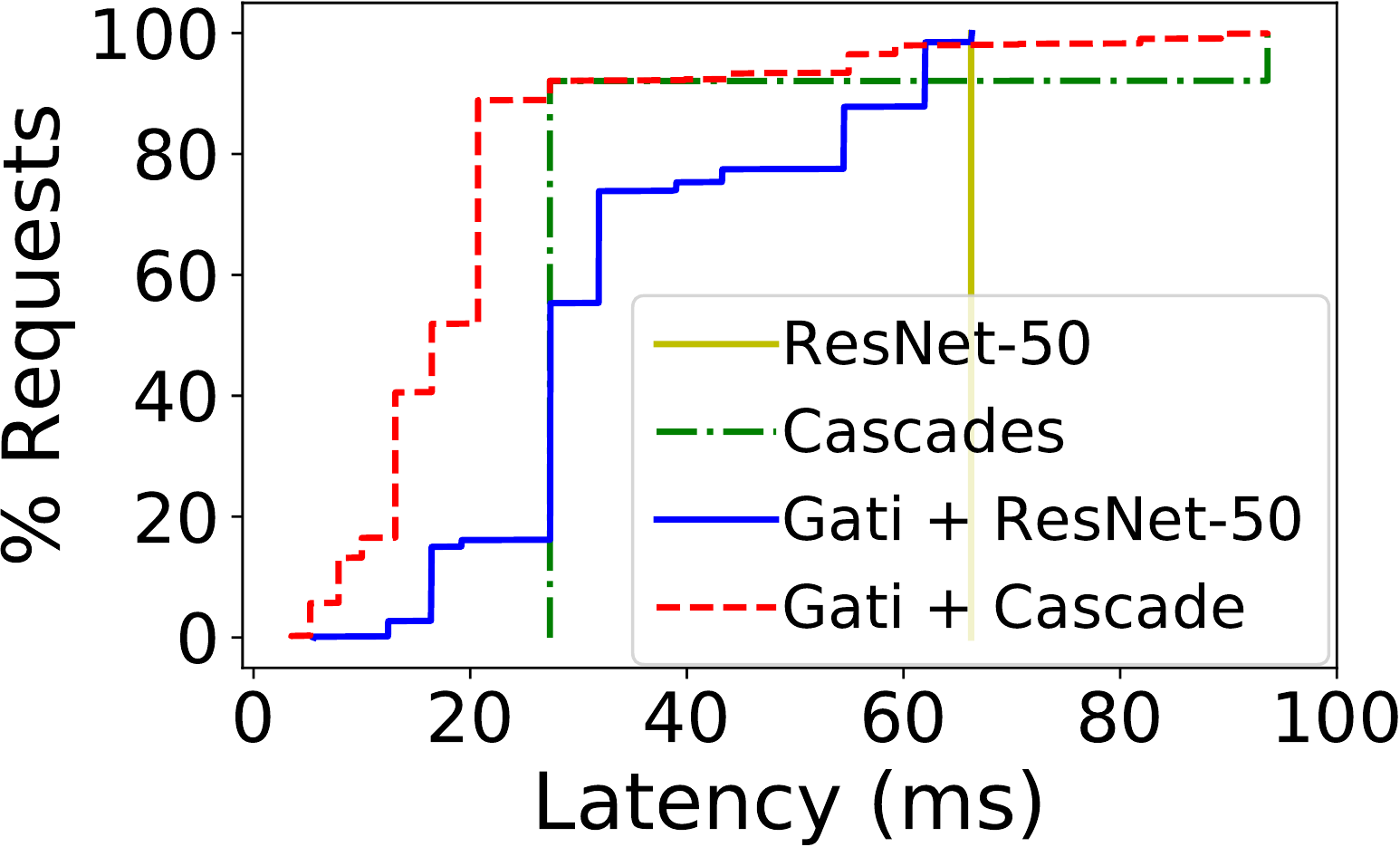}%
                \label{fig:cascades-comparison}
        }
        \hspace{0.2cm}
        \subfloat[][]{%
               \includegraphics[scale=0.22]{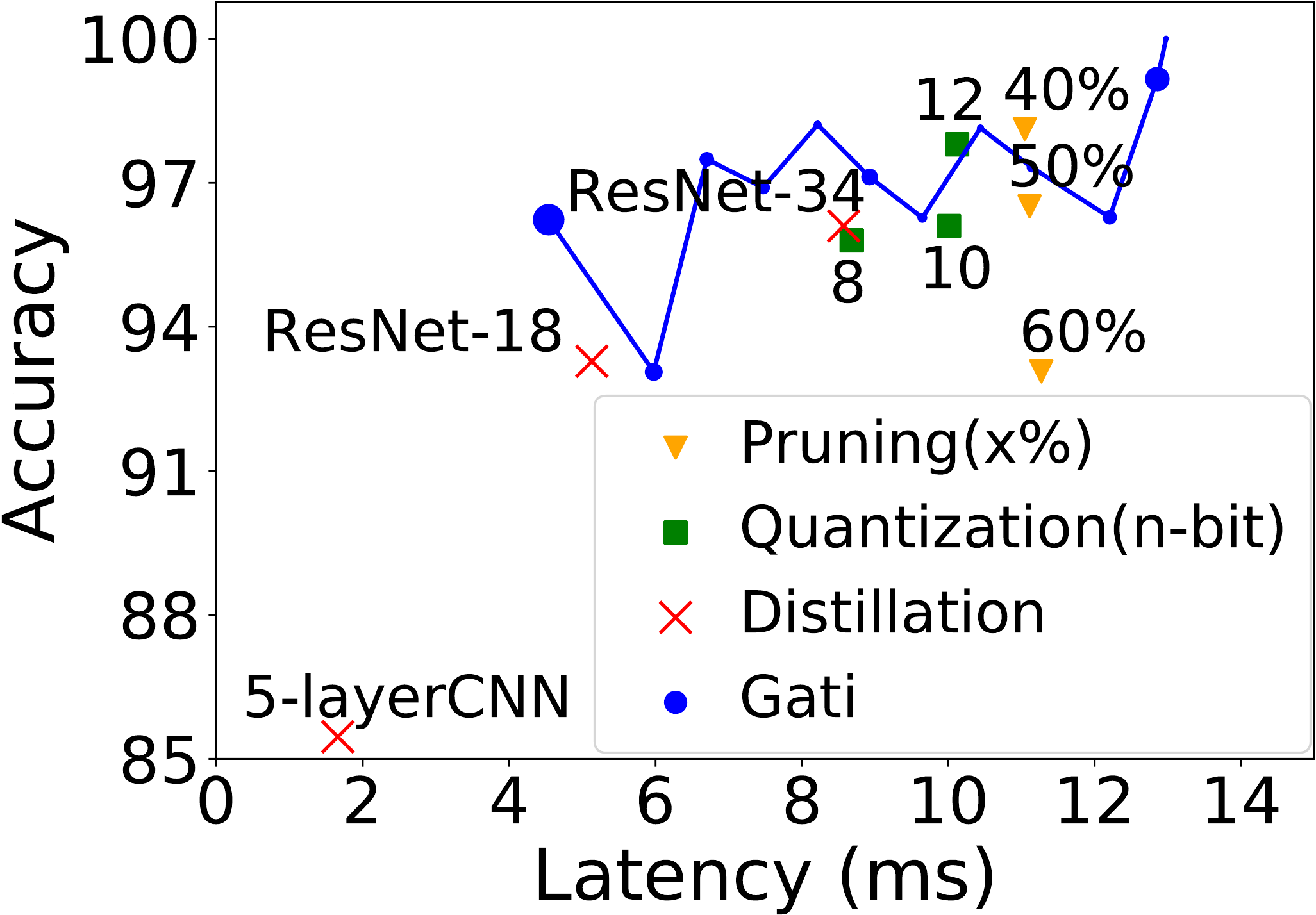}
               \label{fig:res50-macro-results-gpu}
        }
	\hspace{0.2cm}
        \subfloat[][]{%
                \includegraphics[scale=0.22]{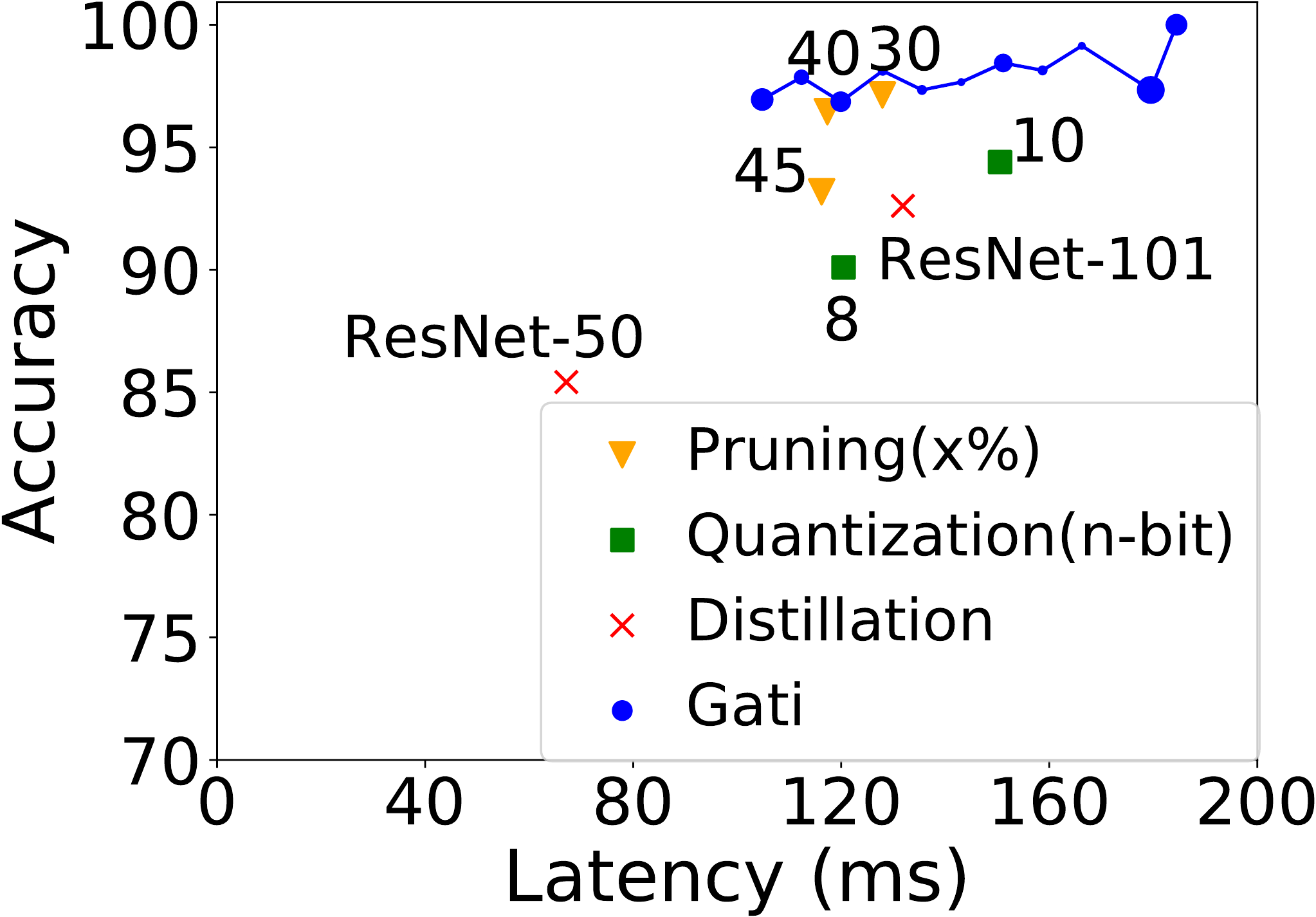}
                \label{fig:res152-macro-cpu}
        }

	\caption{\textbf{(a)[CPU Inference] \name{} vs. a cascade of DNNs (b)[GPU Inference] Accuracy-latency trade-off(ResNet-50 on CIFAR-10).
	(c)[CPU Inference] Accuracy-latency trade-off(ResNet-152 on CIFAR-100). Size of each \name{} marker in b,c is proportional to \% of requests at that latency point.}}
\end{figure*}

\noindent {\bf Against cascades: }
We evaluate \name{} against a cascade of progressively deeper models (ResNet-18, ResNet-50).
We configure the confidence threshold to achieve an accuracy of 97\%.
Figure~\ref{fig:cascades-comparison} shows that cascades can
give lower latencies than \name{} for a greater \% of requests
and even offer a slightly lower average latency.
However, \name{}'s tail latency (99\%-ile) is {\bf 1.68$\times$} lower.
Importantly, we observe that learned caches can {\em complement the models in a cascade}.
In this case we deploy learned caches for both models in the cascade and this
gives a {\bf 1.66$\times$} improvement over cascades' average latency.

\noindent {\bf Using GPUs: }
Figure~\ref{fig:res50-macro-results-gpu} shows that \name{} has an
average latency of \textasciitilde 7.98 ms, which is {\bf 1.63$\times$} lower than running ResNet-50 on a GPU.
\name{} exhibits a spectrum of latencies with an overall accuracy of {\bf 96.8\%} with respect
to the base ResNet-50 model.
Interestingly, the learned caches in this scenario occupy {\bf 8502 MB} memory,
which is 6.66$\times$ times the memory occupied for CPU inference.
This can be attributed to the observation that on GPUs, \name{} used {\bf six} learned caches where predictor networks
had a fully-connected architecture, while \name{} did not pick caches
with fully-connected architecture on CPU due to high lookup latencies.
Among this, we also noticed that \name{} preferred to pick fully-connected
architectures for initial layers in order to maximize the number of early cache hits.
This observation follows from the differences
between CPUs and GPUs highlighted in Table~\ref{tab:exploration_tradeoff}.
We also notice that some baselines do not work as well on GPUs.
For example, comparing Figure~\ref{fig:res50-macro-results-cpu}b and
Figure~\ref{fig:res50-macro-results-gpu}, pruning offers significant latency benefits
on CPUs but not on GPUs.
\emph{\name{} offers similar benefits on both CPUs and GPUs.}


\noindent {\bf One-time cost to train learned caches: }
From our testbed, we found that the time to
train learned caches depended on the layer of the base DNN.
Training took \textasciitilde 1 GPU hour for learned caches
associated with earlier layers since the hidden layer output dimensions
are typically much higher. We found that training later layers took \textasciitilde 20 GPU minutes.
Overall, we found that training learned caches For ResNet-50 required \textasciitilde 50 GPU hours.
Note that individual learned caches can be trained in parallel and are
hence amenable to speedup via scale-out.
Finally composing an optimal set of learned caches involves
solving an optimization problem that takes \textasciitilde 5 seconds.

\noindent {\bf Runtime Overhead: }
For inference on GPUs, we noticed that learned cache computation
on an average increased GPU utilization by \textasciitilde 14\%.
Even with this, we found that GPUs were still not fully utilized.
We note that
the tail latencies are {\bf within 1\%} of the latency
of the base DNN.
Our measurements show that adopting asynchronous cache lookups
reduces tail latency by 50.11\% compared to synchronous
lookups.

\subsubsection{Other Workloads}
\label{subsubsec:other-workloads}

We evaluate \name{} on three other workloads and observe that
\name{} can give similar latency benefits with an accuracy close to 97\%.
For sake of brevity, we present the graphs in Appendix (\secref{appendix:eval})
and summarize our findings in Table~\ref{tab:other_cpu_macro_results}.
We highlight a number of interesting insights from these workloads:

\noindent (i) {\em \name{} has better benefits on deeper models: }
From Figure~\ref{fig:micro}b, we observe that a greater
number of requests get higher latency improvements for a deeper
ResNet-50 model for CPU inference. This directly follows from the discussion in \secref{subsubsec:opportunities_motivation} that
deeper networks are built to achieve higher accuracy for a few "hard"
requests, thereby allowing \name{} to extract greater latency benefits
for "easier" requests through early cache hits.
We observe similar trends for GPU inference.

\noindent (ii) {\em \name{} offers high accuracy on difficult tasks: }
From Figure~\ref{fig:res152-macro-cpu}, we observe that \name{}
works well and minimally trades-off on accuracy 
even when the base DNN has lower accuracy
(94.2\% for CIFAR-10 vs
76.4\% CIFAR-100). However, we observe that obtaining earlier
cache hits is difficult when the base model itself does not
have high fidelity. Despite this, we note that nearly 40\% of requests
get \textasciitilde 2$\times$ latency benefit.

\noindent (iii) {\em \name{} offers benefits even under constrained memory budgets: }
We emulated this in our testbed by
giving reduced memory budgets in the composition phase.
This models scenarios where \name{} is deployed on low end GPUs.
\footnote{This assumes that latencies remain same
on lower end GPUs.}
From Figure~\ref{fig:micro}a, 
we see that when memory is not a bottleneck (10 GB), \name{}
improves average latency by 1.62$\times$ with respect to the base model.
With a memory budget of 1 GB, the average latency
is {\bf 1.44$\times$} better than the base model. 
Further, \name{} responds to reduction in 
memory budget by choosing learned cache variants that have
smaller memory footprints. With a budget of 10 GB, 
6 of the 13 chosen variants have a fully-connected
architecture, which has a greater memory requirement.
With a budget of 1 GB, \name{} chooses 
pooling and convolutional architectures, which have a
smaller memory footprint.

\noindent (iv) {\em \name{} offers benefits across accuracy targets: }
We modified the accuracy target to be 99\% instead of
97\% and observed that \name{} was able still able to give a 1.41$\times$
improvement in average latency over the base ResNet-50 model with an accuracy of 99.12\%.
Though the latency benefit at an accuracy target of 99\% is lower than 97\%
by 1.38$\times$, \name{} is able to deliver on its promise of
meeting the accuracy target and still offer latency benefits.

\subsection{Adapting to Temporal Locality}


In this section, we evaluate the ability of \name{} to exploit
temporal locality in online workloads.

\noindent {\bf Workload: }
We use 4 publically available videos -
Two of these are traffic camera videos (\em{bellevue}~\cite{focus} and \em{bend}~\cite{bend-data})
while the other two are surveillance camera videos (\em{oxford}~\cite{oxford-data} and \em{sittard}~\cite{sittard-data}).
Each of these has 5-6 hours of video footage.
Similar to ~\cite{noscope}, we perform object classification on the video at 1 fps.
All of our evaluations use ResNet-18 model trained on CIFAR-10.
The initial set of learned caches are constructed 
using validation data from CIFAR-10.

\noindent {\bf Parameters: }
The inference service accumulates a batch of 100 requests and samples
20\% for use in the cache adaptation service.
We maintain 60 minutes worth of windowed samples.
The cache adaptation service retrains caches every 15 minutes
and immediately deploys them. 
We use a learning rate of 0.002 and retrain the learned caches for 5 epochs. 

\noindent {\bf Baseline: }
We compare against \textsc{Static-}\name{},
where the
cache adaptation service is disabled.
Latencies get inflated when the inference service 
updates learned caches before executing a request.
We filter such latencies in our results.


\subsubsection{Results}

\label{subsec:eval_online}

\begin{figure}[t!]
        \subfloat[][CPU Inference]{\includegraphics[width=0.48\columnwidth]{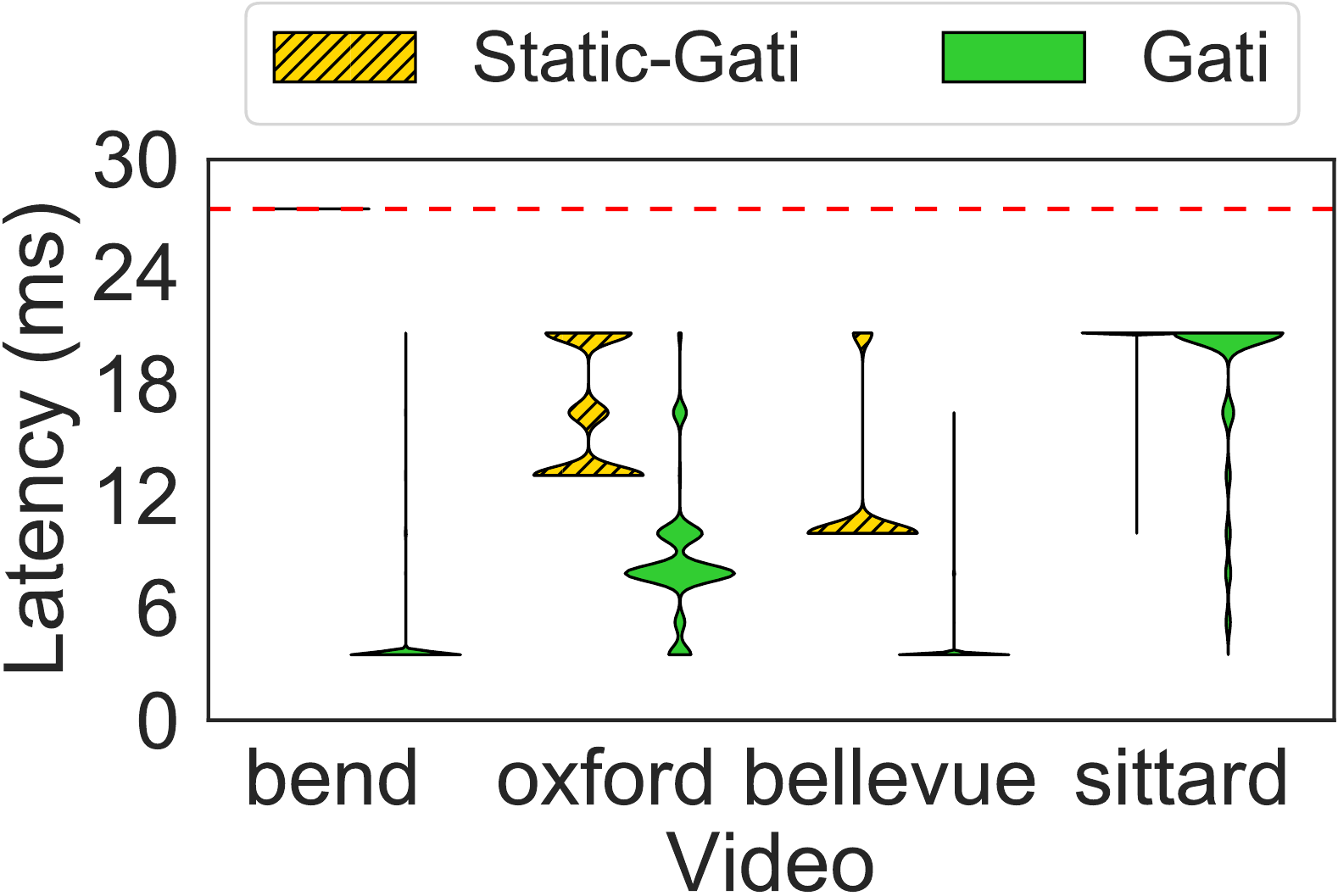}}
	\hspace{0.1cm}
        \subfloat[][GPU Inference]{\includegraphics[width=0.48\columnwidth]{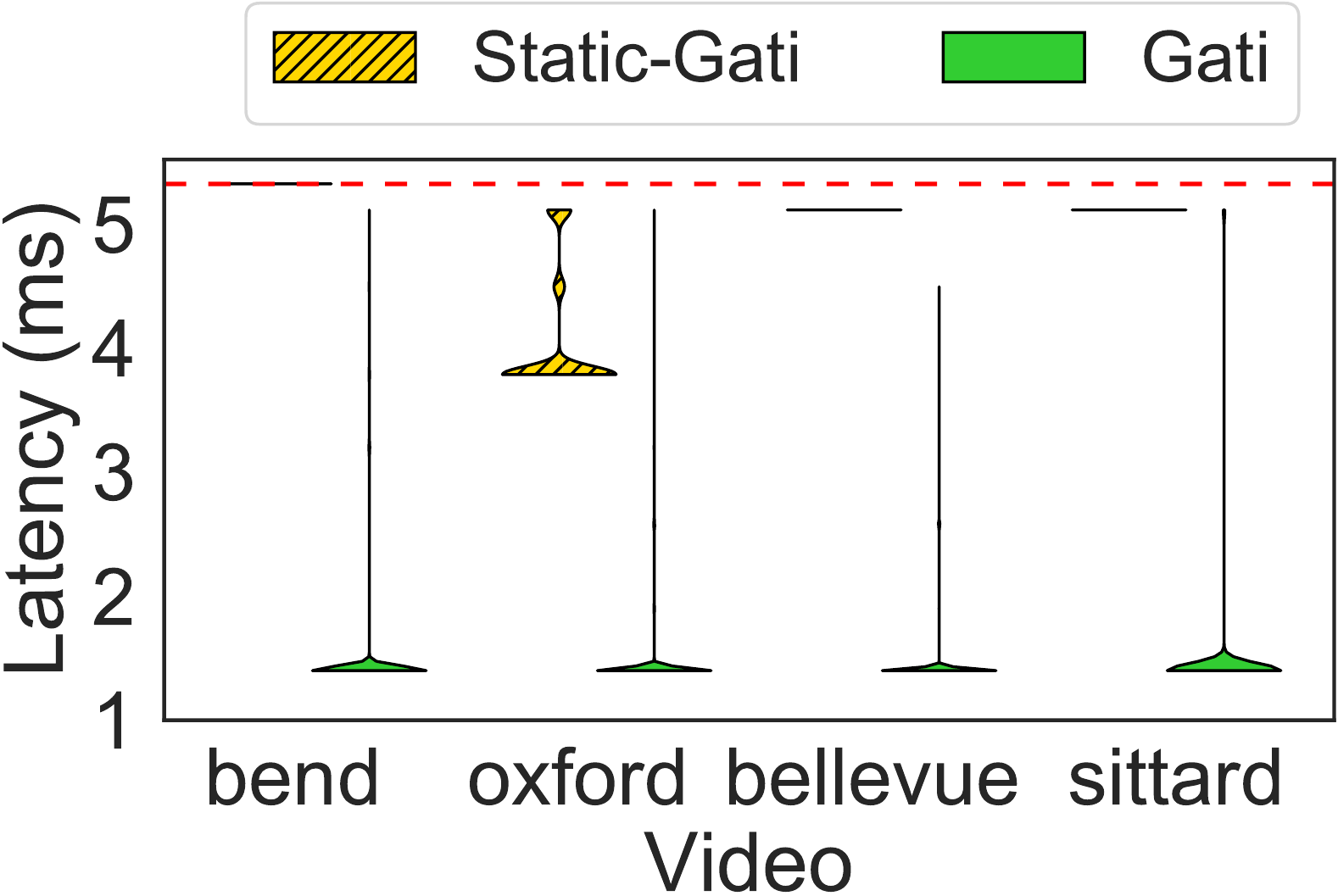}}
        \caption{\label{fig:online_video} \textbf{Latency benefits on various videos. Thickness shows the density of requests at that latency. Dashed line for ResNet-18.}} 
\end{figure}

Figure~\ref{fig:online_video} shows the latency
benefits of adapting to inherent temporal locality on both CPUs and GPUs.
Across all videos, we observe that \name{} is able to improve the
average inference latency by a factor of {\bf 1.45$\times$-7.69$\times$} on CPUs
with respect to the base ResNet-18 model. Additionally, none of the inference requests
required computing the entire DNN.
We also observe that the accuracy in all of the cases is above the target of 97\%.

\name{} also outperforms \textsc{Static}-\name{} by up to a factor
of 4.61$\times$. The relative performance between the two varies according
to the video. We see that in one case (the {\em bend} video),
\textsc{Static}-\name{} is unable to yield any cache hits and the
performance matches that of the base DNN.
This happens when the learned caches are trained on data that
looks different from the inference requests.
Retraining learned caches accounts for this variance in addition to exploiting temporal locality.

\begin{table}[t]
  \centering
  \begin{scriptsize}
  \begin{tabular}{ c | c | c}
    \hline
	  \textbf{Block} & \textbf{\textsc{Static}-\name{} Acc./Hit-Rate} & \textbf{\name{} Acc./Hit-Rate} \\
    \hline
	  1 & 27.95\%/0\% & \textbf{90.56\%/11\%} \\
	  \hline
	  3 & 30.21\%/0\% & \textbf{93.55\%/64\%} \\
	  \hline
	  6 & 85.85\%/31\% & \textbf{94.82\%/83\%} \\
    \hline
  \end{tabular}
  \end{scriptsize}
	\caption{\textbf{[oxford video] Blockwise improvement in the predictor network accuracy yields hit rate improvements
	for \name{} over \textsc{Static}-\name{}.}}
  \label{tab:sources_improvement_video}
\end{table}

\noindent {\bf Sources of Improvement: }
Figure~\ref{fig:online_video}
shows that \name{} is able to get more cache hits at earlier layers
leading to a number of requests with low
latencies. Table~\ref{tab:sources_improvement_video} shows the 
reason behind this trend. We observe that retraining drastically
improves the fidelity of predictions from the predictor network compared
to \textsc{Static}-\name{}, with
the change being highest in earlier layers.
High fidelity predictions allow the selector network to yield
more cache hits and improve the end-to-end latency.

\noindent {\bf Impact of batching: }
To evaluate the impact of batching requests, we modify \name{} to wait
for a batch and then run inference on a GPU.
We execute each batch until both requests have either had a cache hit or reached the end of DNN.
We record the latency from when inference starts on GPU to discount the effects of queuing
while forming batches.
With a batch size of 2,
\name{} gives 2.88$\times$-2.93$\times$ improvement in average latency
in comparison to 3.67$\times$-3.75$\times$ with batch size of 1.
This is because, with batching, some requests might
need to wait for other requests in the batch to finish execution.

\noindent {\bf Retraining is quick: }
Our measurements show that using simple predictor and selector networks
enables retraining learned caches to finish in \textasciitilde 480 ms.
This enables fast adaptation to workloads where temporal locality could
change rapidly. 

\subsubsection{Insights}
\label{subsubsec:online-insights}
We now present some insights showing the impact of the workload characteristics
and system parameters on the ability of \name{} to exploit temporal
locality windows.

\begin{figure}[t!]
        \subfloat[][]{\includegraphics[width=0.49\columnwidth]{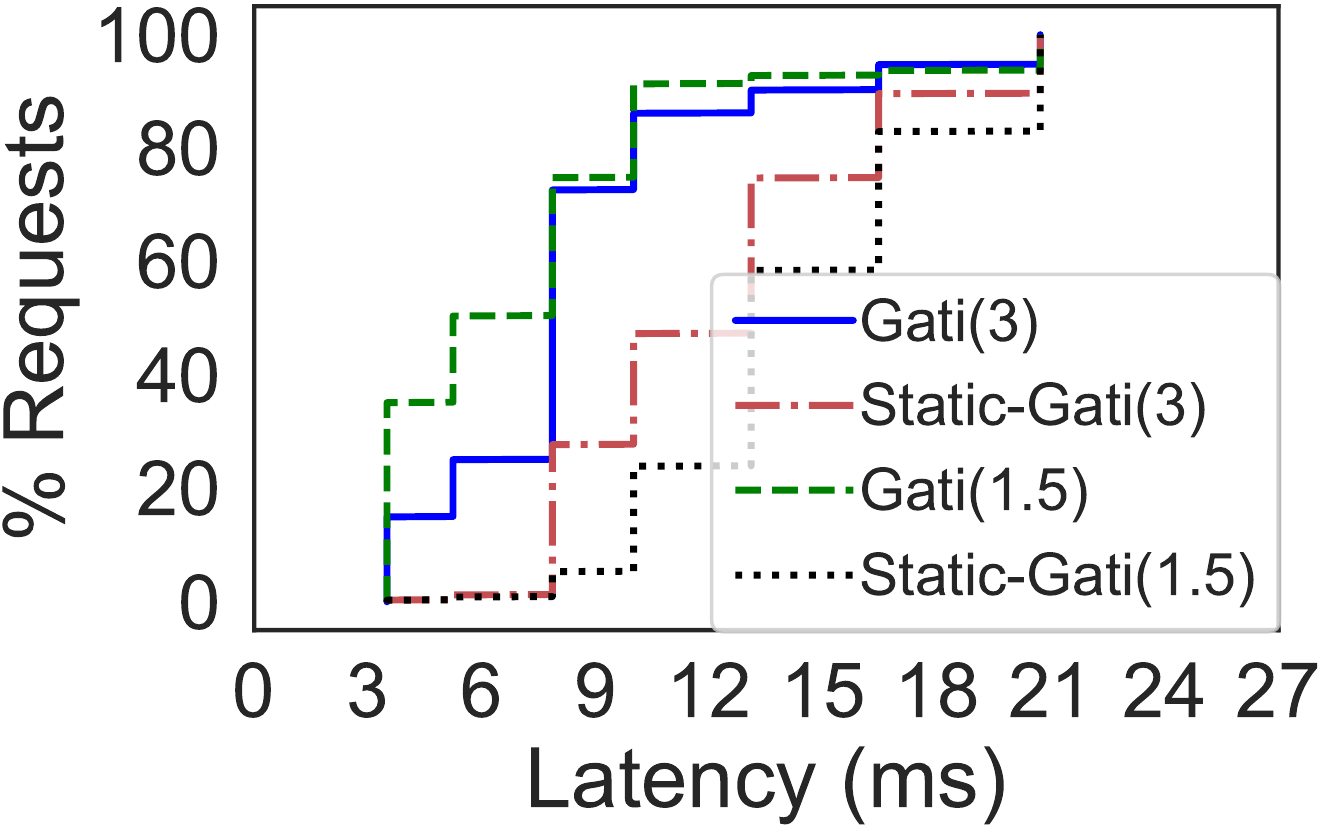}}
        \subfloat[][]{\includegraphics[width=0.49\columnwidth]{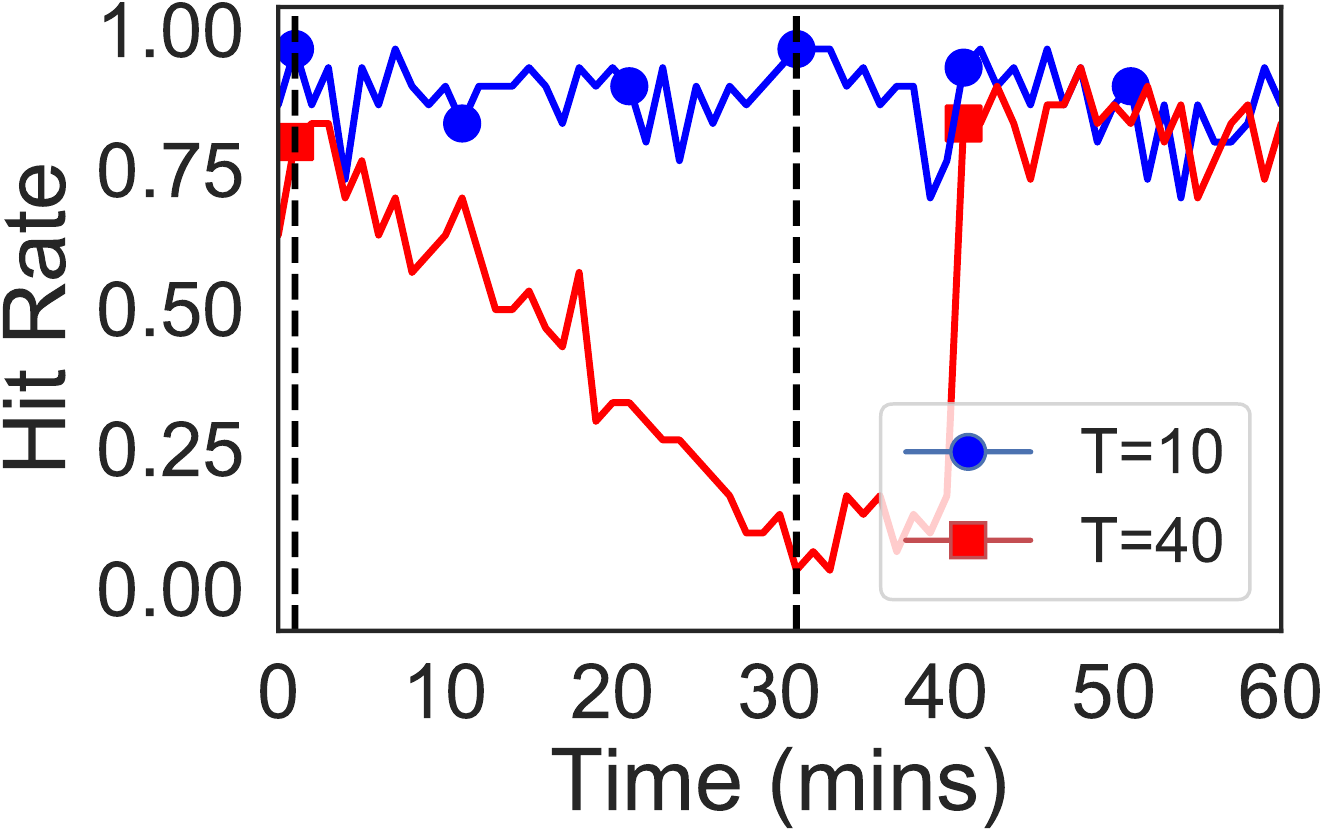}}
	\caption{\label{fig:zipfian-interval} \textbf{(a) Impact of skew in distribution. $\alpha$
	values in legend.
	(b) Impact of retraining intervals [Block 3, ResNet-18]. Markers indicate retraining
	and dashed line shows distribution change.}}
\end{figure}

\noindent (i) {\em Benefits on more skewed output class distributions: }
Since it is hard to control the distribution of objects over time
in real videos, we use a synthetic workload to study 
the impact of varying class distributions. 
We use a Zipfian distribution where the parameter $\alpha$ is used to control the
distribution skew of object classes with lower values indicating greater skew.
We consider two skew levels with $\alpha$ values of 1.5 and 3.0.
The dominant class in the distribution changes every 15 minutes.

Figure~\ref{fig:zipfian-interval}a shows that \name{} performs better when the distribution
is more skewed, offering 3.6$\times$ improvement in latency compared to a 3.2$\times$
gain with lesser skew. This correlates well to the intuition that caches perform better 
when the workload has high occurrences of some popular objects. 

\noindent (ii) {\em Small sampling rates suffice : }
We observe that the end-to-end latency of inference is not
impacted unless the sampling rate becomes less than 10\%,
motivating us to choose 20\% for our experiments.
Interestingly, this points out that a small number of samples are sufficient to capture temporal locality
effects in realistic workloads such as videos.

\noindent (iii) {\em \name{} is sensitive to the periodicity of retraining: }
To evaluate this aspect, we consider the same
Zipfian workload used in (i) with $\alpha$=1.5 and a temporal
window of 20 minutes.
We vary the interval at which we retrain caches.
We observe from Figure~\ref{fig:zipfian-interval}b that a lower interval of 10 mins
helps \name{} maintain high cache hit rates and adapt quickly to changes in distribution.
With a larger update interval (40 mins), we observe that the hit rate drops continuously and
increases again only after the learned caches have been retrained.

\subsection{Incremental Replanning Benefits}
\label{subsec:eval_replanning}
We deploy \name{} with the cache adaptation service disabled
to study the benefits of incremental replanning in isolation.

\noindent {\bf Workload: }
We use the same traffic analysis application example as discussed in~\secref{sec:system-design}.
We use learned caches in the object detection module.
We sample images from the LFW dataset~\cite{lfw},
Stanford Cars dataset~\cite{stanfordcars}, and CIFAR-100 and issue
them as input requests. The input distribution consists of 46\% faces and
45\% vehicles.
Each DNN model outlined in Table~\ref{tab:query_replanning_models}
is deployed on a separate CPU instance.

\noindent {\bf Baselines: }
We consider both
the equal split and smart split policy (\secref{subsubsec:split_policy}).
We evaluate the performance of \name{} 
against a 
baseline that does not use learned caches or replanning.

\begin{figure}[t!]
        \centering
        \subfloat[][Equal Split]{\includegraphics[width=0.45\columnwidth]{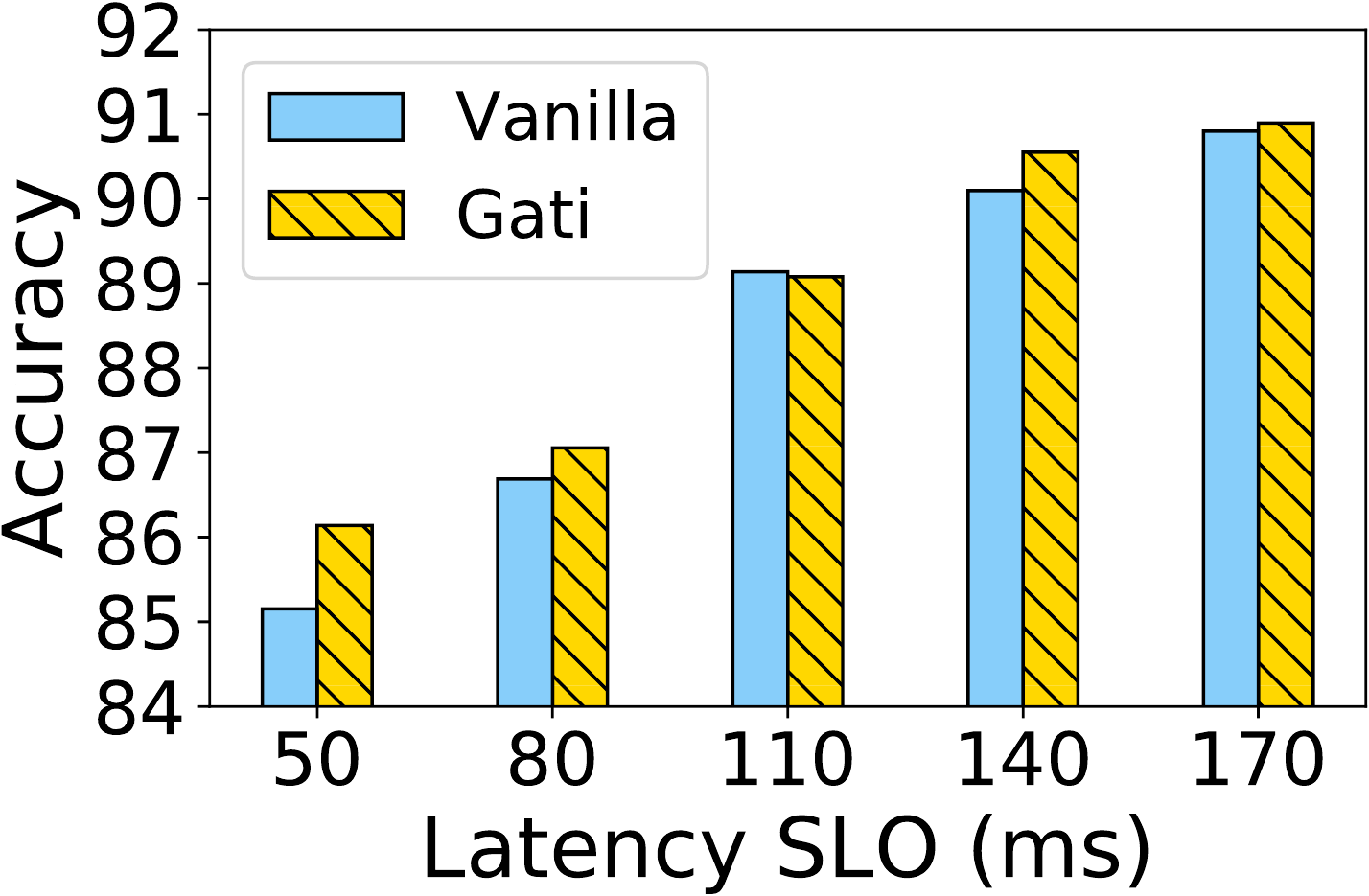}}
	\hspace{0.2cm}
        \subfloat[][Proportional Split]{\includegraphics[width=0.45\columnwidth]{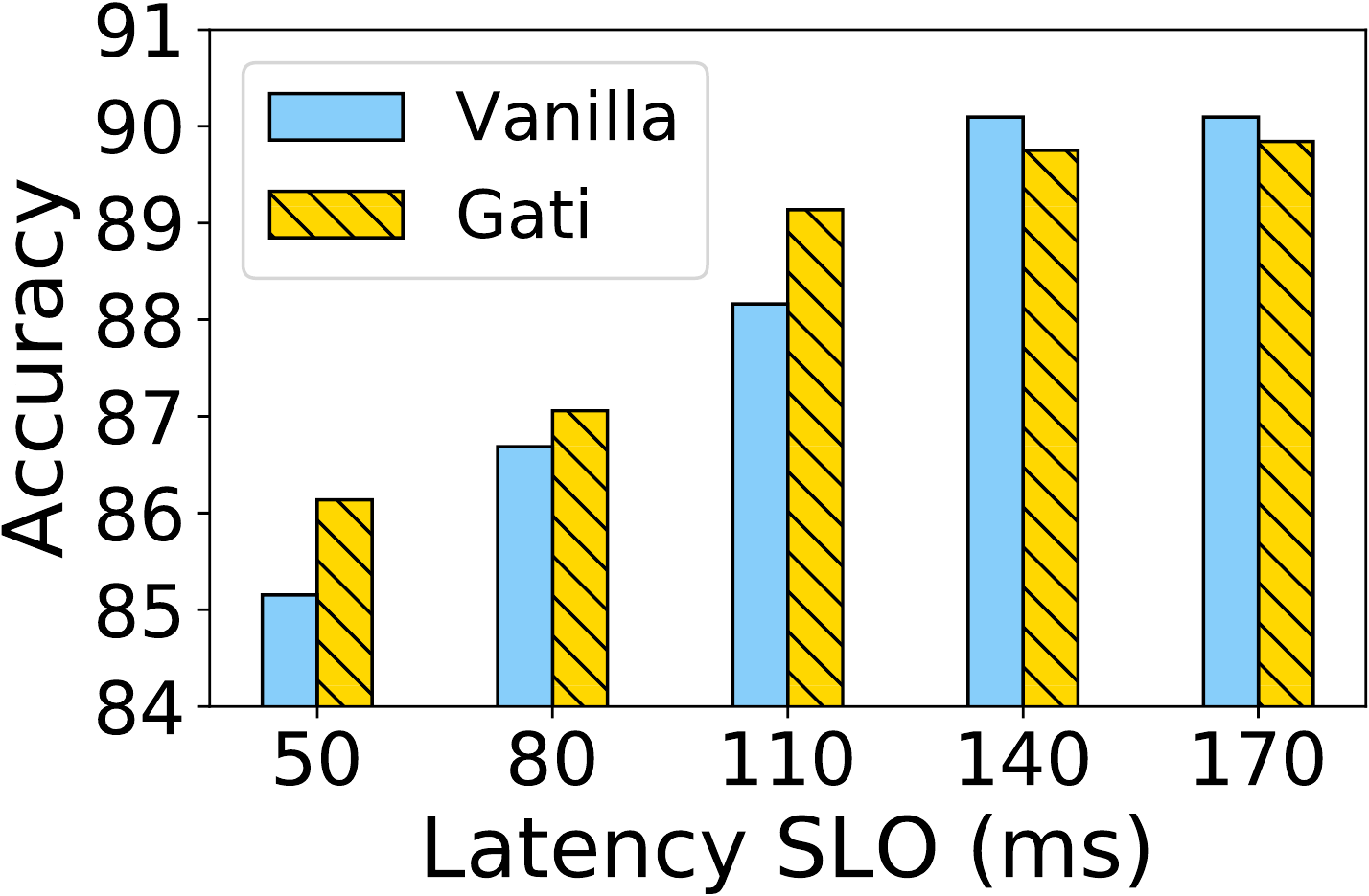}}
	\caption{\label{fig:replanning_benefit} \textbf{Accuracy benefit from incremental replanning in \name{}}}
\end{figure}

\begin{figure}[t!]
        \centering
        \includegraphics[clip,width=0.9\columnwidth]{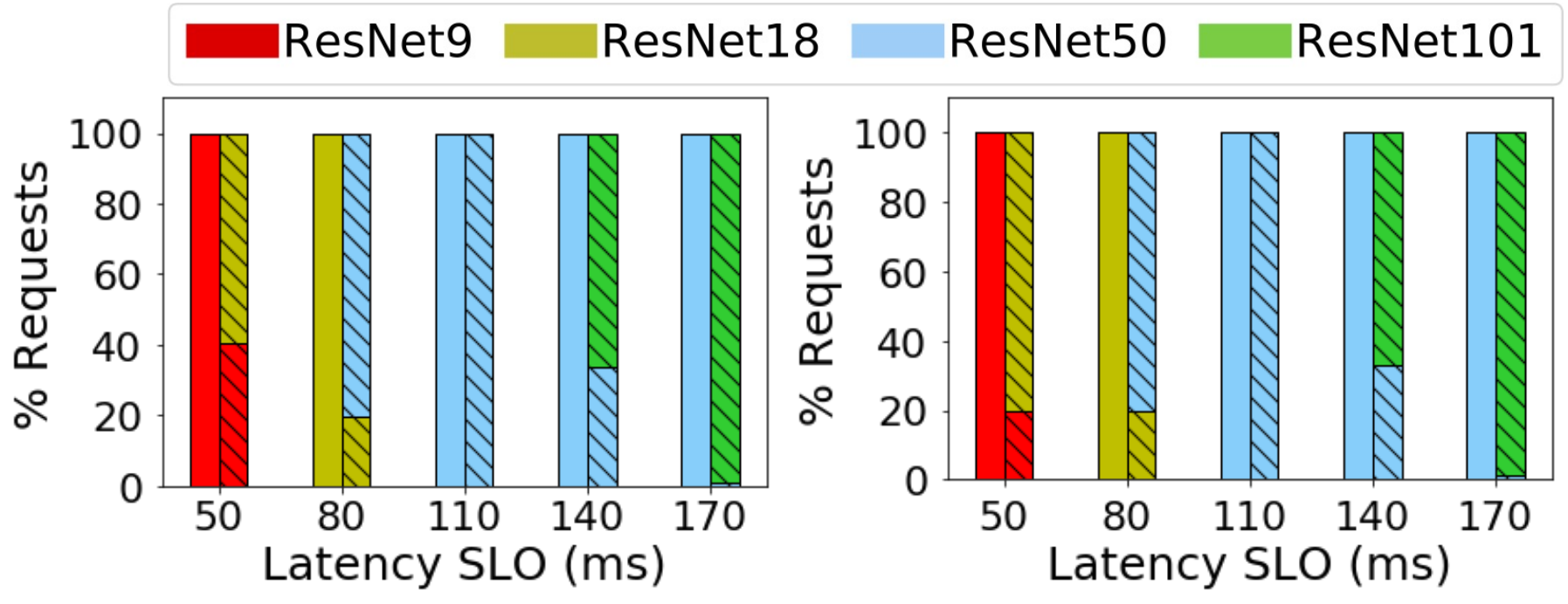}%
	\caption{\textbf {DNNs used for (a) face (b) vehicle recognition for different SLOs. Shaded bars are for \name{}
			  and indicate that incremental replanning allows more accurate DNNs to be executed.}}
        \label{fig:replanning_sources_improvement}
\end{figure}

\noindent {\bf Results: }
Figure~\ref{fig:replanning_benefit} shows
that incremental replanning gives
accuracy benefits for a variety of latency SLOs, with up to 
\textasciitilde {\bf 1\%} improvement
in end-to-end accuracy and this applies to both 
the equal and smart split policy.
We observe greater benefits at tight latency SLOs ($\leq$ 100 ms) since
this is when cheaper models can be replaced with more expensive
ones through replanning.
Figure~\ref{fig:replanning_sources_improvement} provides insight into
why replanning is able to give accuracy benefits.
As an example, at a latency SLO of 80 ms,
the baseline uses SE-LResNet18E-IR for face recognition
and ResNet-18 for vehicle recognition.
Due to saved latencies obtained by learned cache hits,
\name{} is able to use a more accurate SE-LResNet50E-IR model for
~\textasciitilde 79\% of face recognition tasks and a 
more accurate ResNet-50 for ~\textasciitilde 80\% of vehicle recognition tasks.
Both of these lead to greater end-to-end accuracy for the query.


We see that \name{} can reduce
average latencies by up to {\bf 1.26$\times$} along with an accuracy improvement of ~\textasciitilde {\bf 1\%}.
Thus, incremental replanning helps overcome the fundamental latency-accuracy trade-off by improving on both metrics.

\section{Related Work}
\label{sec:related_work}

\noindent{\bf Prediction Serving Systems. }
\name{} is a prediction serving system just like
~\cite{clipper,tfserving,nexus,paritymodels,pretzel,willump,laser}.
\name{} enables late-binding in DNNs through learned caches unlike \cite{clipper} which
treats the model as a black-box and applies caching only at the input layer.
\name{} borrows the idea of expressing queries as DAGs from \cite{nexus},
but uses it to improve accuracy and not for resource efficiency.
\cite{paritymodels} is orthogonal to \name{} since it focuses on failure resilience and improving tail latencies.
NoScope~\cite{noscope} uses a cascade of models to achieve a limited form of late-binding.
We show in \secref{subsubsec:res50_results} that learned caches can complement cascades to give further latency benefits.
\cite{tfserving,focus,willump} focus on optimizing the serving pipeline but do not focus improving
DNN inference latency.
\cite{laser} uses caching to improve predictions for recent data but targets linear models
like logistic regression and not DNNs.

\noindent{\bf Accuracy-latency trade-off. }
\cite{anytime,skipnet} complement DNNs with auxiliary networks just like in \name{},
but provide an option for configurable latency-accuracy trade-off. 
A number of systems~\cite{rafiki,videostorm} also enable such trade-offs.
\name{} however lowers latencies without significant trade-off in accuracy.

\noindent{\bf High performance inference. }
A number of systems~\cite{taso,tvm,tensorrt} optimize the computation
graph for faster DNN inference.
Others specifically focus on RNNs~\cite{cellularbatching,deepcpu} and CNNs~\cite{spgcnn}.
All of these complement \name{}'s goal of late-binding the
amount of computation to the hardness of the input.

\noindent{\bf Query Planning. }
\cite{infaas} outlines an idea of automatically picking
models based on the latency SLO. \name{} builds on this idea and provides inference service owners 
an option to specify an array of models at each node of the query DAG.
In other domains, several other systems
have used query replanning and dynamic adaptation to resource 
or workload changes~\cite{qoop,clarinet,adaptivequeryprocessing,adaptivestreamprocessing,rethinkingadaptive}
to get performance benefits.

\noindent{\bf Using ML in Systems. } 
A number of projects use ML to improve system design.
This includes use of ML for database
indices~\cite{learnedindices}, cache replacement~\cite{lecar,learnedreplacement},
memory management~\cite{learnmemalloc}.
cluster scheduling~\cite{decima,dl2},
resource management~\cite{microsoftrc}, packet
classification~\cite{neuralpacketclassification}, and video content
delivery~\cite{adaptivedelivery,penseive,learninginsitu}.  Per our
knowledge, \name{} is the first work to use ML models as caches
to accelerate DNN inference.

\section{Conclusion}
In this paper we presented \name{}, a low-latency
prediction serving system that uses caching to late-bind DNN inference computation.
We proposed using {\em learned caches}, where
the cache is represented by neural networks
and outlined a design for their construction.
Our evaluations show that \name{} can improve average latency for
easier inputs and exploit temporal locality to help online workloads.


\bibliographystyle{abbrv}
\bibliography{freeze}

\appendix
\section{Appendix}
\label{sec:appendix}

\subsection{Composing learned caches}
\label{subsec:appendix_optimization}

The goal of the composition phase is to jointly select
a global set of learned caches that minimize the expected
average latency, while ensuring minimal degradation in
accuracy, and meeting computation and memory constraints.
We can formulate this as a mix-integer quadratic
program.

Let us assume that we have a DNN with $N$ layers for which we wish to construct learned caches.
Let us consider that for each layer of the DNN, we have an array of $K$ possible variants.
Let us consider that we have a memory budget of $M$.

From the exploration phase, we
obtain the following metrics for each variant - Hit Rate ($H_{i,j}$),
Accuracy ($A_{i,j}$), Lookup Latency ($T_{i,j}$), and Memory Cost
($M_{i,j}$). Additionally, we profile the latency for the computation
of each layer ($L_{i}$).

\subsubsection{Optimization Problem}
\label{subsec:optimization}

Given the above, we wish to decide which layer(s) of the DNN to use learned caches for 
and which variant to use at those layers.

\noindent {\bf Indicator Variables.}

We use the indicator variable $b_{i,j}$ to indicate if a variant $j$ at layer $i$ is chosen. 
$b_{i,j}=1$ means we chose that particular variant at that layer; 0 means we do not choose.

We use the indicator variable $c_{i,j}$ to indicate if layer $j$ is the first layer after layer $i$ to have a learned cache (This corresponds to a 1).
For instance, consider a neural network with 5 layers - $L_1$, $L_2$, .. $L_5$.
Let us assume that we ultimately need caches only for layers $L_1$, $L_3$, and $L_5$. In such a case, $c_{1,3}$ and $c_{3,5}$ will be 1 and the rest will be 0.

\noindent{\bf Derived Metrics.}

The hit rate $H_{i,j}$ is an independent metric that is measured for each variant at each layer.
However, when the cache at an earlier layer gives hits,
it reduces the number of input samples available at a later layer. 
To model this, we measure an {\em effective cache-hit rate} $EH_{i,j}$ as defined below -

\begin{equation}
  \text{$EH_{i,j}$} := H_{i,j} - \sum_{k=1}^{i-1}(c_{k,i}.\sum_{m=1}^{K}(b_{k,m}.EH_{k,m}))
\end{equation}

\noindent The above equation subtracts the effective cache hit rate of the previous layer 
that has a learned cache in a recursive fashion.
The base case is $EH_{1,k}=H_{1,k}$ for all $k \in K$.

\noindent {\bf Constraints}

From the definition of the indicator variable $c_{i,j}$,
we formulate a constraint that at most one layer $j$ 
after a layer $i$ can have $c_{i,j}=1$. 

\begin{equation}
  \forall i \sum_{j=i+1}^{N}c_{i,j} \leq 1
\end{equation}

\noindent {\bf Accuracy constraint.}
\noindent Given that variant at each layer has an accuracy $A_{i,j}$,
we need to ensure that the cumulative accuracy is greater than a minimum {\em accuracy threshold} $A$ -

\begin{equation}
    \sum_{i=1}^{N}\sum_{j=1}^{K}(b_{i,j}.EH_{i,j}.A_{i,j}) + (1 - \sum_{i=1}^{N}\sum_{j=1}^{K}(b_{i,j}.EH_{i,j})) \geq A
\end{equation}

\noindent The first part of the equation measures 
the effective accuracy for all inputs predicted by learned caches 
while the second part captures the accuracy component for inputs that go through the entire DNN (which get 100\% accuracy).

\noindent{\bf Resource constraints.}
There are two resource constraints to consider - (i) memory and (ii) computation.

\noindent{\bf Memory constraint: }
The total memory occupied by the chosen variants 
should be within the memory budget $M$.

\begin{equation}
  \label{eqn:memory-constraint}
  \sum_{i=1}^{N}\sum_{j=1}^{K}(b_{i,j}.M_{i,j}) \leq M
\end{equation}

\noindent{\bf Computational constraint:}
The underlying hardware imposes computational constraints on the cost of that computation in terms of latency 
(captured by $T_{i,j}$), and the degree of parallelism with which the computation can be done.
We assume a simple computational model that we can have atmost one cache lookup at any
given point of time. This lookup happens asynchronous to the base DNN computation.



With this information, we need to ensure that cache look-ups do not coincide in time. Hence, we have -

\begin{equation}
  \label{eqn:comp-constraint}
  \forall i \sum_{k=1}^{K}(b_{i,k}.T_{i,k}) \leq \sum_{k=i+1}^{N}L_{k} - \sum_{k=i+1}^{N}(c_{i,k}.\sum_{m=k+1}^{N}L_{m})
\end{equation}

In the above constraint, the LHS captures the running time of the cache lookup at layer $i$.
The RHS subtracts the remaining running time for layer $i$ to the end of the DNN 
from the remaining running time for layer $j$ to the end of the DNN, where $j$ is the earliest layer after $i$ to have a learned cache.
Essentially, for a layer $i$ that has a chosen variant, 
we measure the running time between layer $i$ and $j$ and constraint 
it to be at least as large as the cache lookup time at layer $i$.




{\bf Objective}
We wish to minimize the expected latency for a given input request.
We capture this as below -

\begin{equation}
\begin{aligned}
	\text{\textbf{min.}} \sum_{i=1}^{N}\sum_{j=1}^{K}(b_{i,j}.EH_{i,j}.(\sum_{k=1}^{i}L_i + T_{i,j})) \\
	+ (1 - \sum_{i=1}^{N}\sum_{j=1}^{K}(b_{i,j}.EH_{i,j}))(\sum_{i=1}^{N}L_i)
\end{aligned}
\end{equation}{}

\subsubsection{Relaxed Formulation}
The problem with above solution is that it not linear with respect to the indicator variables.
\name{} adopts a three-step approach to simplify the above formulation.
The key idea is to separate out the concerns of accuracy, overall latency, and resource contraints (computation and memory).
We use the same indicator variables $b_{i,j}$ and $c_{i,j}$ as described above.


\noindent{\bf Step 1. Accuracy Filter: }
As a first step, we filter out model variants $M_{i,j}$ whose accuracy $A_{i,j}$ is below a minimum threshold $A$.
To formulate this as a constraint -

\begin{equation}
	b_{i,j} = 0, \forall \text{i,j } A_{i,j} < A
\end{equation}

\noindent{\bf Step 2. Score Computation: }
We consider two factors in determining
the importance of a particular variant:
(i) The hit rate of the variant and
(ii) The {\em latency gain} obtained by using the model variant in the event of a cache hit.

\noindent We compute latency gain (LG) for a learned cache variant
as the ratio of running time for the entire DNN to the running time
assuming that a cache hit is obtained at the given learned cache variant.

\begin{equation}
     LG_{i,j} = \sum_{k=1}^{N}L_{k}/(\sum_{k=1}^{i}L_{i} + T_{i,j})
\end{equation}

We prefer higher hit rates and higher latency gains. However, these are
fundamentally at odds with each other since higher latency gains are
obtained using variants at earlier layers of the base DNN where the
hit rates would be lower, and vice versa.  To balance these two
factors, we compute a score ($S$ - higher is better) that captures
the benefit of using a variant:

\begin{equation}
	S_{i,j} = \alpha.(H_{i,j}) + (1 - \alpha).(LG_{i,j})
\end{equation} 

\noindent where $\alpha$ is a knob that lies in [0,1] and controls
the relative importance of hit rate and latency gain.







\noindent{\bf Step 3. Resource Constraints: }
The memory and computational constraint remain the same as in
the initial formulation - Eqn. \ref{eqn:memory-constraint} 
and Eqn. \ref{eqn:comp-constraint} respectively.

\noindent {\bf Objective:}
Finally, the objective of our formulation is to {\em maximize} the 
sum of scores for chosen variants:
\begin{equation}
	\text{\textbf{max.}} \sum_{i=1}^{N}\sum_{j=1}^{K}b_{i,j}.S_{i,j}
\end{equation}

The computed values of $b_{i,j}$ then determine which learned cache
variants should be used along with the base DNN during inference.

\subsection{Late Binding Benefits}
\label{appendix:eval}

\begin{figure}[t!]
        \centering
        \includegraphics[clip,width=0.8\columnwidth]{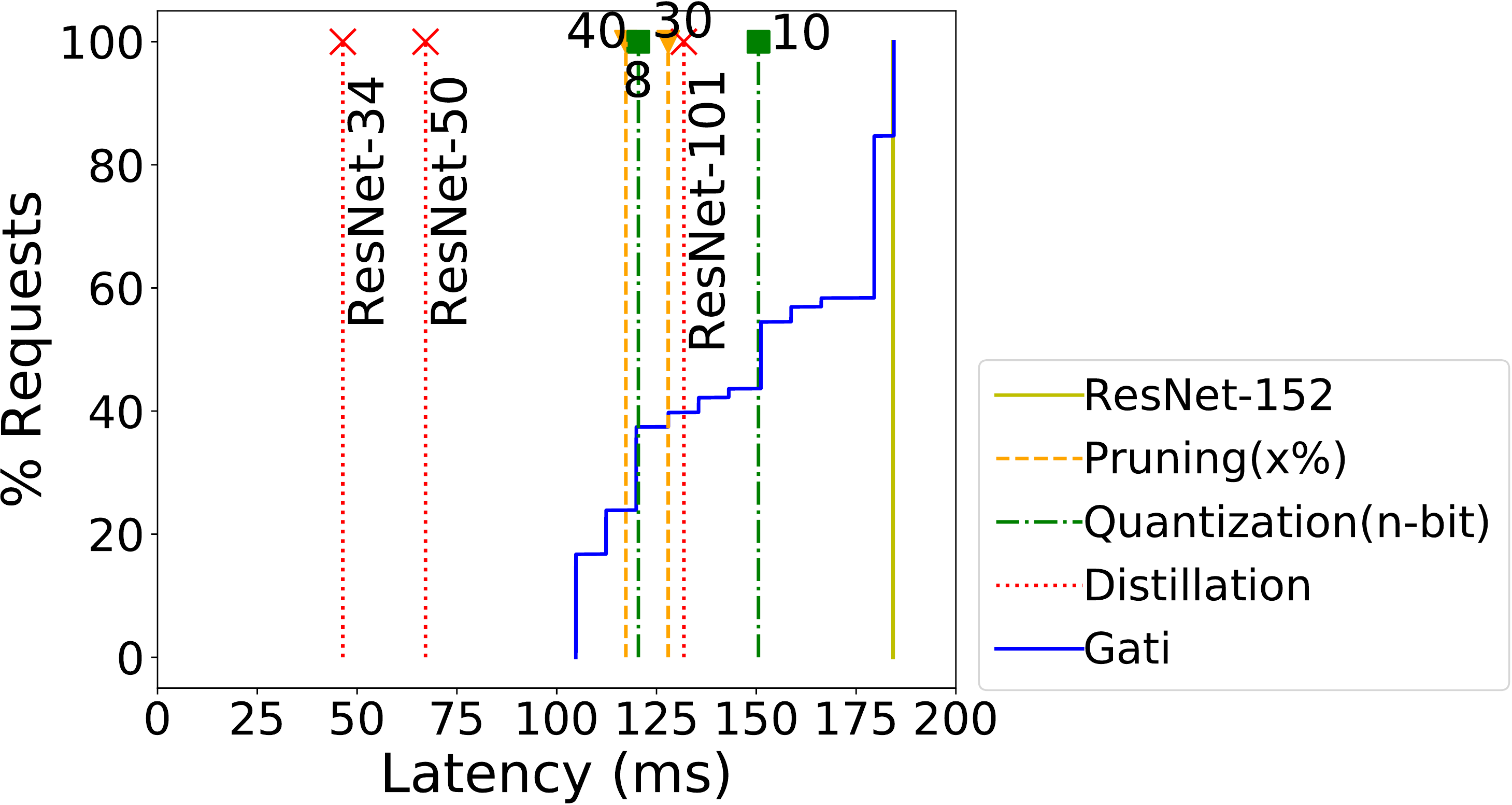}%
        \caption{\textbf {[CPU Inference]CDF of request latencies comparing \name{} vs baselines for ResNet-152 on CIFAR-100 dataset.}}
        \label{fig:res152-macro-results-cpu_a}
\end{figure}

\begin{figure}[t!]
        \centering
        \includegraphics[clip,width=0.65\columnwidth]{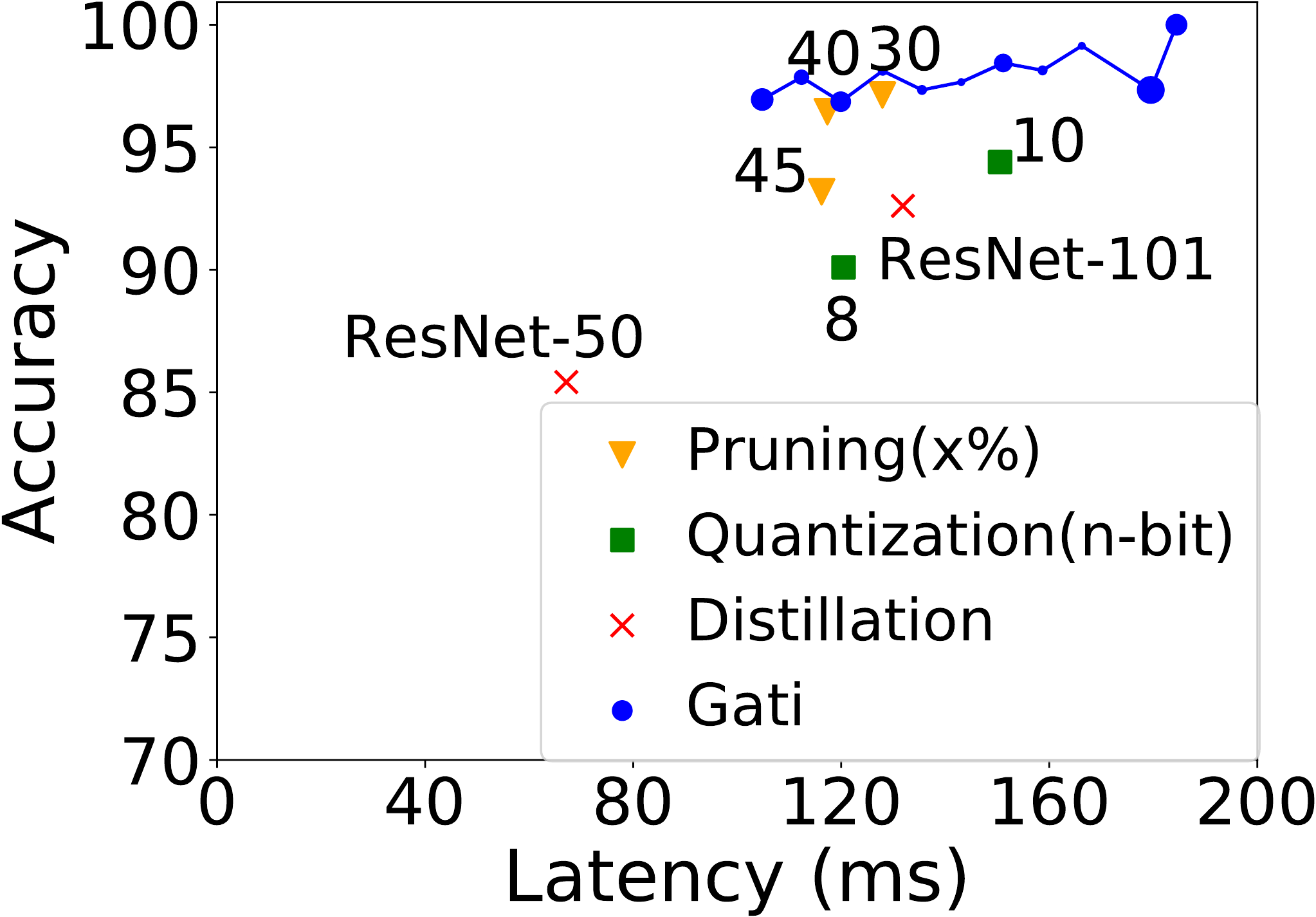}%
        \caption{\textbf {[CPU Inference]Accuracy vs latency trade-off for ResNet-152 on CIFAR-100 dataset. Each dot for \name{} is sized in proportion to the number of inference requests
served at that particular latency point.} }
        \label{fig:res152-macro-results-cpu_b}
\end{figure}

\noindent {\bf (i) CIFAR-100 on ResNet-152: }
From Figure~\ref{fig:res152-macro-results-cpu_a},\ref{fig:res152-macro-results-cpu_b}, we observe that \name{} exhibits an average latency of {\bf 148.34 ms} (CPU), 
which is {\bf 1.24$X$} lower than the latency of running the entire DNN. \name{} exhibits a spectrum of latencies
 with an overall accuracy of {\bf 97.79\%} (CPU) with respect to the base ResNet-152 model. We also observe that 
 for ResNet-152, {\bf 15.32\%} of inputs run through the entire DNN, which is a large increase comparing to {\bf 3.51\%} (ResNet-18)
 and {\bf 1.53\%} (ResNet-50) on CIFAR-10 dataset. The reason is that CIFAR-100 has a lesser proportion of easier examples,
 which reduces the chances for early hits.

\begin{figure}[t!]
        \centering
        \includegraphics[clip,width=0.8\columnwidth]{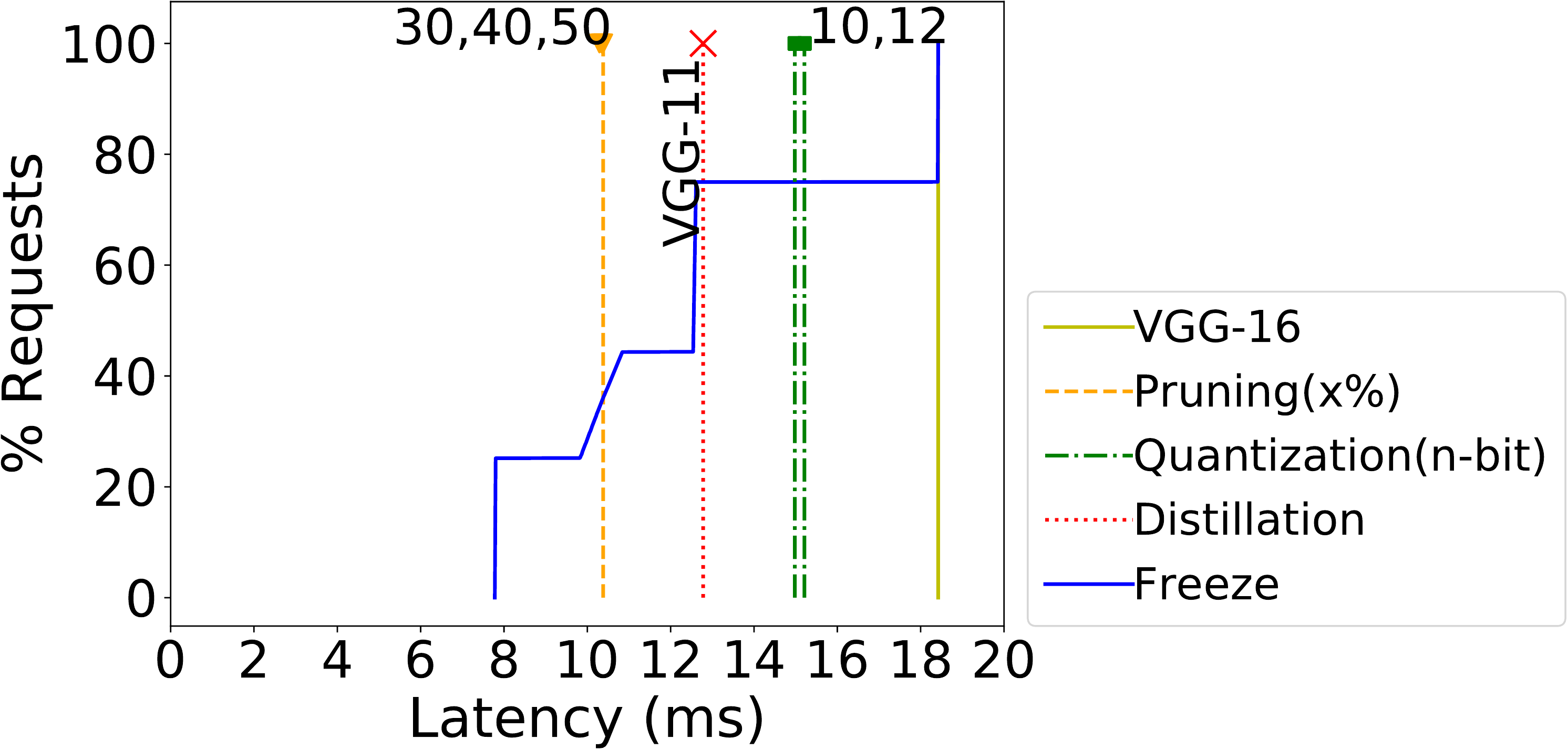}%
        \caption{\textbf {[CPU Inference]CDF of request latencies comparing \name{} vs baselines for VGG-16 on Google Voice dataset.}}
        \label{fig:vgg-macro-results-cpu_a}
\end{figure}

\begin{figure}[t!]
        \centering
        \includegraphics[clip,width=0.8\columnwidth]{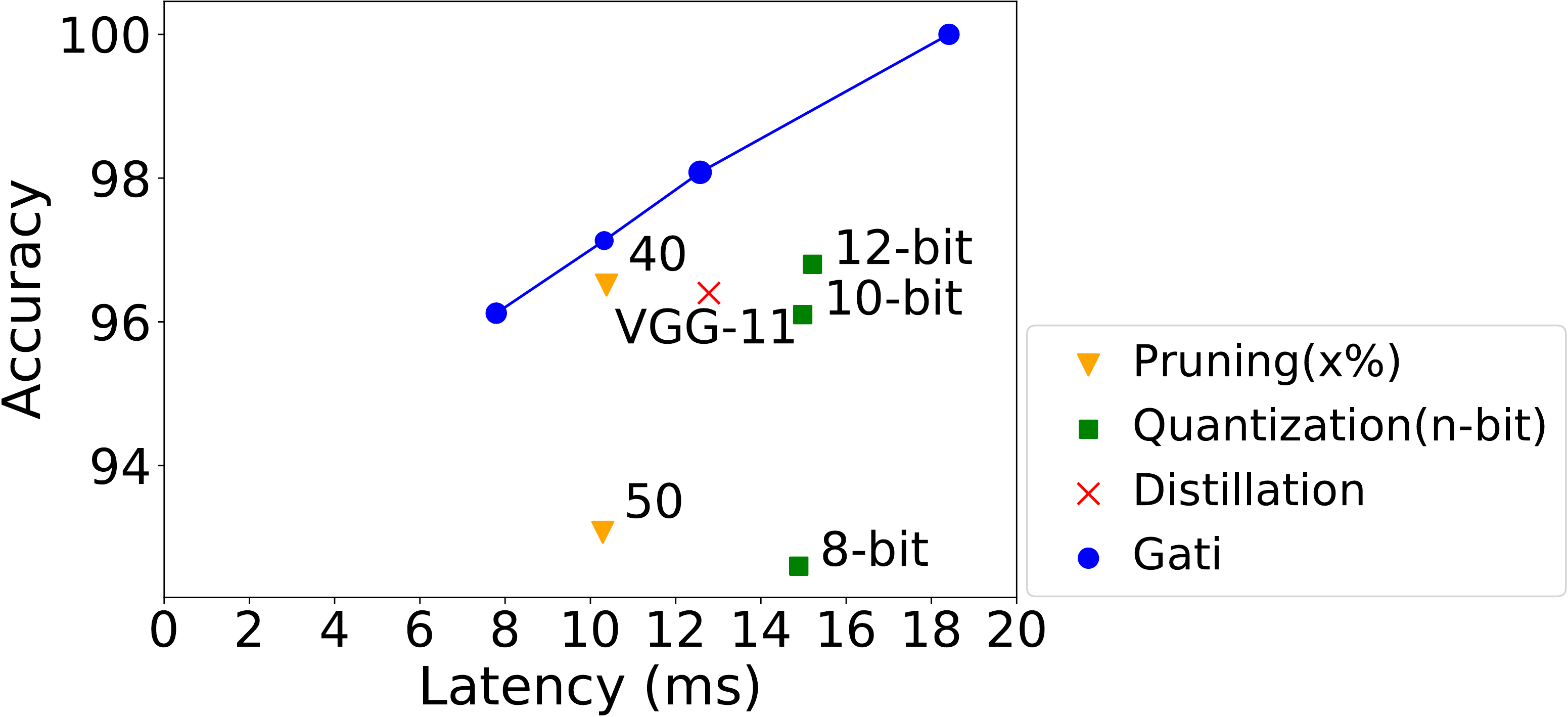}%
        \caption{\textbf{[CPU Inference]Accuracy vs latency trade-off for VGG-16 on Google Voice dataset dataset. Size of each \name{} marker is proportional to \% of requests at the corresponding latency point.}}
        \label{fig:vgg-macro-results-cpu_b}
\end{figure}

\begin{figure}[t!]
        \centering
        \includegraphics[clip,width=0.8\columnwidth]{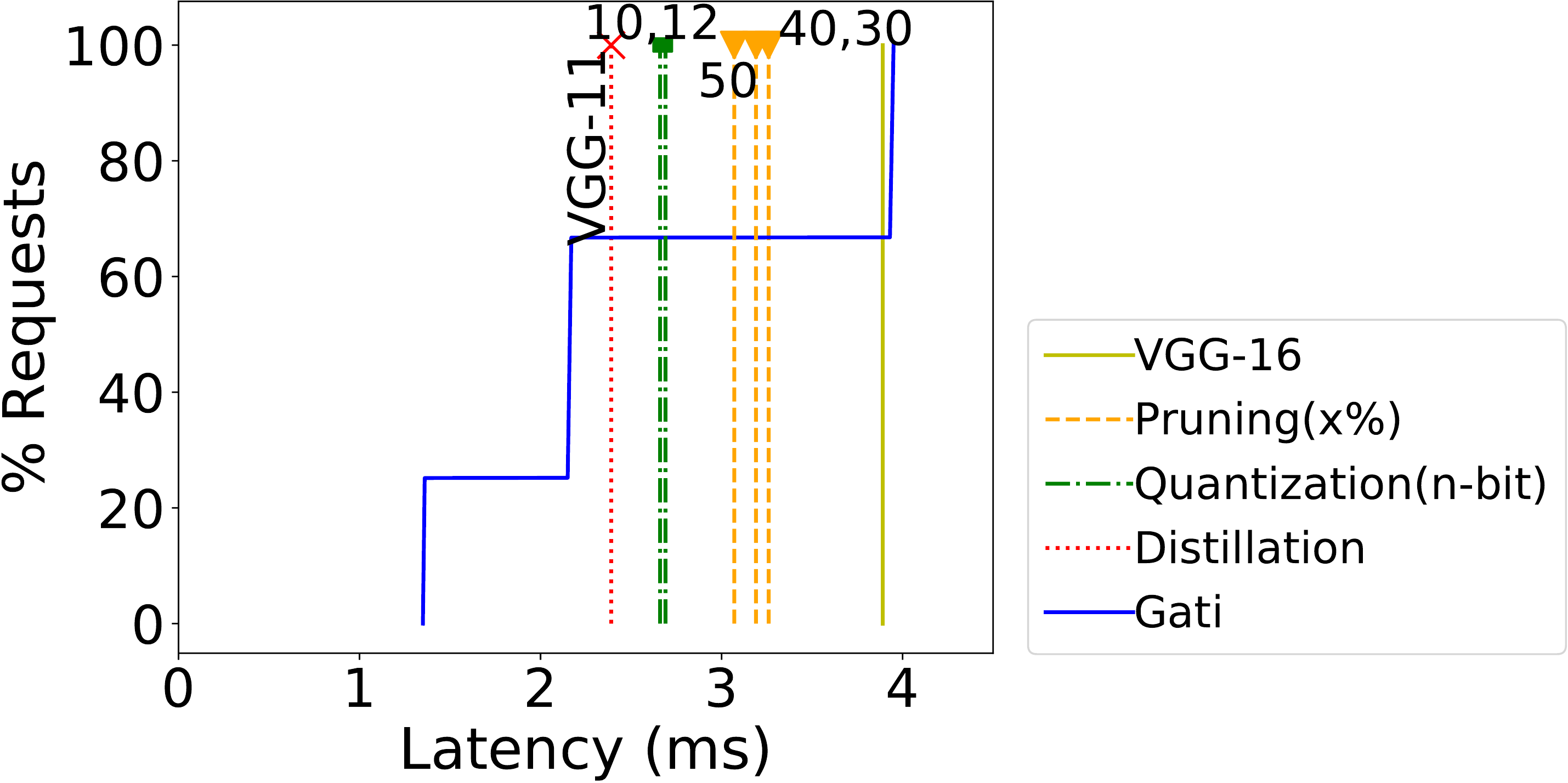}%
        \caption{\textbf{[GPU Inference]CDF of request latencies comparing \name{} vs baselines for VGG-16 on Google Voice dataset.}}
        \label{fig:vgg-macro-results-gpu_a}
\end{figure}

\begin{figure}[t!]
        \centering
        \includegraphics[clip,width=0.8\columnwidth]{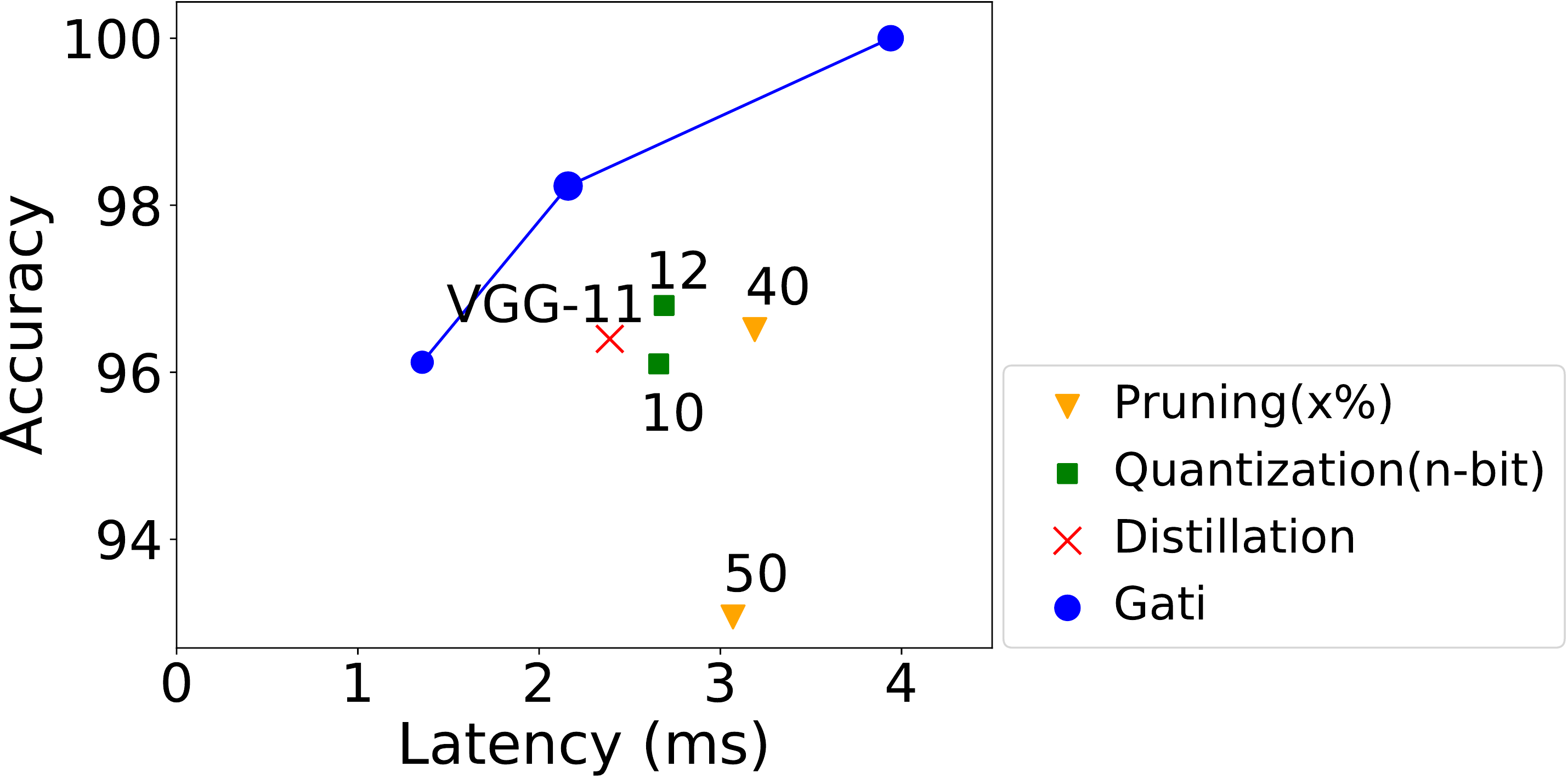}%
        \caption{\textbf{[GPU Inference]Accuracy vs latency trade-off for VGG-16 on Google Voice dataset dataset. Size of each \name{} marker is proportional to \% of requests at the corresponding latency point.}}
        \label{fig:vgg-macro-results-gpu_b}
\end{figure}

\noindent {\bf (ii) Voice Commands on VGG-16: }
From Figure~\ref{fig:vgg-macro-results-cpu_a},\ref{fig:vgg-macro-results-cpu_b},\ref{fig:vgg-macro-results-gpu_a},\ref{fig:vgg-macro-results-gpu_b}
we observe that \name{} exhibits an average latency of {\bf 9.42 ms} (CPU) and {\bf 2.52 ms} (GPU), which is
{\bf 1.96$X$} and {\bf 1.54 $X$} lower than the latency of running the entire DNN.
\name{} exhibits a spectrum of latencies with an overall accuracy of {\bf 97.88\%} (CPU)
and {\bf 98.28\%} (GPU) with respect to the base VGG-16 model.

\begin{figure}[t!]
        \centering
        \includegraphics[clip,width=0.8\columnwidth]{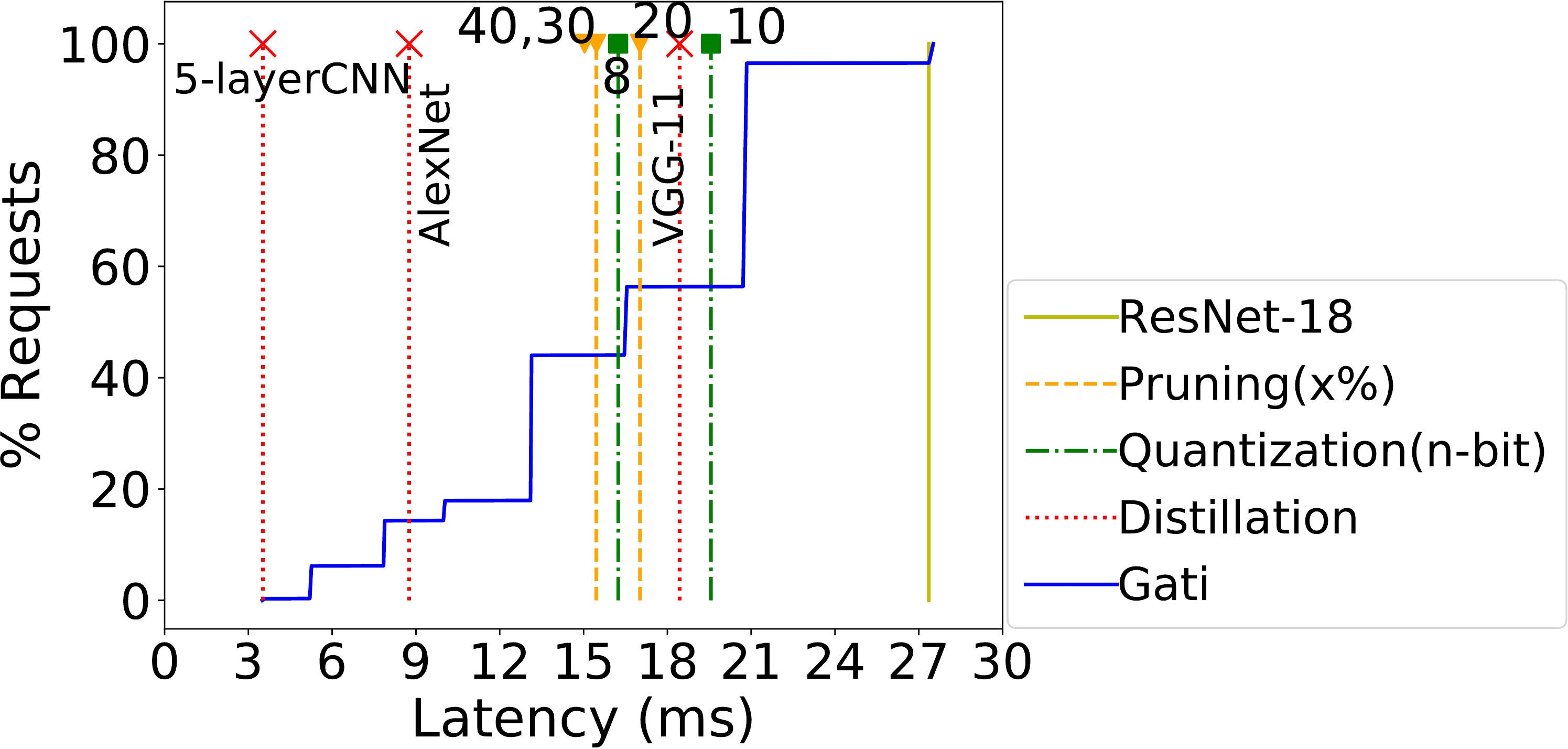}%
        \caption{\textbf{[CPU Inference]CDF of request latencies comparing \name{} vs baselines for ResNet-18 on CIFAR-10 dataset.}}
        \label{fig:res18-macro-results-cpu_a}
\end{figure}

\begin{figure}[t!]
        \centering
        \includegraphics[clip,width=0.8\columnwidth]{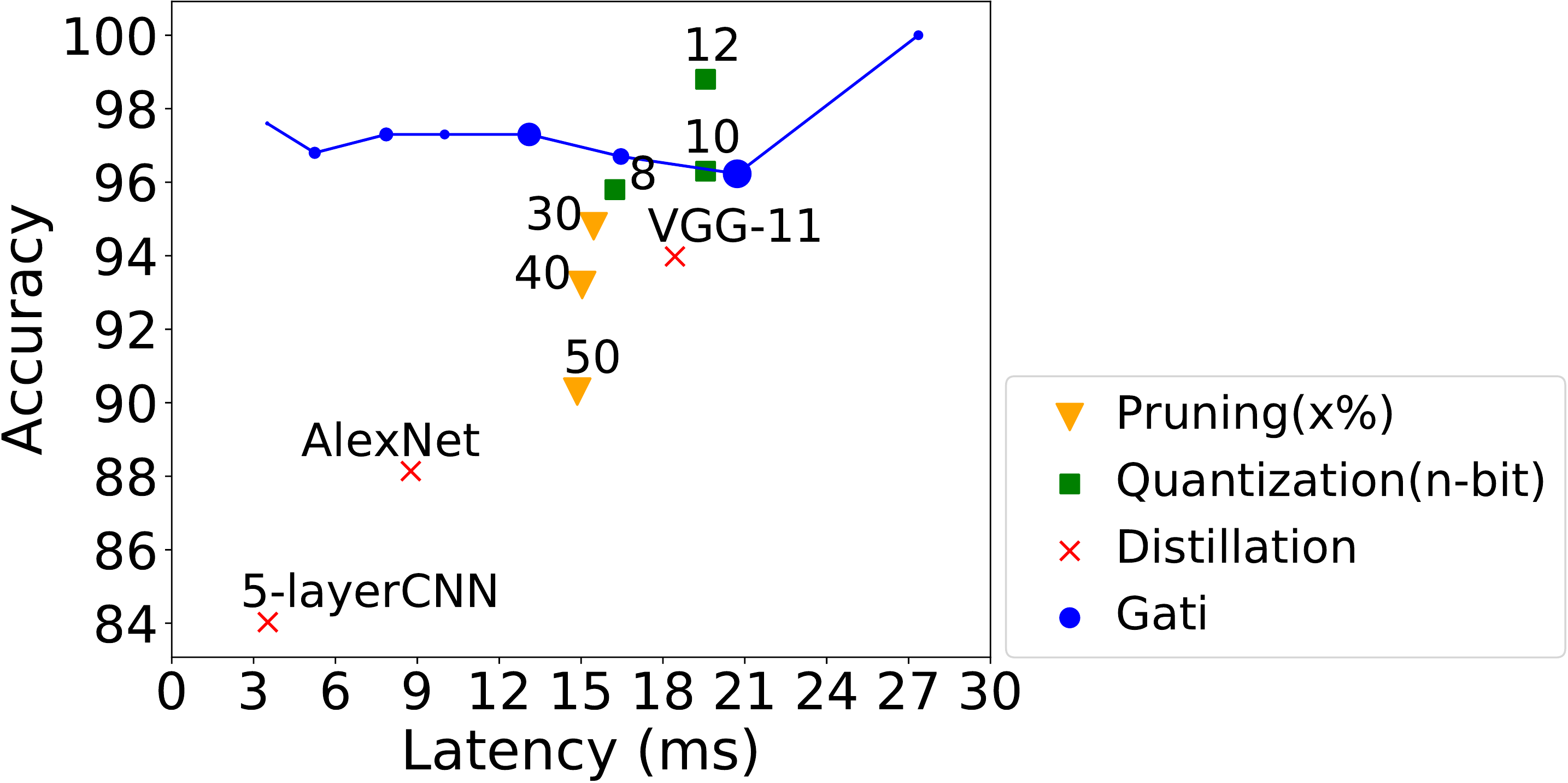}%
        \caption{\textbf{[CPU Inference]Accuracy vs latency trade-off for ResNet-18 on CIFAR-10 dataset. Size of each \name{} marker is proportional to \% of requests at the corresponding latency point.}}
        \label{fig:res18-macro-results-cpu_b}
\end{figure}

\begin{figure}[t!]
        \centering
        \includegraphics[clip,width=0.8\columnwidth]{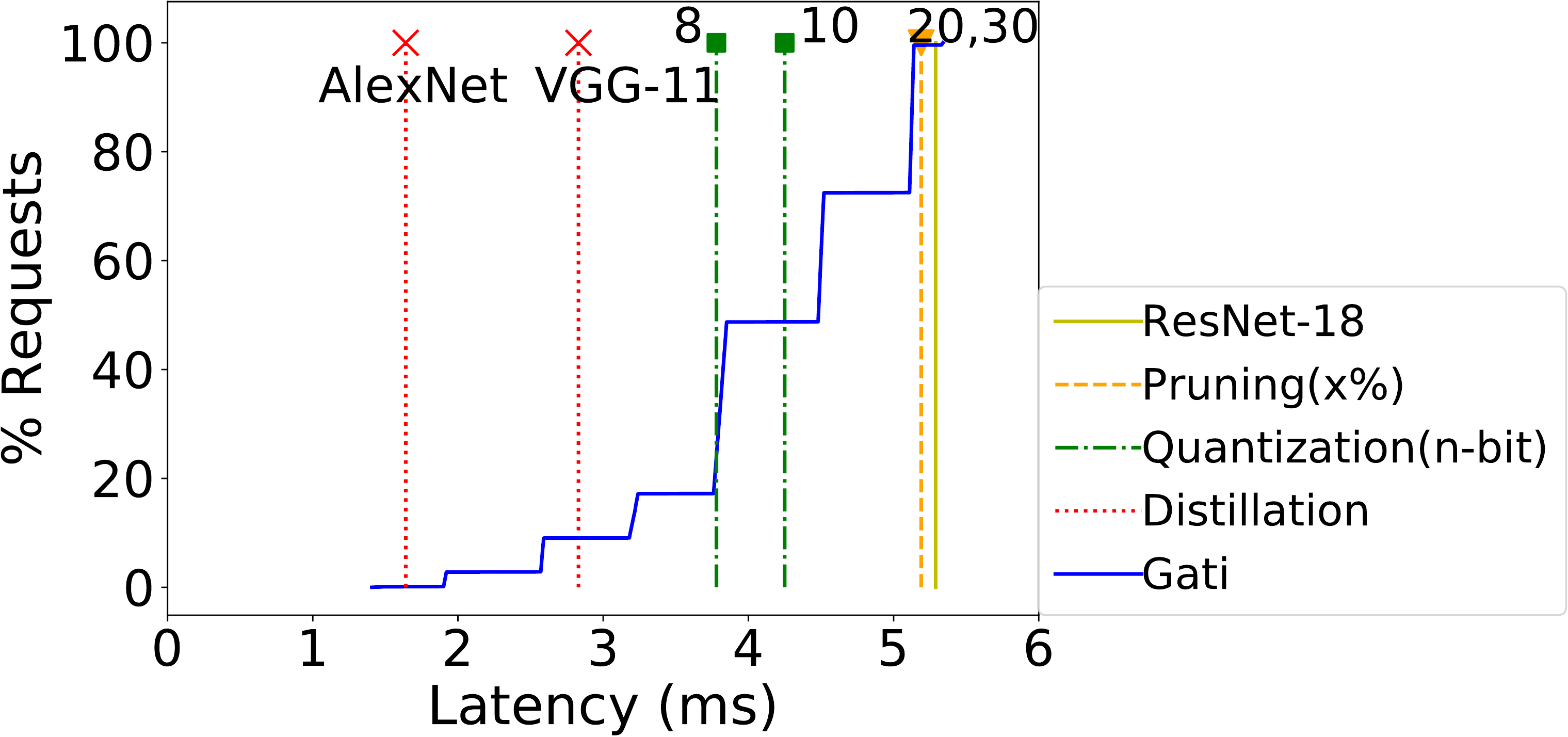}%
        \caption{\textbf{[GPU Inference]CDF of request latencies comparing \name{} vs baselines for ResNet-18 on CIFAR-10 dataset.}}
        \label{fig:res18-macro-results-gpu_a}
\end{figure}

\begin{figure}[t!]
        \centering
        \includegraphics[clip,width=0.8\columnwidth]{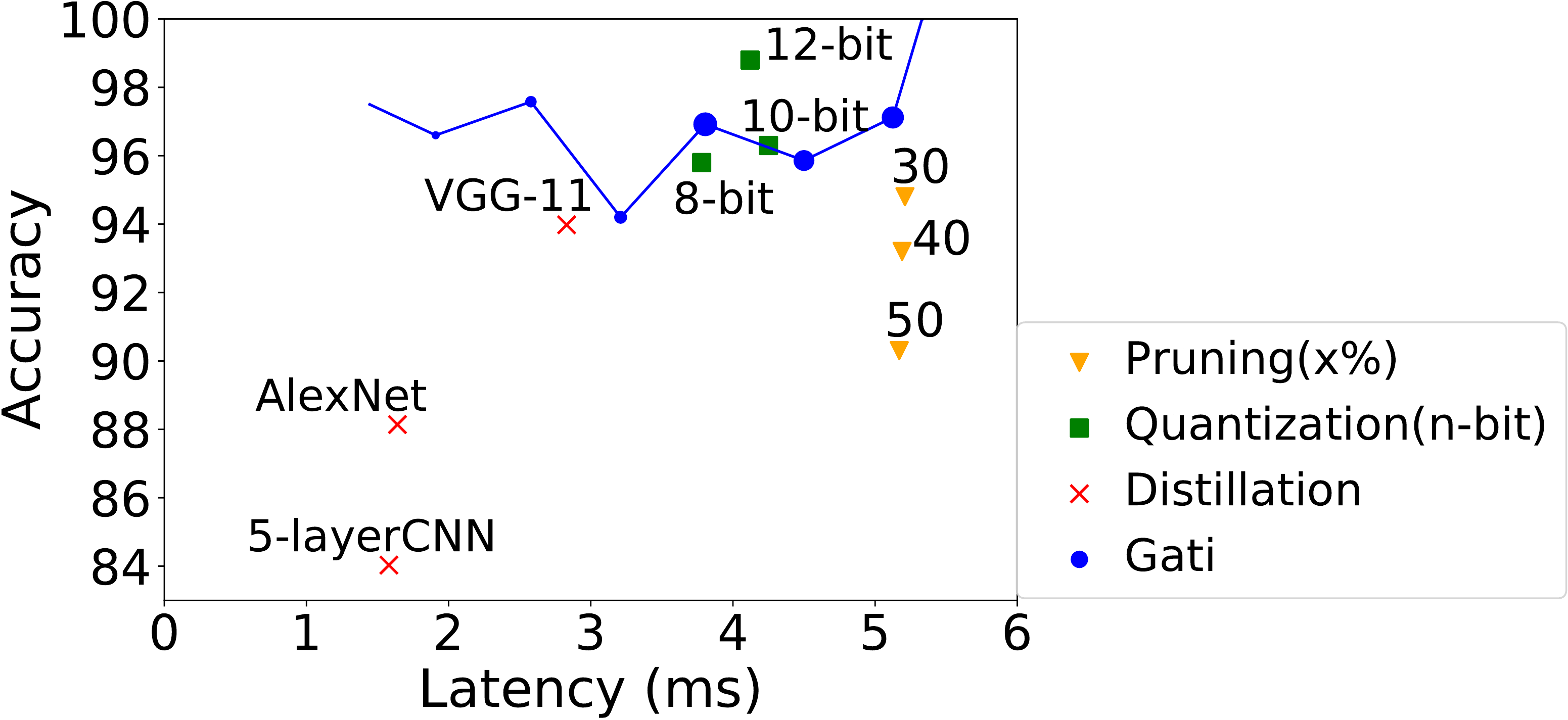}%
        \caption{\textbf{[GPU Inference]Accuracy vs latency trade-off for ResNet-18 on CIFAR-10 dataset. Size of each \name{} marker is proportional to \% of requests at the corresponding latency point.}}
        \label{fig:res18-macro-results-gpu_b}
\end{figure}

\noindent {\bf (iii) CIFAR-10 on ResNet-18: }
From Figure~\ref{fig:res18-macro-results-cpu_a},\ref{fig:res18-macro-results-cpu_b},\ref{fig:res18-macro-results-gpu_a},\ref{fig:res18-macro-results-gpu_b}
we observe that \name{} exhibits an average latency of {\bf 16.04 ms} (CPU) and {\bf 4.15 ms} (GPU), which is
{\bf 1.72$X$} and {\bf 1.28 $X$} lower than the latency of running the entire DNN.
\name{} exhibits a spectrum of latencies with an overall accuracy of {\bf 96.86\%} (CPU) 
and {\bf 96.54\%} (GPU) with respect to the base ResNet-18 model.

\subsubsection{Learned Caches Design Analysis}

\label{subsec:lc_analysis}

\begin{figure}[t!]
        \centering
        \includegraphics[clip,width=0.8\columnwidth]{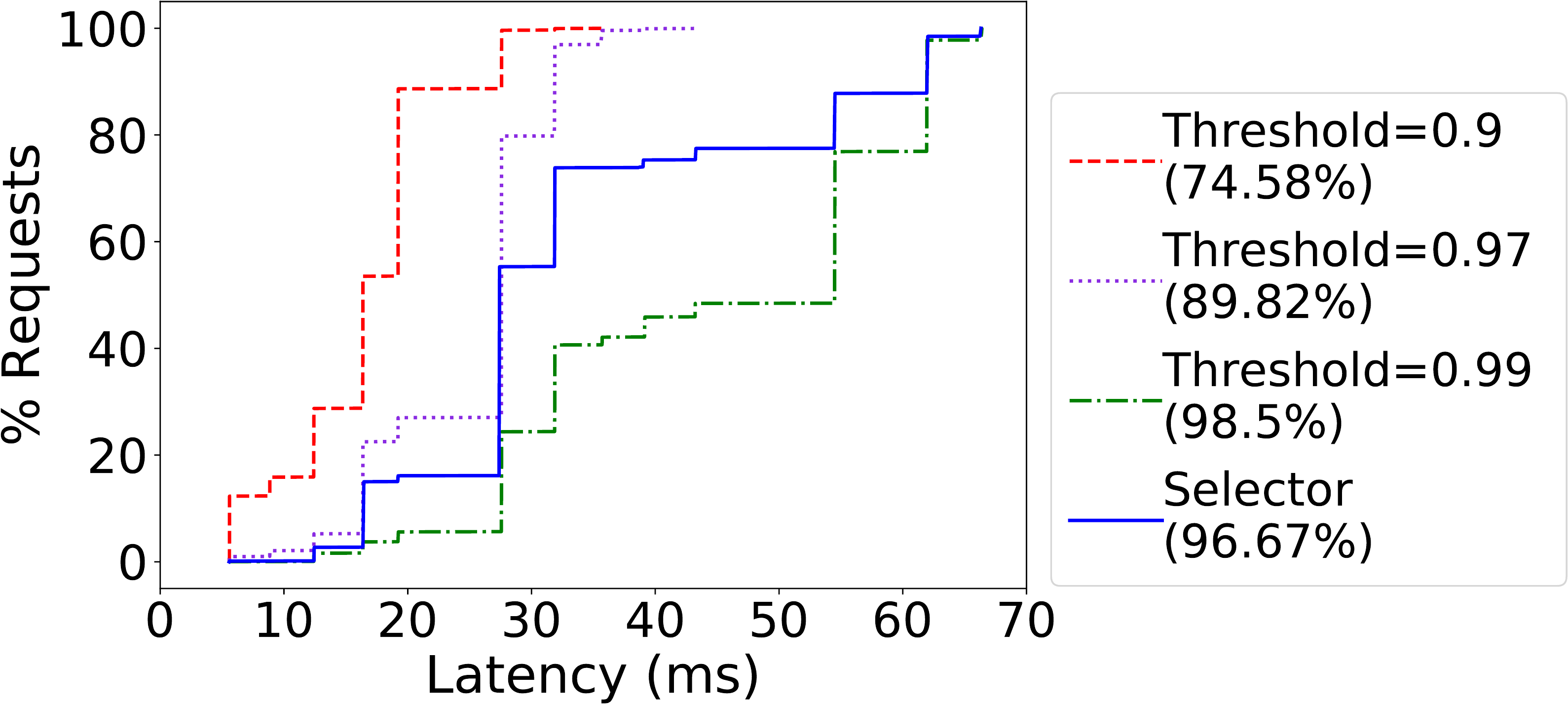}%
        \caption{\textbf{[CPU Inference]Benefits of decoupling prediction and selection for ResNet-50 on CIFAR-10 dataset. Accuracies for different schemes are labeled in brackets. }}
        \label{fig:decoupling-predictor-and-selector}
\end{figure}

\begin{figure}[t!]
        \centering
        \includegraphics[clip,width=0.6\columnwidth]{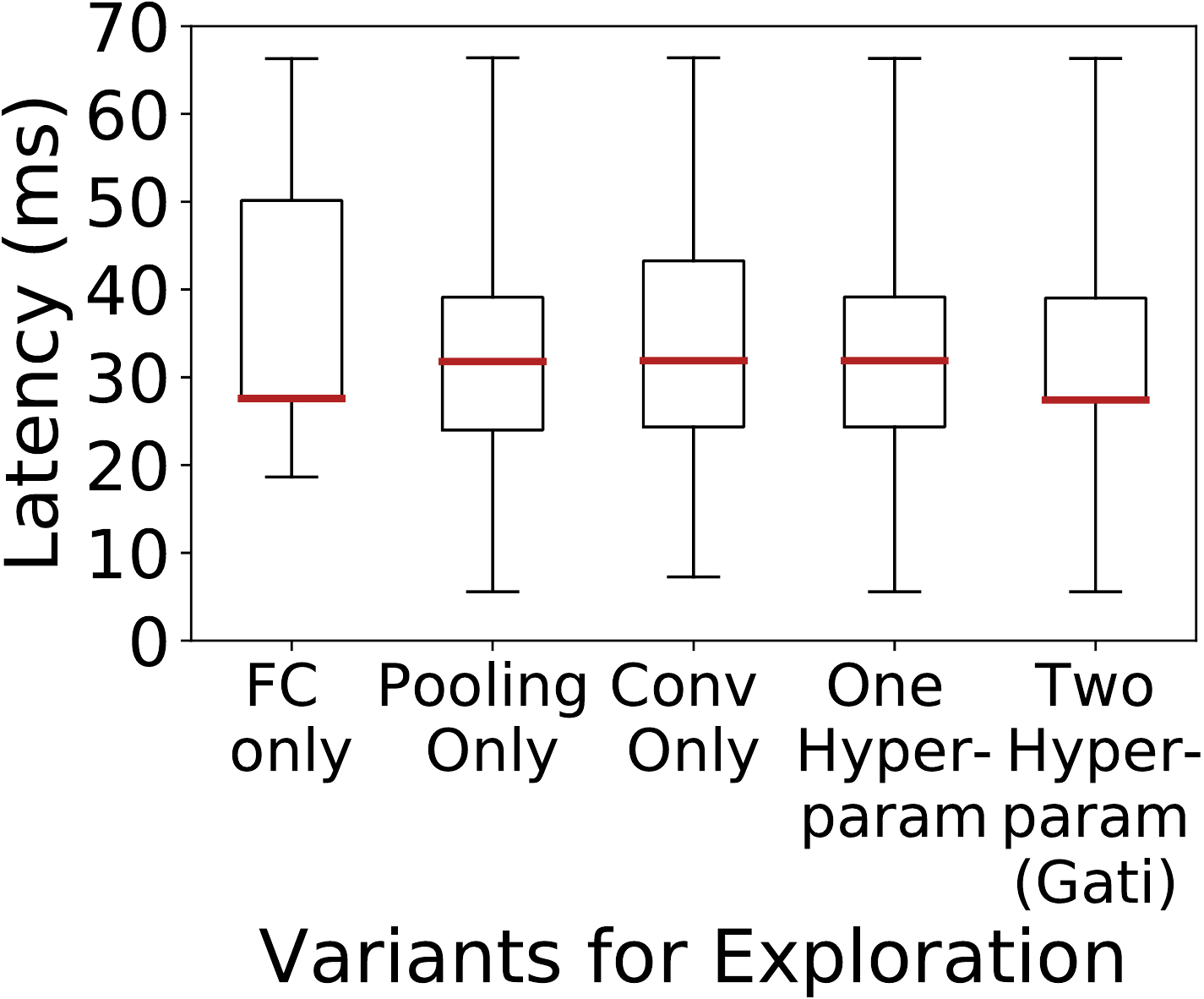}%
        \caption{\textbf{[CPU Inference]Benefits of exploring multiple predictor network architectures for ResNet-50 on CIFAR-10 dataset.}}
        \label{fig:exploration-benefits}
\end{figure}

\begin{figure}[t!]
        \centering
        \includegraphics[clip,width=0.6\columnwidth]{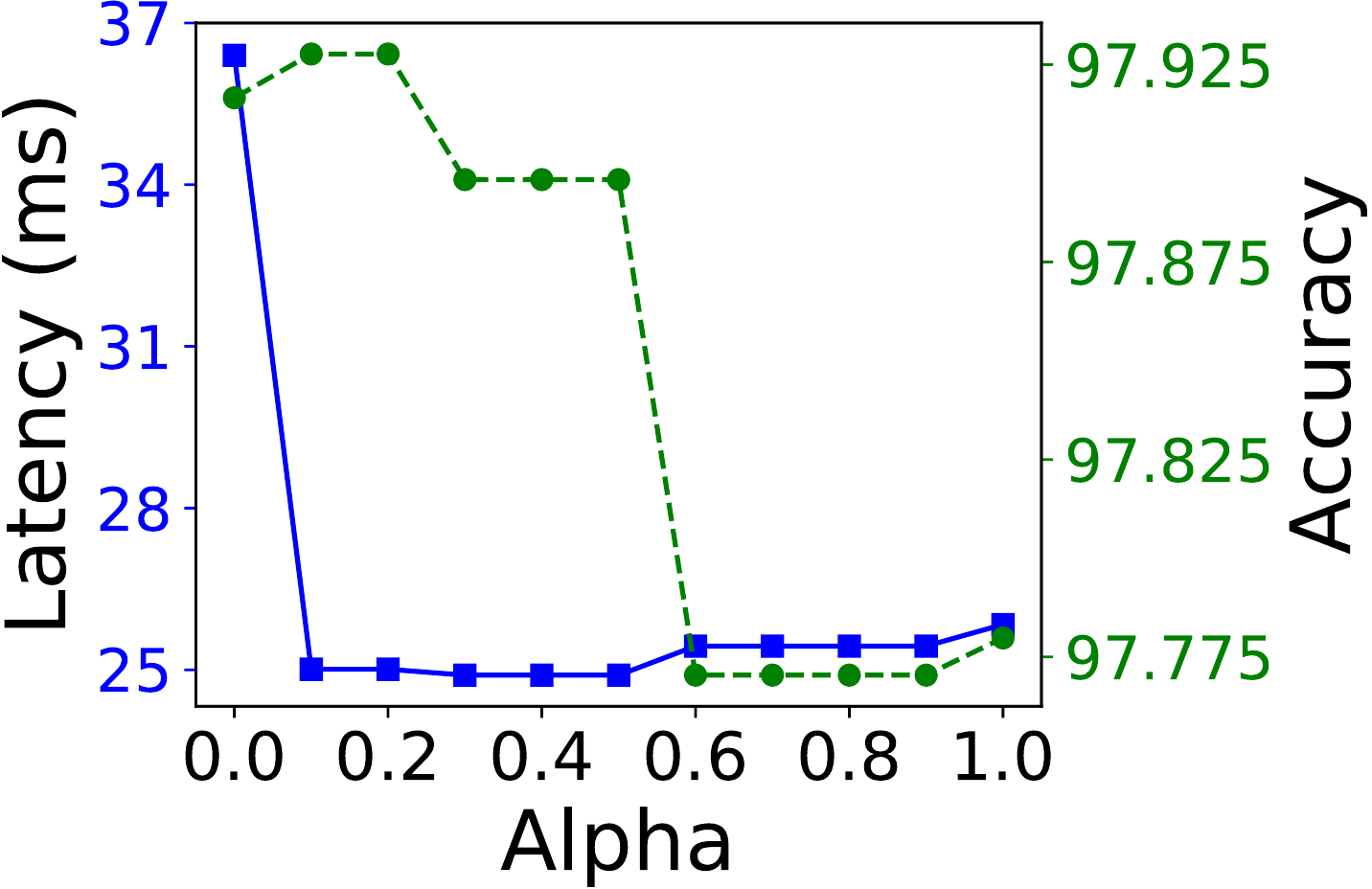}%
        \caption{\textbf{[CPU Inference]Latency-accuracy trade-off for different $\alpha$ values in the composition phase for ResNet-50 on CIFAR-10 dataset.
        }}
        \label{fig:alpha-regime}
\end{figure}









We analyze the impact of specific design choices 
for learned caches adopted by \name{}
that enable better latencies while ensuring that the accuracy is
close to the target accuracy.

\noindent {\bf Using distillation loss function in predictor network: }
We compare the benefit of using a loss function inspired by distillation
against a standard cross-entropy loss function for ResNet-50 on CIFAR-10.
We observe that the distillation loss function gives a greater
number of cache hits at earlier layers.
The distillation loss function
results in more accurate predictor networks, thereby allowing the selector network
to infer more points as confident cache hits.
Overall, we observe that this loss function yields {\bf 5.9\%} improvement in
average latency.

\noindent {\bf Decoupling prediction and selection in learned caches: }
Decoupling the decisions helps give cache hits with the desired
accuracy and hit rate properties.
We compare this choice against a baseline that uses only the
predictor network and establishes a threshold over the softmax scores
from the predictor network to infer cache hits.
We notice from Figure~\ref{fig:decoupling-predictor-and-selector} that applying
different thresholds on predictor networks
induces trade-offs between accuracy and hit rate that is hard to control.
The selector network provides a binary classification mechanism
that allows for optimal control of the trade-off between accuracy and hit rate.

\noindent {\bf Exploring multiple predictor network architectures: }
We evaluate the benefit of exploration by comparing against baselines
that consider either only one type of network architecture
or only one hyper-parameter for each architecture.
While there is not much impact overall accuracy, we observe
from Figure~\ref{fig:exploration-benefits}, that exploring more
variants leads to better latency benefits, since it provides
more data points for the composition phase to pick from.
The minimum latency (5.57 ms) by considering all variants matches
the minimum latency by considering only variants with pooling architecture.
Similarly, the median latency (27.5 ms) by considering all variants matches
the minimum latency by considering only variants with fully-connected architecture.
Exploration thus gives combined benefits of exploring individual variants.
A trade-off incurred in exploring multiple variants is that it requires more
resources during the initial cache construction phase. Developers can limit
the number of variants to be explored depending upon the resource availability.

\noindent {\bf Optimal cache composition: }
During the composition phase, \name{} evaluates multiple $\alpha$ values
and picks a value that minimizes the expected latency while maximizing accuracy.
For ResNet-50 on CIFAR-10, Figure~\ref{fig:alpha-regime} shows that $\alpha$ 
values of 0.2 provide optimal results.
We observe similar values for GPU based inference.
\name{} selects the final set of learned cache variants based on this value.

We evaluate the benefit of formulating the composition phase
as an optimization problem by comparing it to two greedy approaches -
(i) A latency greedy approach that greedily picks variants
with maximum latency gain while respecting computational and memory constraints.
(ii) A hit-rate greedy approach that greedily picks variants
with maximum hit-rate by respecting computational and memory constraints.
We observe that the optimization
problem has the most optimal latency profile, offering up to {\bf 5\%} improvement
in average latency.
This can be tied down to a better visibility
of the trade-offs between the various metrics for learned caches
that an optimization formulation can capture.

\noindent {\bf Hardware-aware profiling: }
In the composition phase, \name{} profiles the lookup latency
on the same target hardware where the model will be deployed for prediction serving.
We observe that profiling in such a hardware-aware manner
yields a {\bf 12.93\%} improvement in average latency.

\label{subsec:appendix_results}

\end{document}